\documentclass{article}
\pdfoutput=1
\usepackage[utf8]{inputenc} 
\usepackage[T1]{fontenc}    
\usepackage{hyperref}       
\usepackage{url}            
\usepackage{booktabs}       
\usepackage{amsfonts}       
\usepackage{nicefrac}       
\usepackage{microtype}      
\usepackage[square,sort,comma,numbers]{natbib}

\makeatletter
\renewcommand{\paragraph}{%
  \@startsection{paragraph}{4}%
  {\z@}{-2ex \@plus 10ex \@minus 0ex}{-0.5em}%
  {\normalfont\normalsize\bfseries}%
}
\makeatother

\title{Rényi Differential Privacy Mechanisms for Posterior Sampling}

\author{
    Joseph Geumlek\\
    University of California, San Diego\\
    \texttt{jgeumlek@cs.ucsd.edu} 
     \and
     Shuang Song \\
     University of California, San Diego\\
     \texttt{shs037@eng.ucsd.edu} 
     \and
     Kamalika Chaudhuri \\
     University of California, San Diego\\
     \texttt{kamalika@cs.ucsd.edu} 
}

\usepackage{amsmath,amssymb, amsthm}
\usepackage[utf8]{inputenc}

\usepackage{natbib}
\usepackage{amsmath}
\usepackage{amssymb}
\usepackage{amsfonts}
\usepackage{amsthm}
\usepackage{fullpage}
\usepackage{algorithm}
\usepackage{algorithmic}
\usepackage{graphicx}
\usepackage{color}
\usepackage{mdframed}

\usepackage{enumitem}
\setlistdepth{9}

\setlist[itemize,1]{label=$ $}
\setlist[itemize,2]{label=$ $}
\setlist[itemize,3]{label=$ $}
\setlist[itemize,4]{label=$ $}
\setlist[itemize,5]{label=$ $}
\setlist[itemize,6]{label=$ $}
\setlist[itemize,7]{label=$ $}
\setlist[itemize,8]{label=$ $}
\setlist[itemize,9]{label=$ $}

\renewlist{itemize}{itemize}{9}

\newtheorem{Thm}{Theorem}
\newtheorem{Lem}[Thm]{Lemma}
\newtheorem{Cor}[Thm]{Corollary}

\newtheorem{lemma}{Lemma}
\newtheorem{Observation}{Observation}
\newtheorem{theorem}{Theorem}
\newenvironment{proof2}{\noindent {\sc Proof:}}{$\Box$ \medskip} 

\newtheorem{assumption}{Assumption}

\newcommand{\calA}{{\mathcal{A}}}

\newcommand{\calX}{{\mathcal{X}}}
\newcommand{\calY}{{\mathcal{Y}}}

\def\argmax{\mathrm{argmax}}

\def\E{\mathbb{E}}

\def\x{\mathbf{x}}
\def\X{\mathbf{X}}

\def\Diff{Diff} 

\usepackage{xr}

\usepackage{chngcntr}
\usepackage{apptools}
\AtAppendix{\counterwithin{theorem}{section}}
\usepackage{geometry}

\usepackage{subcaption} 
\usepackage[usenames, dvipsnames]{xcolor}
\newtheorem{Defn}[theorem]{Definition}
\newtheorem{definition}[theorem]{Definition}
\newtheorem{observation}[theorem]{Observation}

\renewcommand{\exp}[1]{\text{exp}\left( #1 \right)}
\newcommand{\Expect}[2]{\mathbb{E}_{#1}\left[ #2 \right]}
\newcommand{\KL}[2]{D_{\text{KL}} \left(#1 \| #2\right)}

\newcommand{\reg}{\beta}

\renewcommand{\ln}[1]{\log\left(#1\right)}
\newcommand{\rdp}{\text{RDP}}
\newcommand{\rd}{\text{R\'enyi Divergence}}
\usepackage{bbm}

\newcommand{\minorheading}[1]{\subsection{#1}}
\begin{document}
\maketitle

\begin{abstract}
With the newly proposed privacy definition of Rényi Differential Privacy (RDP) in \citep{mironov2017renyi}, we re-examine the inherent privacy of releasing a single sample from a posterior distribution. We exploit the impact of the prior distribution in mitigating the influence of individual data points. In particular, we focus on sampling from an exponential family and specific generalized linear models, such as logistic regression. We propose novel RDP mechanisms as well as offering a new RDP analysis for an existing method in order to add value to the RDP framework. Each method is capable of achieving arbitrary RDP privacy guarantees, and we offer experimental results of their efficacy. 
\end{abstract}

\section{Introduction}
As data analysis continues to expand and permeate ever more facets of life, the concerns over the privacy of one's data grow too. Many results have arrived in recent years to tackle the inherent conflict of extracting usable knowledge from a data set without over-extracting or leaking the private data of individuals. Before one can strike a balance between these competing goals, one needs a framework by which to quantify what it means to preserve an individual's privacy.

Since 2006, Differential Privacy (DP) has reigned as the privacy framework of choice \citep{dwork2006our}. It quantifies privacy by measuring how indistinguishable the mechanism is across whether or not any one individual is in or out of the data set. This gave not just privacy semantics, but also robust mathematical guarantees. However, the requirements have been cumbersome for utility, leading to many proposed relaxations. One common relaxation is approximate DP, which allows arbitrarily bad events to occur with probability at most $\delta$. A more recent relaxation is Rényi Differential Privacy (RDP) proposed in \citep{mironov2017renyi}, which uses the measure of Rényi divergences to smoothly vary between bounding the average and maximum privacy loss. However, RDP has very few mechanisms compared to the more established approximate DP. We expand the RDP repertoire with novel mechanisms inspired by Rényi divergences, as well as re-analyzing an existing method in this new light.

Inherent to DP and RDP is that there must be some uncertainty in the mechanism; they can not be deterministic. Many privacy methods have been motivated by exploiting pre-existing sources of randomness in machine learning algorithms. One promising area has been Bayesian data analysis, which focuses on maintaining and tracking the uncertainty within probabilistic models. Posterior sampling is prevalent in many Bayesian methods, serving to introduce randomness that matches the currently held uncertainty.

We analyze the privacy arising from posterior sampling as applied to two domains: sampling from exponential family and Bayesian logistic regression. Along with these analyses, we offer tunable mechanisms that can achieve stronger privacy guarantees than directly sampling from the posterior. These mechanisms work via controlling the relative strength of the prior in determining the posterior, building off the common intuition that concentrated prior distributions can prevent overfitting in Bayesian data analysis. We experimentally validate our new methods on synthetic and real data.

\section{Background}



\minorheading{Privacy Model.} We say two data sets $\X$ and $\X'$ are {\em{neighboring}} if they differ in the private record of a single {\em{individual}} or person. We use $n$ to refer to the number of records in the data set.

\begin{Defn}{Differential Privacy (DP) \citep{dwork2006our}.}
A randomized mechanism $\calA(\X)$ is said to be $(\epsilon,\delta)$-differentially private if for any subset $U$ of the output range of $\calA$ and any neighboring data sets $\X$ and $\X'$, we have $p(\calA(\X) \in U) \le \exp{\epsilon} p(\calA(\X') \in U) + \delta$.

\end{Defn}

DP is concerned with the difference the participation of a individual might have on the output distribution of the mechanism. When $\delta>0$, it is known as approximate DP while the $\delta = 0$ case is known as pure DP. The requirements for DP can be phrased in terms of a privacy loss variable, a random variable that captures the effective privacy loss of the mechanism output.

\begin{Defn}{Privacy Loss Variable \citep{bun2016concentrated}.}
We can define a random variable $Z$ that measures the privacy loss of a given output of a mechanism across two neighboring data sets $\X$ and $\X'$.
\begin{equation}
Z = \log \frac{p(\calA(\X) = o)}{p(\calA(\X') = o)} \bigg|_{o \sim \calA(\X)}
\end{equation}
\end{Defn}

$(\epsilon,\delta)$-DP is the requirement that for any two neighboring data sets $Z \le \epsilon$ with probability at least $1-\delta$. The exact nature of the trade-off and semantics between $\epsilon$ and $\delta$ is subtle, and choosing them appropriately is difficult. For example, setting $\delta = 1/n$ permits $(\epsilon,\delta)$-DP mechanisms that always violate the privacy of a random individual. However, there are other ways to specify that a random variable is mostly small. One such way is to bound the Rényi divergence of $\calA(\X)$ and $\calA(\X')$.



\begin{Defn}{Rényi Divergence \citep{bun2016concentrated}.}
The Rényi divergence of order $\lambda$ between the two distributions $P$ and $Q$ is defined as
\begin{align}
D_\lambda(P||Q) &= \frac{1}{\lambda-1} \log \int P(o)^\lambda Q(o)^{1-\lambda} do \mbox{.}
\end{align}
\end{Defn}

As $\lambda \rightarrow \infty$, Rényi divergence becomes the {\em{max divergence}}; moreover, setting $P =\calA(\X)$ and $Q=\calA(\X')$ ensures that $D_{\lambda}(P||Q) = \frac{1}{\lambda-1} \log \E_Z[e^{(\lambda-1)Z}]$, where $Z$ is the privacy loss variable. Thus, a bound on the Rényi divergence over all orders $\lambda\in(0,\infty)$ is equivalent to $(\epsilon,0)$-DP, and as $\lambda \rightarrow 1$, this approaches the expected value of $Z$ equal to $KL(\calA(\X)||\calA(\X'))$.  This leads us to Rényi Differential Privacy, a flexible privacy notion that covers this intermediate behavior.

\begin{Defn}{Rényi Differential Privacy (RDP) \citep{mironov2017renyi}.}
A randomized mechanism $\calA(\X)$ is said to be $(\lambda,\epsilon)$-Rényi differentially private if for any neighboring data sets $\X$ and $\X'$ we have $D_\lambda(\calA(\X)||\calA(\X')) \le \epsilon$.
\end{Defn}

The choice of $\lambda$ in RDP is used to tune how much concern is placed on unlikely large values of $Z$ versus the average value of $Z$. One can consider a mechanism's privacy as being quantified by the entire curve of $\epsilon$ values associated with each order $\lambda$,  but the results of \citep{mironov2017renyi} show that almost identical results can be achieved when this curve is known at only a finite collection of possible $\lambda$ values.

\minorheading{Posterior Sampling.} In Bayesian inference, we have a model class $\Theta$, and are given observations $x_1, \ldots, x_n$ assumed to be drawn from a $\theta \in \Theta$. Our goal is to 
maintain our beliefs about $\theta$ given the observational data in the form of the posterior distribution $p(\theta | x_1, \ldots, x_n)$. This is often done in the form of drawing samples from the posterior.

Our goal in this paper is to develop privacy preserving mechanisms for two popular and simple posterior sampling methods. The first is sampling from the exponential family posterior, which we address in Section~\ref{sec:expfam}; the second is sampling from posteriors induced by a subset of Generalized Linear Models, which we address in Section~\ref{sec:glm}.






\minorheading{Related Work.} Differential privacy has emerged as the gold standard for privacy in a number of data analysis applications -- see~\cite{dwork2014algorithmic, sarwate2013signal} for surveys. Since enforcing pure DP sometimes requires the addition of high noise, a number of relaxations have been proposed in the literature. The most popular relaxation is approximate DP~\cite{dwork2006our}, and a number of uniquely approximate DP mechanisms have been designed by~\cite{dwork2009differential, thakurtha2013differentially, CHS14, bun2015differentially} among others. However, while this relaxation has some nice properties, recent work~\cite{mironov2017renyi, mcsherry17} has argued that it can also lead privacy pitfalls in some cases. Approximate differential privacy is also related to, but is weaker than, the closely related $\delta$-probabilistic privacy~\cite{machanavajjhala2008privacy} and $(1, \epsilon, \delta)$-indistinguishability~\cite{chaudhurimishra2006}.

Our privacy definition of choice is Rényi differential privacy~\cite{mironov2017renyi}, which is motivated by two recent relaxations -- concentrated DP~\cite{dwork2016concentrated} and z-CDP~\cite{bun2016concentrated}. Concentrated DP has two parameters, $\mu$ and $\tau$, controlling the mean and concentration of the privacy loss variable. Given a privacy parameter $\alpha$, z-CDP essentially requires $(\lambda, \alpha \lambda)$-RDP for all $\lambda$. While~\cite{bun2016concentrated, dwork2016concentrated, mironov2017renyi} establish tighter bounds on the privacy of existing differentially private and approximate DP mechanisms, we provide mechanisms based on posterior sampling from exponential families that are uniquely RDP. RDP is also a generalization of the notion of KL-privacy~\cite{wang2016average}, which has been shown to be related to generalization in machine learning.

There has also been some recent work on privacy properties of Bayesian posterior sampling; however most of the work has focused on establishing pure or approximate DP. \cite{Dimitrakakis2014} establishes conditions under which some popular Bayesian posterior sampling procedures directly satisfy pure or approximate DP. \cite{wang2015privacy} provides a pure DP way to sample from a posterior that satisfies certain mild conditions by raising the temperature. \cite{foulds2016theory, zhang2016differential} provide a simple statistically efficient algorithm for sampling from exponential family posteriors. \cite{without_sensitivity} shows that directly sampling from the posterior of certain GLMs, such as logistic regression, with the right parameters provides approximate differential privacy. While our work draws inspiration from all~\cite{Dimitrakakis2014, wang2015privacy, without_sensitivity}, the main difference between their and our work is that we provide RDP guarantees.


\section{RDP Mechanisms based on Exponential Family Posterior Sampling}
\label{sec:expfam}

In this section, we analyze the Rényi divergences between distributions from the same exponential family, which will lead to our RDP mechanisms for sampling from exponential family posteriors.


\subsection{Background: Exponential Families} \label{sec:expfambackground}

This section will give a in-depth explanation of exponential families and the properties of them we exploit in our analysis.

An exponential family is a family of probability distributions over $x \in \calX$ indexed by the parameter $\theta \in \Theta \subseteq \mathbb{R}^d$ that can be written in this canonical form for some choice of functions $h: \calX \rightarrow \mathbb{R}$, $S: \calX \rightarrow \mathbb{R}^d$, and $A: \Theta \rightarrow \mathbb{R}$:

\begin{equation}
p(x_1,\ldots,\x_n | \theta) = (\prod_{i=1}^n h(x_i)) \exp{ (\sum_{i=1}^n S(x_i)) \cdot \theta - n\cdot A(\theta)  } \label{eqn:genexpfam} \mbox{ .}
\end{equation}

We call $h$ the base measure, $S$ the sufficient statistics of $x$, and $A$ as the log-partition function of this family.  Note that the data $\{x_1,\ldots,\x_n\}$ interact with the parameter $\theta$ solely through the dot product of $\theta$ and the sum of their sufficient statistics. When the parameter $\theta$ is used in this dot product unmodified (as in \eqref{eqn:genexpfam}), we call this a natural parameterization. Our analysis will be restricted to the families that satisfy the following two properties:

\begin{Defn}
An exponential family is minimal if the coordinates of the function $S$ are not almost surely linearly dependent, and the interior of $\Theta$ is non-empty.
\end{Defn}

\begin{Defn}
For any for $\Delta \in \mathbb{R}$, an exponential family is $\Delta$-bounded if
\begin{equation}
\Delta \ge \sup_{x,y\in \calX} ||S(x) - S(y)|| \mbox{.}
\end{equation}
This constraint can be relaxed with some caveats explored in the appendix.
\end{Defn}

When a family is minimal, the log-partition function $A$ has many interesting characteristics. It can be defined as $A(\theta) = \log \int_{\calX} h(x) \exp{S(x)\cdot \theta} dx$, and serves to normalize the distribution. Its derivatives form the cumulants of the distribution, that is to say $\nabla A(\theta) = \kappa_1 = \mathbb{E}_{x|\theta} [ S(x) ] $ and $\nabla^2 A(\theta) = \kappa_2 = \mathbb{E}_{x|\theta} [ (S(x) - \kappa_1)(S(x) - \kappa_1)^\intercal ]$. This second cumulant is also the covariance of $S(x)$, which demonstrates that $A(\theta)$ must be a convex function since covariances must be positive semidefinite.

In Bayesian data analysis, we are interested in finding our posterior distribution over the parameter $\theta$ that generated the data. We must introduce a prior distribution $p(\theta|\eta)$ to describe our initial beliefs on $\theta$, where $\eta$ is a parameterization of our family of priors.

\begin{align}
p(\theta|x_1,\ldots,x_n,\eta) &\propto  p(x_1,\ldots,x_n | \theta) p(\theta | \eta)  \\
&\propto (\prod_{i=1}^n h(x_i))\exp{ (\sum_{i=1}^n S(x_i)) \cdot \theta - n \cdot A(\theta)  } p(\theta | \eta) \\
&\propto \exp{ (\sum_{i=1}^n S(x_i), n) \cdot (\theta,-A(\theta))  } p(\theta | \eta) \\
\end{align}

Notice that we can ignore the $(\prod_{i=1}^n h(x_i))$ as it is a constant that will be normalized out. If we let our prior take the form of another exponential family $p(\theta | \eta) = \exp{T(\theta) \cdot \eta - B(\eta)}$ where $T(\theta) = (\theta,-A(\theta))$ and $B(\eta) = \log \int_\Theta \exp{T(\theta) \cdot \eta} d\theta$, the we can perform these manipulations,

\begin{align}
p(\theta|x_1,\ldots,x_n,\eta) &\propto \exp{ (\sum_{i=1}^n S(x_i), n) \cdot T(\theta)   + \eta \cdot T(\theta) - B(\eta) } \\
&\propto  \exp{ \bigg(\eta + (\sum_{i=1}^n S(x_i), n)\bigg) \cdot T(\theta)  - B(\eta) } \label{eqn:conjpriorwork}
\end{align}

and see that expression \eqref{eqn:conjpriorwork} can be written as

\begin{align}
p(\theta|\eta') = \exp{ T(\theta) \cdot \eta' - C(\eta' )}  \label{eqn:conjprior}
\end{align}

where $\eta' = \eta + \sum_{i=1}^n (S(x_i), 1)$  and $C(\eta')$ is chosen such that the distribution is normalized.

This family of posteriors is precisely the same exponential family that we chose for our prior. We call this a conjugate prior, and it offers us an efficient way of finding the parameter of our posterior: $\eta_{posterior} = \eta_{prior} + \sum_{i=1}^n (S(x_i), 1)$. Within this family, $T(\theta)$ forms the sufficient statistics of $\theta$, and the derivatives of $C(\eta)$ give the cumulants of these sufficient statistics.

\paragraph{Beta-Bernoulli System.} A specific example of an exponential family that we will be interested in is the Beta-Bernoulli system, where an individual's data is a single i.i.d. bit modeled as a Bernoulli variable with parameter $\rho$, along with a Beta conjugate prior.
 
\begin{equation}
p(x_1,\ldots,\x_n|\rho) = \prod_{i=1}^n \rho^{x_i} (1-\rho)^{1-x_i}
\end{equation}

Letting $\theta = \log(\frac{\rho}{1-\rho})$ and $A(\theta) = \log(1+\exp{\theta}) = -\log(1-\rho)$ , we can rewrite the equation as follows:

\begin{align}
p(x_1,\ldots,\x_n|\rho) &= \prod_{i=1}^n (\frac{\rho}{1-\rho})^{x_i} (1-\rho) \\
&= \exp{\sum_{i=1}^n x_i \log(\frac{\rho}{1-\rho}) + \log(1-\rho) } \\
&= \exp{ (\sum_{i=1}^n x_i) \cdot \theta - A(\theta)} \label{eqn:bernexpfam} \mbox{.}
\end{align}

This system satisfies the properties we require, as this natural parameterization with $\theta$ is both minimal and $\Delta$-bounded for $\Delta = 1$.

As our mechanisms are interested mainly in the posterior, the rest of this section will be written with respect the family specified by equation \eqref{eqn:conjprior}.

Now that we have the notation for our distributions, we can write out the expression for the Rényi divergence of two posterior distributions $P$ and $Q$ (parameterized by $\eta_P$ and $\eta_Q$) from the same exponential family. This expression allows us to directly compute the Rényi divergences of posterior sampling methods, and forms the crux of the analysis of our exponential family mechanisms. 

\begin{observation}
Let $P$ and $Q$ be two posterior distributions from the same exponential family that are parameterized by $\eta_P$ and $\eta_Q$. Then,

\begin{align}
D_\lambda(P||Q) &= \frac{1}{\lambda - 1} \log\Bigg( \int_\Theta P(\theta)^\lambda Q(\theta)^{1-\lambda}d\theta\Bigg) = \frac{C(\lambda \eta_P + (1-\lambda)\eta_Q)- \lambda C(\eta_P)}{\lambda - 1}+ C(\eta_Q) \label{eqn:exprenyi} \mbox{.}
\end{align}
\end{observation}

To help analyze the implication of equation \eqref{eqn:exprenyi} for Rényi Differential Privacy, we define some sets of prior/posterior parameters $\eta$ that arise in our analysis.

\begin{Defn}
We say a posterior parameter $\eta$ is normalizable if $C(\eta) = \log \int_\Theta \exp{T(\theta)\cdot\eta)} d\theta$ is finite.

Let $E$ denote the set of all normalizable $\eta$ for the conjugate prior family.
\end{Defn}

\begin{Defn}
Let $pset(\eta_0,n)$ be the convex hull of all parameters $\eta$ of the form $\eta_0 + n(S(x),1)$ for $x \in \calX$. When $n$ is an integer this represents the hull of possible posterior parameters after observing $n$ data points starting with the prior $\eta_0$.
\end{Defn}

\begin{Defn}
Let $Diff$ be the difference set for the family, where $Diff$ is the convex hull of all vectors of the form $( S(x) - S(y), 0)$ for $x,y \in \calX$.
\end{Defn}

\begin{Defn}
Two posterior parameters $\eta_1$ and $\eta_2$ are neighboring iff $\eta_1 -\eta_2 \in Diff$.\\
They are $r$-neighboring iff $(\eta_1 -\eta_2)/r \in Diff$.
\end{Defn}

\subsection{Mechanisms and Privacy Guarantees} \label{sec:expfammechs}

We begin with our simplest mechanism, Direct Sampling, which samples according to the true posterior. This mechanism is presented as Algorithm \ref{alg:expfamdirect}.

\begin{algorithm} 
\caption{Direct Posterior}\label{alg:expfamdirect}
\begin{algorithmic}[1]
\REQUIRE $\eta_0$, $\{x_1,\ldots,x_n\}$
\STATE Sample $\theta \sim p(\theta | \eta')$ where $\eta' = \eta_0 + \sum_{i=1}^n (S(x_i),1)$
\end{algorithmic}
\end{algorithm}

Even though Algorithm \ref{alg:expfamdirect} is generally not differentially private \cite{Dimitrakakis2014}, Theorem \ref{thm:privacyDirectExpFamily} suggests that it offers RDP  for $\Delta$-bounded exponential families and certain orders $\lambda$.

\begin{theorem} \label{thm:privacyDirectExpFamily}

For a $\Delta$-bounded minimal exponential family of distributions $p(x|\theta)$ with continuous log-partition function $A(\theta)$, there exists $\lambda^* \in (1,\infty]$ such Algorithm \ref{alg:expfamdirect} achieves $(\lambda,\epsilon(\eta_0,n,\lambda))$-RDP for $\lambda < \lambda^*$.

$\lambda^*$ is the supremum over all $\lambda$ such that all $\eta$ in the set $\eta_0 + (\lambda - 1)\Diff$ are normalizable.

\end{theorem}

\begin{Cor} \label{cor:bernlambdastar}
For the Beta-Bernoulli system with a prior $Beta(\alpha_0,\beta_0)$,  Algorithm \ref{alg:expfamdirect} achieves $(\lambda,\epsilon)$-RDP iff $\lambda > 1$ and $\lambda < 1 + min(\alpha_0, \beta_0)$.

\end{Cor}

Notice the implication of Corollary \ref{cor:bernlambdastar}: for any $\eta_0$ and $n>0$, there exists finite $\lambda$ such that direct posterior sampling does not guarantee $(\lambda,\epsilon)$-RDP for any finite $\epsilon$. This also prevents $(\epsilon,0)$-DP as an achievable goal as well. Algorithm \ref{alg:expfamdirect} is inflexible; it offers us no way to change the privacy guarantee. 

This motivates us to propose two different modifications to Algorithm \ref{alg:expfamdirect} that are capable of achieving arbitrary privacy parameters. Algorithm \ref{alg:expfamdiffuse} modifies the contribution of the data $\X$ to the posterior by introducing a coefficient $r$, while Algorithm \ref{alg:expfamconc} modifies the contribution of the prior $\eta_0$ by introducing a coefficient $m$. These simple ideas have shown up before in variations: \citep{wang2015privacy} introduces a temperature scaling that acts similarly to $r$, while \citep{without_sensitivity, Dimitrakakis2014} analyze concentration constraints for prior distributions much like our coefficient $m$.

\begin{algorithm} 
\caption{Diffused Posterior}\label{alg:expfamdiffuse}
\begin{algorithmic}[1]
\REQUIRE $\eta_0$, $\{x_1,\ldots,x_n\}, \epsilon, \lambda$
\STATE Find $r \in (0,1]$ such that $\forall r\text{-neighboring } \eta_P, \eta_Q \in pset(\eta_0,rn)$, $D_\lambda(p(\theta| \eta_P) || p(\theta|\eta_Q)) \le \epsilon$
\STATE Sample $\theta \sim p(\theta | \eta')$ where $\eta' = \eta_0  + r \sum_{i=1}^n (S(x_i),1)$
\end{algorithmic}
\end{algorithm}

\begin{theorem} \label{thm:expdiffuseprivacy}

For any $\Delta$-bounded minimal exponential family with prior $\eta_0$ in the interior of $E$, any $\lambda > 1$, and any $\epsilon > 0$, there exists $r^* \in (0,1]$ such that using $r \in (0,r^*]$ in Algorithm \ref{alg:expfamdiffuse} will achieve $(\lambda,\epsilon)$-RDP.

\end{theorem}

\begin{algorithm} 
\caption{Concentrated Posterior}\label{alg:expfamconc}
\begin{algorithmic}[1]
\REQUIRE $\eta_0$, $\{x_1,\ldots,x_n\}, \epsilon, \lambda$
\STATE Find $m \in (0,1]$ such that $\forall \text{ neighboring } \eta_P, \eta_Q \in pset(\eta_0/m,n)$, $D_\lambda(p(\theta| \eta_P) || p(\theta|\eta_Q)) \le \epsilon$
\STATE Sample $\theta \sim p(\theta | \eta')$  where $\eta' = \eta_0/m  +  \sum_{i=1}^n (S(x_i),1)$
\end{algorithmic}
\end{algorithm}

\begin{theorem}  \label{thm:expconcprivacy}

For any $\Delta$-bounded minimal exponential family with prior $\eta_0$ in the interior of $E$, any $\lambda > 1$, and any $\epsilon > 0$, there exists $m^* \in (0,1]$ such that using $m \in (0,m^*]$ in Algorithm \ref{alg:expfamconc} will achieve $(\lambda,\epsilon)$-RDP.

\end{theorem}

Theorems \ref{thm:expdiffuseprivacy} and \ref{thm:expconcprivacy} can be interpreted as demonstrating that any RDP privacy level can be achieved by setting $r$ or $m$ arbitrarily close to zero. A small $r$ implies a weak contribution from the data, while a small $m$ implies a strong prior that outweighs the contribution from the data. Setting $r = 1$ and $m = 1$ reduces to Algorithm \ref{alg:expfamdirect}, in which a sample is released from the true posterior without any modifications for privacy.

We have not yet specified how to find the appropriate values of $r$ or $m$, and the condition requires checking the supremum of divergences across the possible $pset$ range of parameters arising as posteriors. However, with an additional assumption this supremum of divergences can be efficiently computed.

\begin{theorem} \label{thm:divconvexity}

Let $e(\eta_P,\eta_Q,\lambda) =  D_\lambda\left(p(\theta|\eta_P) || p(\theta|\eta_Q)\right)$. For a fixed $\lambda$ and fixed $\eta_P$, the function $e$ is a convex function over $\eta_Q$.

If for any direction $v \in Diff$, the function $g_v(\eta) = v^\intercal \nabla^2C(\eta) v$ is convex over $\eta$, then for a fixed $\lambda$, the function $f_\lambda(\eta_P) = \sup_{\eta_Q r-\text{neighboring } \eta_P} e(\eta_P,\eta_Q,\lambda)$ is convex over $\eta_P$ in the directions spanned by $\Diff$.

\end{theorem}

\begin{Cor} \label{cor:bernboundary}
The Beta-Bernoulli system satisfies the conditions of Theorem \ref{thm:divconvexity} since the functions $g_v(\eta)$ have the form $(v^{(1)})^2(\psi_1(\eta^{(1)})+\psi_1(\eta^{(2)} - \eta^{(1)}))$, and  $\psi_1$ is the digamma function. Both $pset$ and $\Diff$ are defined as convex sets. The expression $\sup_{r-\text{neighboring } \eta_P,\eta_Q \in pset(\eta_0,n)} D_\lambda(p(\theta|\eta_P)||p(\theta|\eta_Q))$ is therefore equivalent to the maximum of  $D_\lambda(p(\theta|\eta_P)||p(\theta|\eta_Q))$ where $\eta_P \in \eta_0 + \{ (0,n), (n,n)\}$ and $\eta_Q \in \eta_P \pm (r, 0)$.

The higher dimensional Dirichlet-Categorical system also satsifies the conditions of Theorem \ref{thm:divconvexity}. This result is located in the appendix. 
\end{Cor}

We can do a binary search over $(0,1]$ to find an appropriate value of $r$ or $m$. At each candidate value, we only need to consider the boundary situations to evaluate whether this value achieves the desired RDP privacy level. These boundary situations depend on the choice of model, and not the data size $n$. For example, in the Beta-Bernoulli system, evaluating the supremum involves calculating the R\'{e}nyi diverengence across at most 4 pairs of distributions, as in Corollary \ref{cor:bernboundary}. In the $d$ dimensional Dirichlet-Categorical setting, there are $O(d^3)$ distribution pairs to evaluate.

Eventually, the search process is guaranteed to find a non-zero choice for $r$ or $m$ that achieves the desired privacy level, although the utility optimality of this choice is not guaranteed. If stopped early and none of the tested candidate values satisfy the privacy constraint, the analyst can either continue to iterate or decide not to release anything.


\subsection{Extension: Public Data for Exponential Families}

The use of a conjugate prior makes the interaction of observed data versus the prior easy to see. The prior $\eta_0$ can be expressed as $(\alpha\chi,\alpha)$, where $\chi$ is a vector expressing the average sufficient statistics of pseudo-observations and $\alpha$ represents a count of these pseudo-observations. After witnessing the $n$ data points, the posterior becomes a prior that has averaged the data sufficient statistics into a new $\chi$ and added $n$ to $\alpha$.

If the data analyst had some data in addition to $\X$ that was not privacy sensitive, perhaps from a stale data set for which privacy requirements have lapsed, then this data can be used to form a better prior for the analysis.

Not only would this improve utility by adding information that can be fully exploited, it would also in most cases improve the privacy guarantees as well. A stronger prior, especially a prior farther from the boundaries where $C(\eta)$ becomes infinite, will lead to smaller Rényi divergences. This is effectively the same behavior as the Concentrated Sampling mechanism, which scales the prior to imagine more pseudo-observations had been seen. This also could apply to settings in which the analyst can adaptively pay to receive non-private data, since this method will inform us once our prior formed from this data becomes strong enough to sample directly at our desired RDP level.

This also carries another privacy implication for partial data breaches. If an adversary learns the data of some individuals in the data set, the Direct Sampling mechanism's privacy guarantee for the remaining individuals can actually improve. Any contributions of the affected individuals to the posterior become in effect yet more public data placed in the prior. The privacy analysis and subsequent guarantees will match the setting in which this strengthened prior was used.

\subsection{Extension: Releasing the result of a Statistical Query}

Here we are given a sensitive database $\X = \{ x_1, \ldots, x_n \}$ and a predicate $\phi(\cdot)$ which maps each $x_i$ into the interval $ [ 0, 1 ]$. Our goal is to release a Rényi DP approximation to the quantity: $F(\X) = \frac{1}{n} \sum_{i=1}^{n} \phi(x_i)$.

Observe that directly releasing $F(\X)$ is neither DP nor Rényi DP, since this is a deterministic algorithm; our goal is to release a random sample from a suitable distribution so that the output is as close to $F(\X)$ as possible.

The task of releasing a privatized result of a statistical query can be embedded into our Beta-Bernoulli system. This allows the privatized statistical query release to be done using either Algorithm \ref{alg:expfamdiffuse} or Algorithm \ref{alg:expfamconc}.

We can extend the Beta-Bernoulli model to allow the sufficient statistics $S(x)$ to range over the interval $[0,1]$ instead of just the discrete set $\{0,1\}$. This alteration still results in a $\Delta$-bounded exponential family, and the privacy results hold.

The sampled posterior will be a Beta distribution that will concentrate around the mean of the data observations and the pseudo-observations of the prior. The process is described in the Beta-Sampled Statistical Query algorithm. The final transformation maps the natural parameter $\theta \in (-\infty,\infty)$ onto the mean of the distribution $\rho \in (0,1)$.

\begin{algorithm*} \label{alg:statquery}
\caption{Beta-Sampled Statistical Query}\label{alg:betastatquery}
\begin{algorithmic}[1]
\REQUIRE $\eta_0$, $\{x_1,\ldots,x_n\},f, \epsilon, \lambda$
\STATE Compute $\X_f = \{f(x_1), \ldots, f(x_n)\}$.
\STATE Sample $\theta$ via Algorithm \ref{alg:expfamdiffuse} or Algorithm \ref{alg:expfamconc} applied to $\X_f$ with $\eta_0$, $\epsilon$, and $\lambda$.
\STATE Release $\rho = \frac{\exp{\theta}}{1+\exp{\theta}}$.
\end{algorithmic}
\end{algorithm*}

\section{\rdp\ for Generalized Linear Models with Gaussian Prior}
\label{sec:glm}
In this section, we reinterpret some existing algorithms in~\citep{without_sensitivity} in the light of \rdp, and use ideas from~\citep{without_sensitivity} to provide new \rdp\ algorithms for posterior sampling for a subset of generalized linear models with Gaussian priors. 


\subsection{Background: Generalized Linear Models (GLMs)}
The goal of generalized linear models (GLMs) is to predict an outcome $y$ given an input vector $x$; $y$ is assumed to be generated from a distribution in the exponential family whose mean depends on $x$ through 
$\Expect{}{y | x} = g^{-1}(w^\top x),$ where $w$ represents the weight of linear combination of $x$, and $g$ is called the link function. For example, in logistic regression, the link function $g$ is logit and $g^{-1}$ is the sigmoid function; and in linear regression, the link functions is the identity function. Learning in GLMs means learning the actual linear combination $w$.


Specifically, the likelihood of $y$ given $x$ can be written as
$p(y|w,x) = h(y) \exp{{y w^\top x - A(w^\top x)}}$,
where $x\in\calX$, $y\in\calY$, $A$ is the log-partition function, and $h(y)$ the scaling constant.
%
%
%
%
%
Given a dataset $D = \{(x_1,y_1),\dots,(x_n,y_n)\}$ of $n$ examples with $x_i \in \calX$ and $y_i \in \calY$, our goal is to learn the parameter $w$. Let $p(D|w)$ denote $p(\{y_1,\dots,y_n\}|w,\{x_1,\dots,x_n\}) = \prod_{i=1}^n p(y_i|w,x_i)$. We set the prior $p(w)$ as a multivariate Gaussian distribution with covariance $\Sigma = (n\reg)^{-1} I$, i.e., $p(w) \sim \mathcal{N}(0, (n\reg)^{-1} I)$. 
The posterior distribution of $w$ given $D$ can be written as
\begin{align}\label{eqn:logistic def}
p(w | D) = \frac{p(D | w) p(w)}{\int_{\mathbb{R}^d} p(D | w') p(w') dw'}
\propto 
\exp{-\frac{n\reg \|w\|^2}{2}} 
\prod_{i=1}^n p(y_i|w,x_i).
\end{align}


\subsection{Mechanisms and Privacy Guarantees}

First, we introduce some assumptions that characterize the subset of GLMs and the corresponding training data on which \rdp\ can be guaranteed.
\begin{assumption}\label{assump:LR_privacy}
\begin{enumerate}
\item $\calX$ is a bounded domain such that $\|x\|_2\leq c$ for all $x\in\calX$, and $x_i\in\calX$ for all $(x_i,y_i)\in D$.
\item $\calY$ is a bounded domain such that $\calY \subseteq [y_{\min}, y_{\max}]$, and $y_i\in\calY$ for all $(x_i,y_i)\in D$..
\item $g^{-1}$ has bounded range such that $g^{-1} \in [\gamma_{\min}, \gamma_{\max}]$.
\end{enumerate}
Then, let $B = \max \{|y_{\min} - \gamma_{\max}|, |y_{\max} - \gamma_{\min}|\}$.
\end{assumption}

\paragraph{Example: Binary Regression with Bounded $\calX$}
Binary regression is used in the case where $y$ takes value $\calY = \{0, 1\}$. There are three common types of binary regression, 
logistic regression with $g^{-1}(w^\top x) = {1}/({1+\exp{-w^\top x}})$, 
probit regression with $g^{-1}(w^\top x) = \Phi(w^\top x)$ where $\Phi$ is the Gaussian cdf, 
and complementary log-log regression with $g^{-1}(w^\top x) = 1 - \exp{-\exp{w^\top x}}$.
In these three cases, $\calY = \{0,1\}$, $g^{-1}$ has range $(0, 1)$ and thus $B = 1$.
Moreover, it is often assumed for binary regression that any example lies in a bounded domain, i.e., $\|x\|_2\leq c$ for $x\in \calX$.


Now we establish the privacy guarantee for sampling directly from the posterior in~\eqref{eqn:logistic def} in Theorem~\ref{lem:LR_privacy}. We also show that this privacy bound is tight for logistic regression; a detailed analysis is in Appendix.
\begin{theorem}\label{lem:LR_privacy0}
Suppose we are given a GLM and a dataset $D$ of size $n$ that satisfies Assumption~\ref{assump:LR_privacy},
and a Gaussian prior with covariance $\Sigma = (n\reg)^{-1} I$, then
sampling with posterior in \eqref{eqn:logistic def} satisfies $(\lambda, \frac{2 c^2 B^2}{n\reg}\lambda)$-\rdp\ for all $\lambda \geq 1$.
\end{theorem}


Notice that direct posterior sampling cannot achieve $(\lambda,\epsilon)$-\rdp\ for arbitrary $\lambda$ and $\epsilon$. We next present Algorithm~\ref{alg:lr_concentrate} and \ref{alg:lr_diffuse}, as analogous to Algorithm~\ref{alg:expfamconc} and \ref{alg:expfamdiffuse} for exponential family respectively, that guarantee any given \rdp\ requirement. Algorithm~\ref{alg:lr_concentrate} achieves a given \rdp\ level by setting a stronger prior, while Algorithm~\ref{alg:lr_diffuse} by raising the temperature of the likelihood.

%
%

\begin{algorithm}[H]
\caption{Concentrated Posterior}\label{alg:lr_concentrate}
\begin{algorithmic}[1]
\REQUIRE 
Dataset $D$ of size $n$;
Gaussian prior with covariance $(n\reg_0)^{-1} I$; 
$(\lambda, \epsilon)$.
\STATE Set $\reg = \max\{\frac{2 c^2 B^2 \lambda}{n\epsilon}, \reg_0\}$ in \eqref{eqn:logistic def}.\\
\STATE Sample $w\sim p(w|D)$ in \eqref{eqn:logistic def}.
\end{algorithmic}
\end{algorithm}

\begin{algorithm}[H]
\caption{Diffuse Posterior}\label{alg:lr_diffuse}
\begin{algorithmic}[1]
\REQUIRE 
Dataset $D$ of size $n$;
Gaussian prior with covariance $(n\reg)^{-1} I$;
$(\lambda, \epsilon)$.
\STATE Replace $p(y_i|w,x_i)$ with $p(y_i|w,x_i)^\rho$ in \eqref{eqn:logistic def} where $\rho = \min\{1, \sqrt{\frac{\epsilon n \reg}{2c^2 B^2 \lambda}}\}$.
\STATE Sample $w\sim p(w|D)$ in \eqref{eqn:logistic def}.
\end{algorithmic}
\end{algorithm}


It follows directly from Theorem~\ref{lem:LR_privacy} that under Assumption~\ref{assump:LR_privacy}, Algorithm~\ref{alg:lr_concentrate} satisfies $(\lambda,\epsilon)$-\rdp.


\begin{theorem}\label{lem:LR_privacy}
Suppose we are given a GLM and a dataset $D$ of size $n$ that satisfies Assumption~\ref{assump:LR_privacy},
and a Gaussian prior with covariance $\Sigma = (n\reg)^{-1} I$, then
Algorithm~\ref{alg:lr_diffuse} guarantees $(\lambda, \epsilon)$-\rdp.
In fact, it guarantees $(\tilde{\lambda}, \frac{\epsilon}{\lambda}\tilde{\lambda})$-\rdp\ for any $\tilde{\lambda} \geq 1$.
\end{theorem}

We show that the \rdp\ guarantee in Theorem~\ref{lem:LR_privacy0} 
is tight for logistic regression.
\begin{theorem}\label{lem:LR_tightness}
For any $d > 1$ and any $n \geq 1$, there exists neighboring datasets $D$ and $D'$, each of size $n$, there exists $\lambda_0$, such that for any $\lambda > \lambda_0$, $\lambda$-\rd\ between logistic regression posteriors under $D$ and $D'$ with Gaussian prior is larger than $\frac{c^2 }{2n\reg} (\lambda-1)$.
\end{theorem}
This implies the tightness of $(\lambda, \frac{2 c^2}{n\reg}\lambda)$, the RDP guarantee of posterior sampling for logistic regression posterior.


\section{Experiments}

In this section, we present the experimental results for our proposed algorithms for both exponential family and GLMs. Our experimental design focuses on two goals -- first, analyzing the relationship between $\lambda$ and $\epsilon$ in our privacy guarantees and second, exploring the privacy-utility trade-off of our proposed methods in relation to existing methods.

\subsection{Synthetic Data: Beta-Bernoulli Sampling Experiments}

In this section, we consider posterior sampling in the Beta-Bernoulli system. We compare three algorithms. As a baseline, we select a modified version of the algorithm in~\citep{foulds2016theory}, which privatizes the sufficient statistic of the data to create a privatized posterior. Instead of Laplace noise that is used by\cite{foulds2016theory}, we use Gaussian noise to do the privatization; \cite{mironov2017renyi} shows that if Gaussian noise with variance $\sigma^2$ is added, then this offers an RDP guarantee of $(\lambda, \lambda \frac{\Delta^2}{\sigma^2})$ for $\Delta$-bounded exponential families. We also consider the two algorithms presented in Section~\ref{sec:expfammechs} -- Algorithm~\ref{alg:expfamdiffuse} and~\ref{alg:expfamconc}; observe that Algorithm~\ref{alg:expfamdirect} is a special case of both. 500 iterations of binary search were used to select $r$ and $m$ when needed.

\begin{figure*}[!t]
\centering
\begin{subfigure}[b]{0.49\textwidth}
\includegraphics[width=\textwidth]{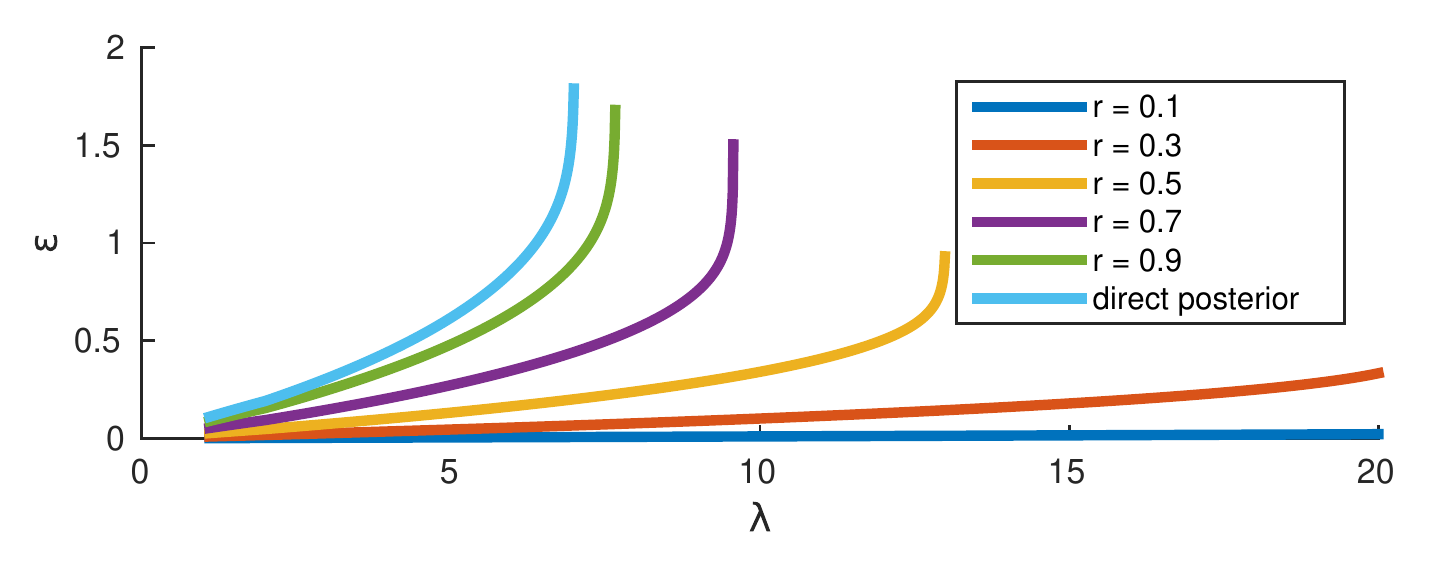}
\caption{Algorithm \ref{alg:expfamdiffuse}}
\end{subfigure}
\begin{subfigure}[b]{0.49\textwidth}
\includegraphics[width=\textwidth]{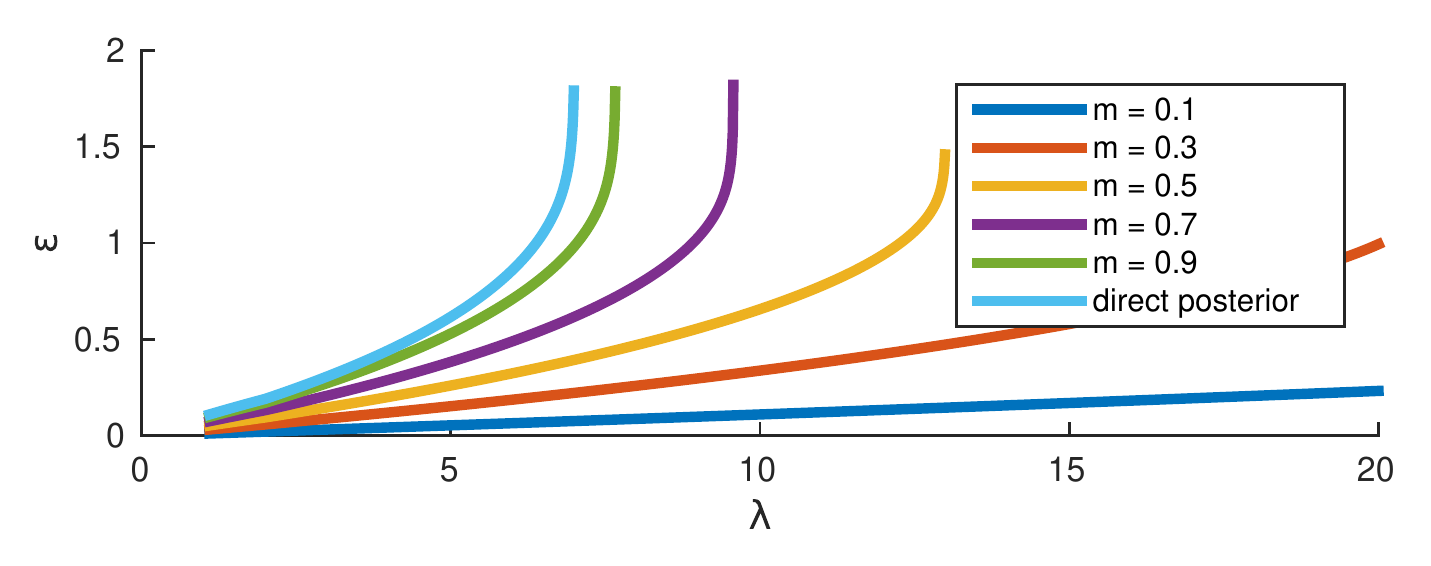}
\caption{Algorithm \ref{alg:expfamconc}}
\end{subfigure}
\caption{Illustration of Potential $(\lambda,\epsilon)$-RDP Curves for Exponential Family Sampling.}
\label{fig:elcurve}
\end{figure*}

\paragraph{Achievable Privacy Levels.} We plot the $(\lambda,\epsilon)$-RDP parameters achieved by Algorithms \ref{alg:expfamdiffuse} and \ref{alg:expfamconc} for a few values of $r$ and $m$. These parameters are plotted for a prior $\eta_0 = (6,18)$ and the data size $n=100$ which are selected arbitrarily for illustrative purposes. We plot over six values $\{0.1,0.3,0.5,0.7,0.9,1\}$ of the scaling constants $r$ and $m$. The results are presented in Figure \ref{fig:elcurve}. Our primary observation is the presence of the vertical asymptotes for our proposed methods. Recall that any privacy level is achievable with our algorithms given small enough $r$ or $m$; these plots demonstrate the interaction of $\lambda$ and $\epsilon$. As $r$ and $m$ decrease, the $\epsilon$ guarantees improve at each $\lambda$ and even become finite at larger orders $\lambda$, but a vertical asymptote still exists. The results for the baseline are not plotted: it achieves RDP along any line of positive slope passing through the origin.

\begin{figure*}[!t]
\centering
\begin{subfigure}[b]{0.24\textwidth}
\includegraphics[width=\textwidth]{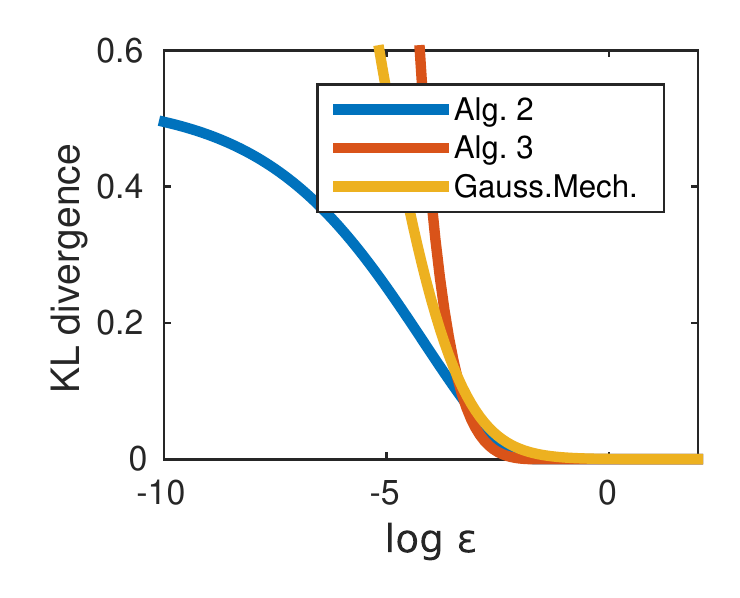}
\caption{KL: $\lambda = 2 < \lambda^*$}
\end{subfigure}
\begin{subfigure}[b]{0.24\textwidth}
\includegraphics[width=\textwidth]{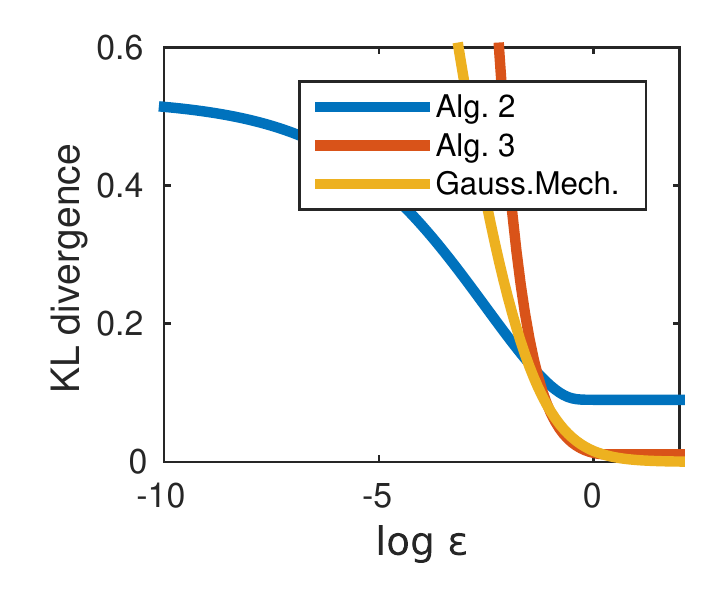}
\caption{KL: $\lambda = 15 > \lambda^*$}
\end{subfigure}
\begin{subfigure}[b]{0.24\textwidth}
\includegraphics[width=\textwidth]{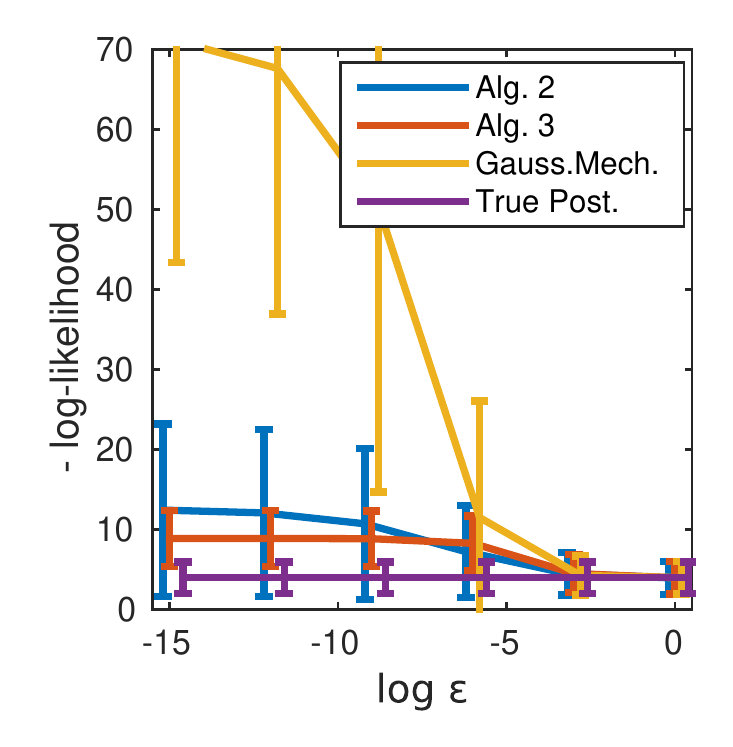}
\caption{$-\log p(\X_H)$: $\lambda = 2$}
\end{subfigure}
\begin{subfigure}[b]{0.24\textwidth}
\includegraphics[width=\textwidth]{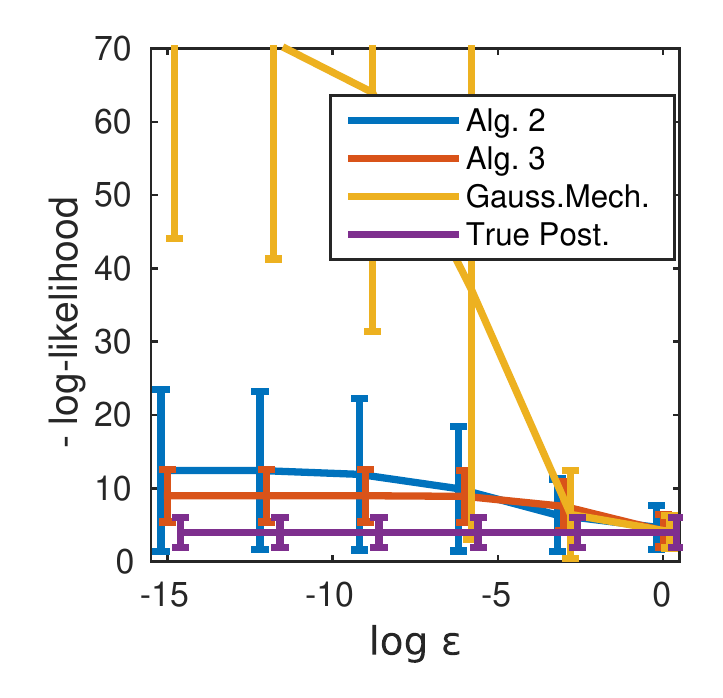}
\caption{$-\log p(\X_H)$: $\lambda = 15$}
\end{subfigure}
\caption{Exponential Family Synthetic Data Experiments.}
\label{fig:expfamexp}
\end{figure*}

\paragraph{Privacy-Utility Tradeoff.} We next evaluate the privacy-utility tradeoff of the algorithms by plotting $KL(P||\calA)$ as a function of $\epsilon$ with $\lambda$ fixed, where $P$ is the true posterior and $\calA$ is the output distribution of a mechanism. For Algorithms~\ref{alg:expfamdiffuse} and~\ref{alg:expfamconc}, the KL divergence can be evaluated in closed form. For the Gaussian mechanism, numerical integration was used to evaluate the KL divergence integral.  We have arbitrarily chosen $\eta_0 =(6,18)$ and data set $\X$ with 100 total trials and 38 successful trials. We have plotted the resulting divergences over a range of $\epsilon$ for $\lambda = 2$ in (a) and for $\lambda = 15$ in (b) of Figure \ref{fig:expfamexp}. When $\lambda = 2 < \lambda^*$, both Algorithms \ref{alg:expfamdiffuse} and \ref{alg:expfamconc} reach zero KL divergence once direct sampling is possible. The Gaussian mechanism must always add nonzero noise. As $\epsilon \rightarrow 0$, Algorithm \ref{alg:expfamconc} approaches a point mass distribution heavily penalized by the KL divergence. Due to its projection step, the Gaussian Mechanism follows a bimodal distribution as $\epsilon \rightarrow 0$. Algorithm \ref{alg:expfamdiffuse} degrades to the prior, with modest KL divergence. When $\lambda = 15 > \lambda^*$, the divergences for Algorithms \ref{alg:expfamdiffuse} and \ref{alg:expfamconc} are bounded away from 0, while the Gaussian mechanism still approaches the truth as $\epsilon \rightarrow \infty$. In a non-private setting, the KL divergence would be zero.

Finally, we plot $\log p(\X_H| \theta)$ as a function of $\epsilon$, where $\theta$ comes from one of the mechanisms applied to $\X$. Both $\X$ and $\X_H$ consist of 100 Bernoulli trials with proportion parameter $\rho=0.5$. This experiment was run 10000 times, and we report the mean and standard deviation. Similar to the previous section, we have a fixed prior of $\eta_0 = (6,18)$. The results are shown for $\lambda = 2$ in (c) and for $\lambda = 15$ in (d) of \ref{fig:expfamexp}. These results agree with the limit behaviors in the KL test. This experiment is more favorable for Algorithm \ref{alg:expfamconc}, as it degrades only to the log likelihood under the mode of the prior. In this plot, we have included sampling from the true posterior as a non-private baseline.

%

\subsection{Real Data: Bayesian Logistic Regression Experiments}

We now experiment with Bayesian logistic regression with Gaussian prior on three real datasets. We consider three algorithms -- Algorithm~\ref{alg:lr_concentrate} and \ref{alg:lr_diffuse}, as well as the OPS algorithm proposed in~\citep{wang2015privacy} as a sanity check. OPS achieves pure differential privacy when the posterior has bounded support; for this algorithm, we thus truncate the Gaussian prior to make its support the $L_2$ ball of radius $c/\reg$, which is the smallest data-independent ball guaranteed to contain the MAP classifier. 

\paragraph{Achievable Privacy Levels.} We consider the achievable \rdp\ guarantees for our algorithms and OPS under the same set of parameters $\reg$, $c$, $\rho$ and $B=1$. \citep{wang2015privacy} shows that with the truncated prior, OPS guarantees $\frac{4 c^2 \rho}{\reg}$-differential privacy, which implies $(\lambda, \frac{4 c^2 \rho}{\reg})$-\rdp\ for all $\lambda \in [1, \infty]$; whereas our algorithm guarantees $(\lambda, \frac{2c^2 \rho^2}{n\beta}\lambda)$-\rdp\ for all $\lambda \geq 1$. Therefore our algorithm achieves better \rdp\ guarantees at $\lambda \leq \frac{2n}{\rho}$, which is quite high in practice as $n$ is the dataset size.

%
%
%


\paragraph{Privacy-Utility: Test Log-Likelihood and Error.}

 %
We conduct Bayesian logistic regression on three real datasets: Abalone, Adult and MNIST. We perform binary classification tasks: abalones with less than $10$ rings vs. the rest for Abalone, digit $3$ vs. digit $8$ for MNIST, and income $\leq 50$K vs. $>50$K for Adult. We encode all categorical features with one-hot encoding, resulting in $9$ dimensions for Abalone, $100$ dimensions for Adult and $784$ dimensions in MNIST. We then scale each feature to range from $[-0.5, 0.5]$, and normalize each example to norm $1$. $1/3$ of the each dataset is used for testing, and the rest for training. Abalone has $2784$ training and $1393$ test samples, Adult has $32561$ and $16281$, and MNIST has $7988$ and $3994$ respectively.
%
%

\begin{figure*}
\centering
\begin{subfigure}[b]{0.325\textwidth}
\includegraphics[width=\textwidth]{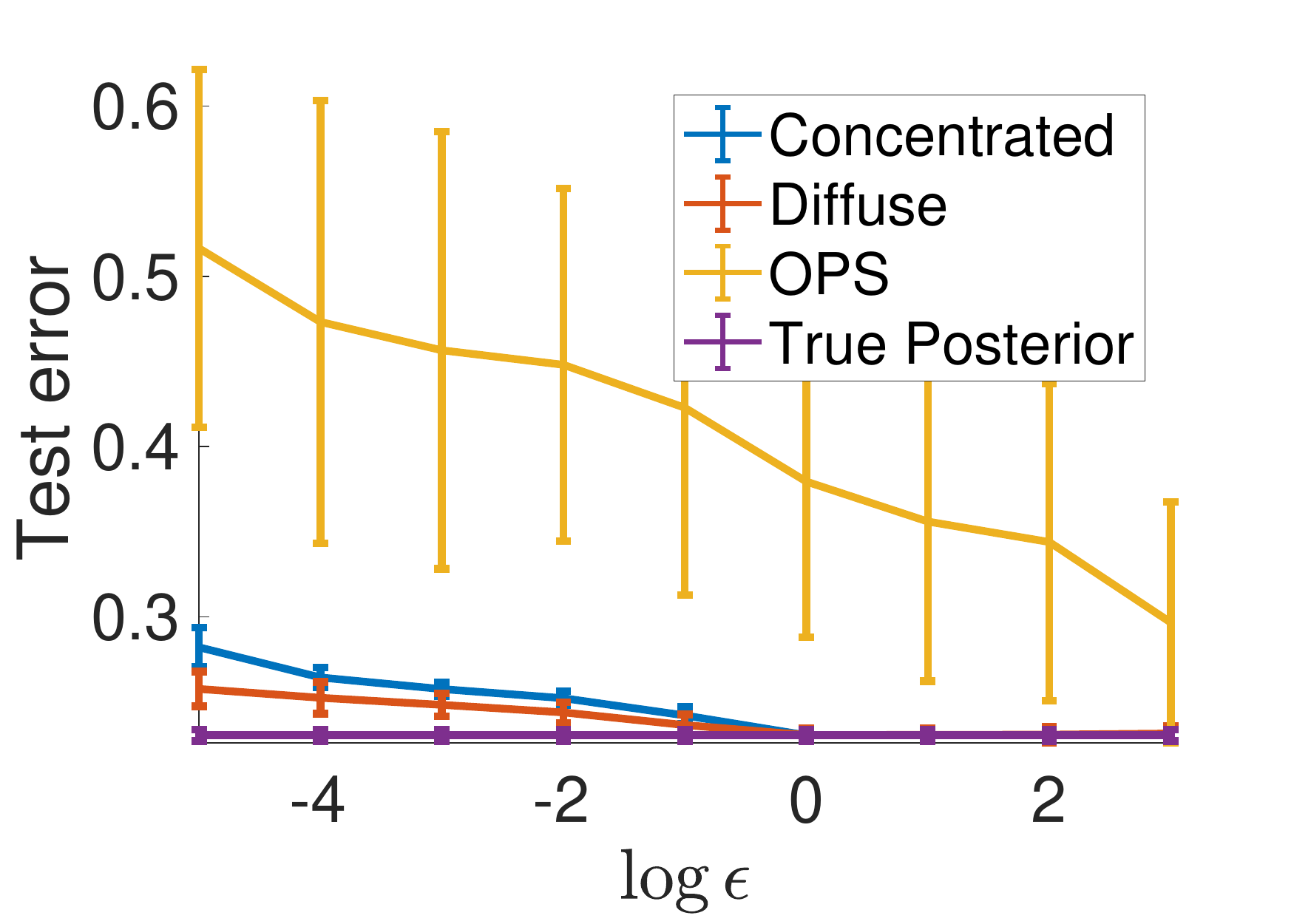}
\end{subfigure}
\hfill
\begin{subfigure}[b]{0.325\textwidth}
\includegraphics[width=\textwidth]{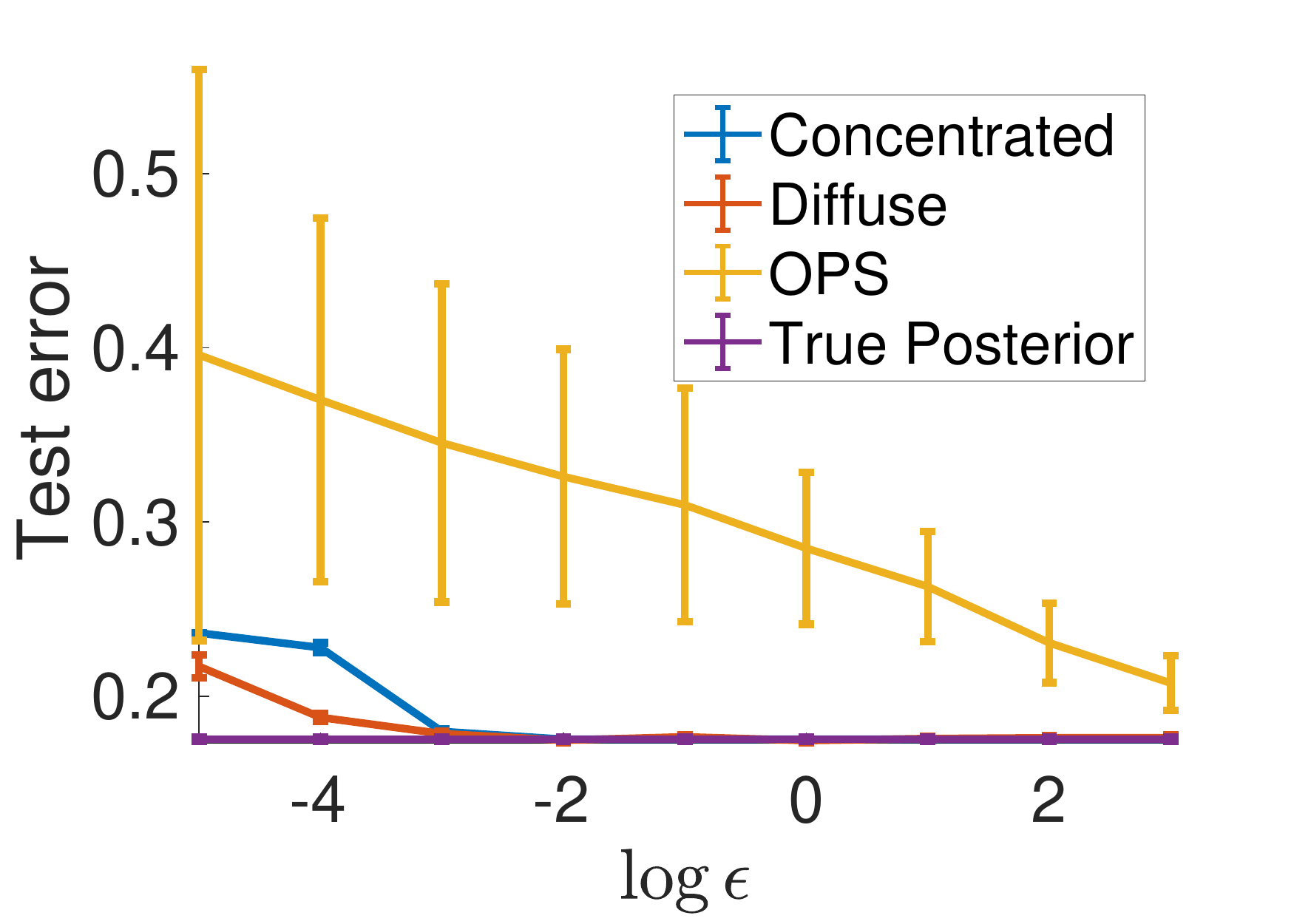}
\end{subfigure}
\hfill
\begin{subfigure}[b]{0.325\textwidth}
\includegraphics[width=\textwidth]{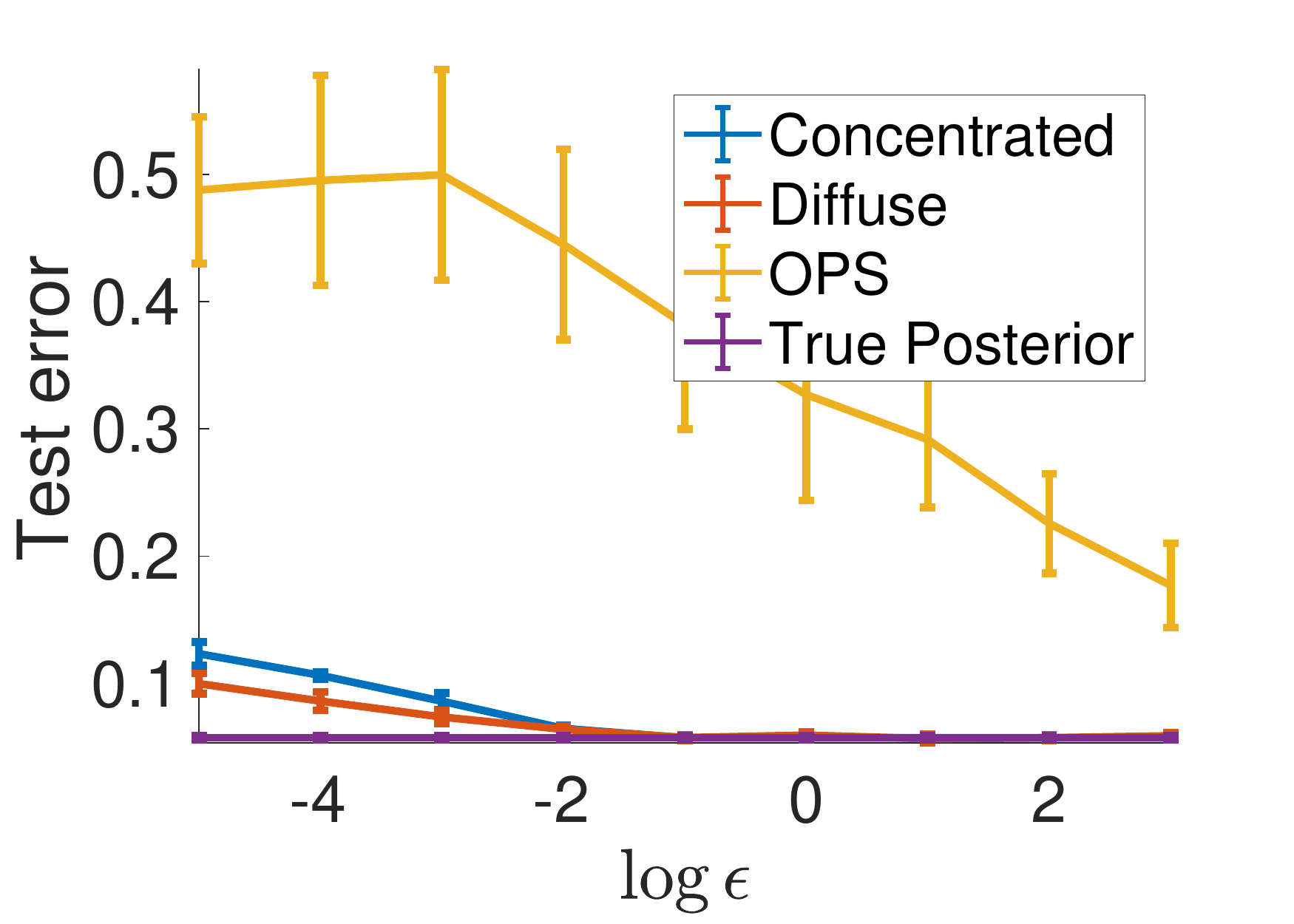}
\end{subfigure}
%
\begin{subfigure}[b]{0.325\textwidth}
\includegraphics[width=\textwidth]{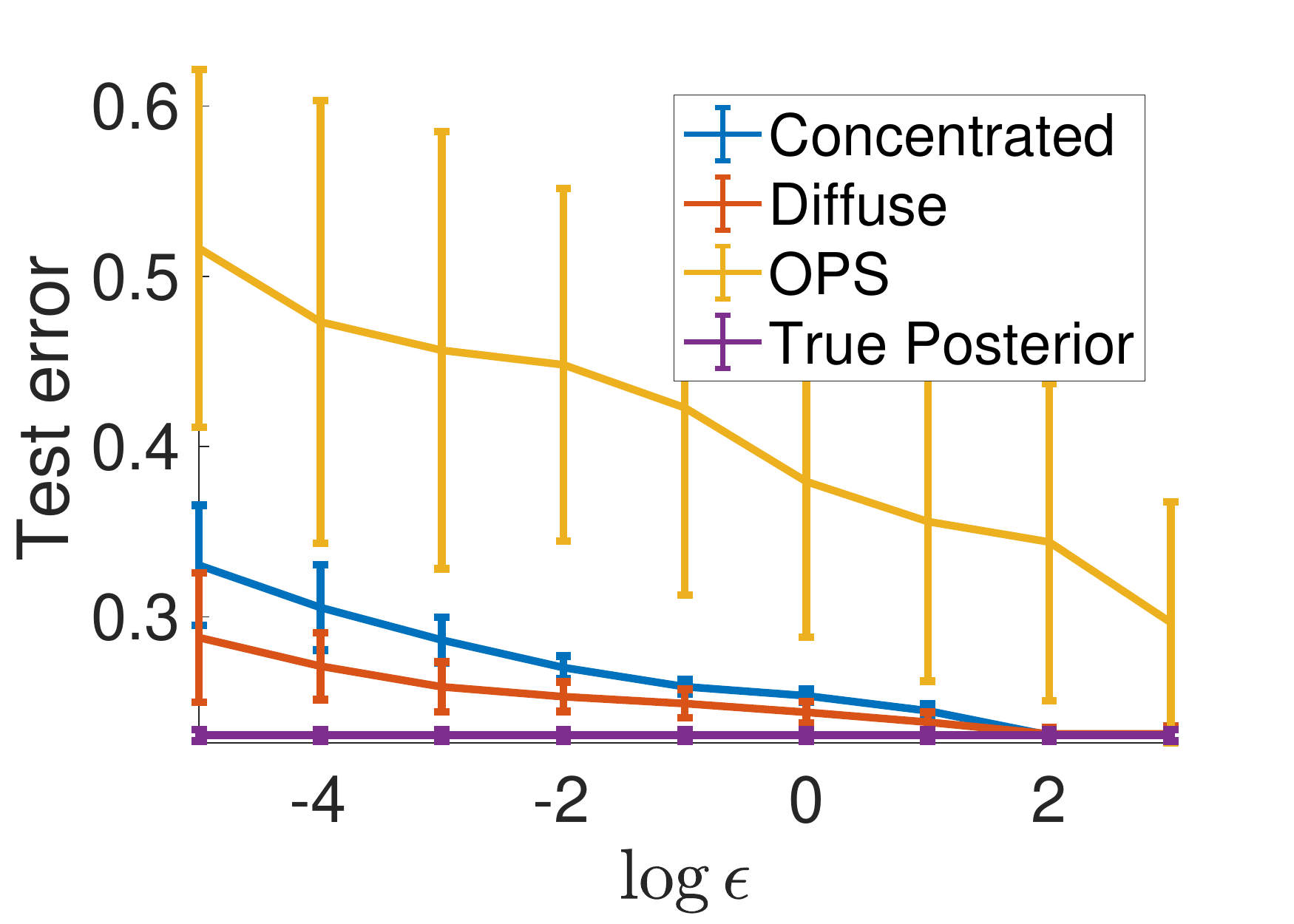}
\end{subfigure}
\hfill
\begin{subfigure}[b]{0.325\textwidth}
\includegraphics[width=\textwidth]{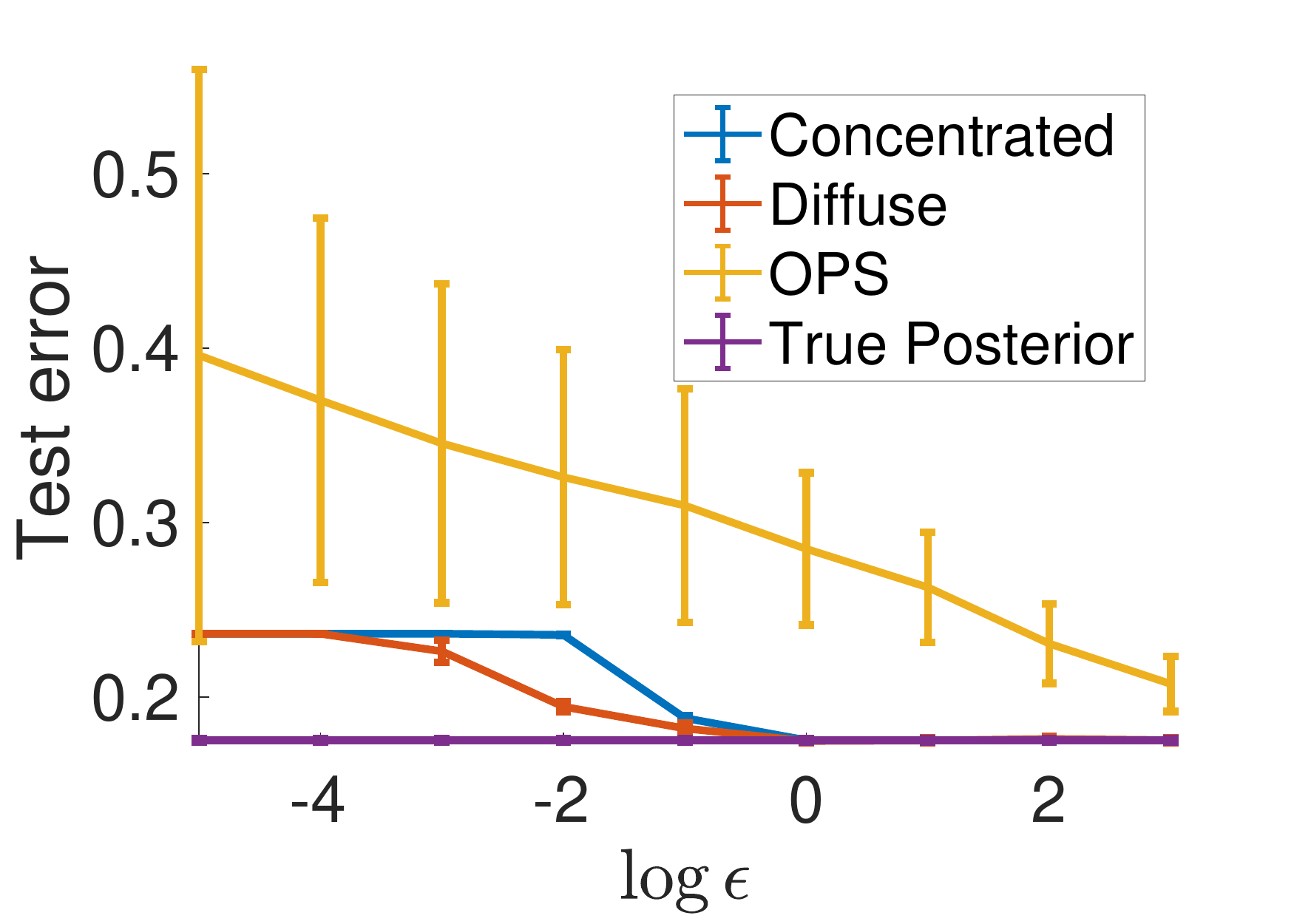}
\end{subfigure}
\hfill
\begin{subfigure}[b]{0.325\textwidth}
\includegraphics[width=\textwidth]{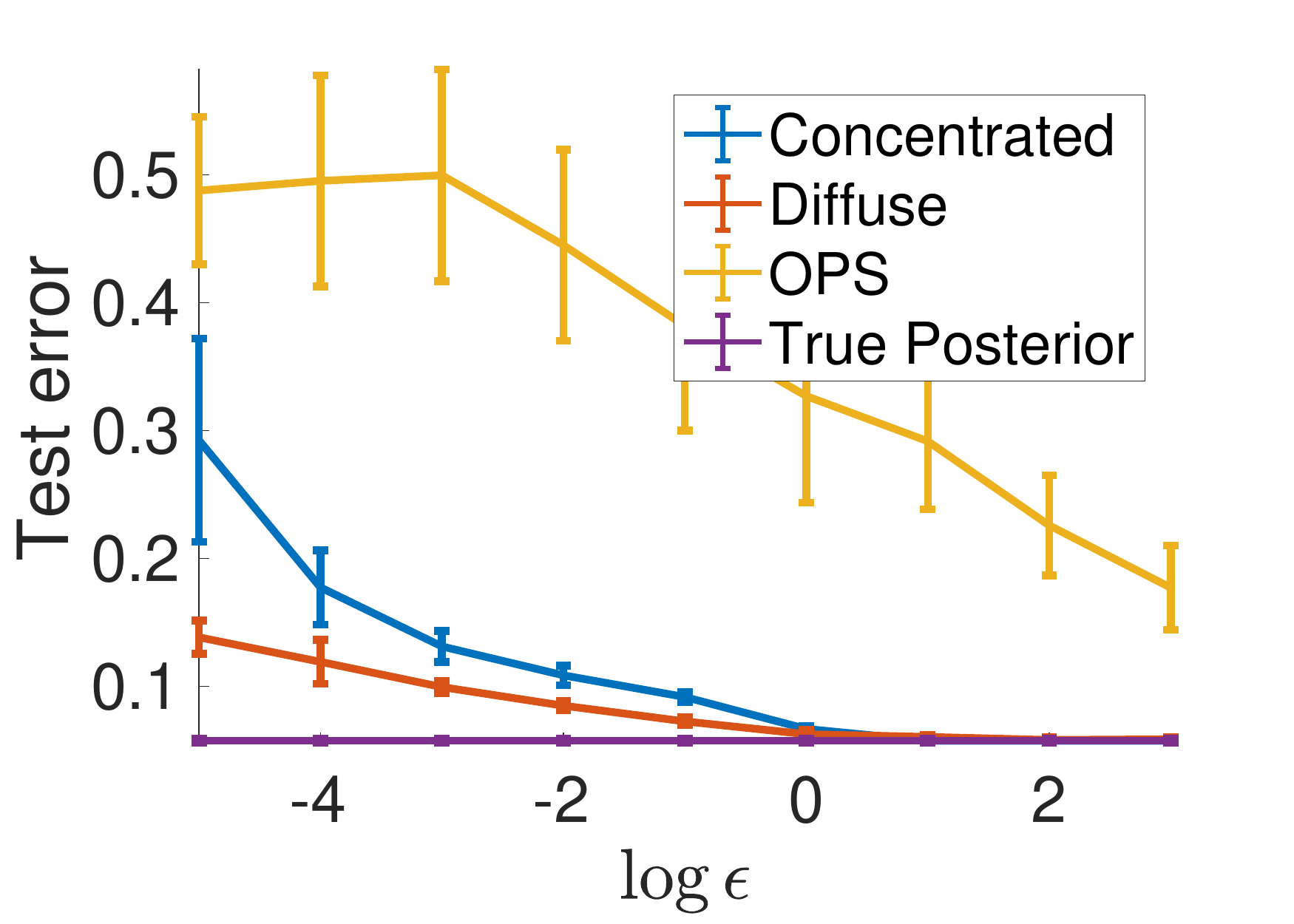}
\end{subfigure}
\begin{subfigure}[b]{0.325\textwidth}
\includegraphics[width=\textwidth]{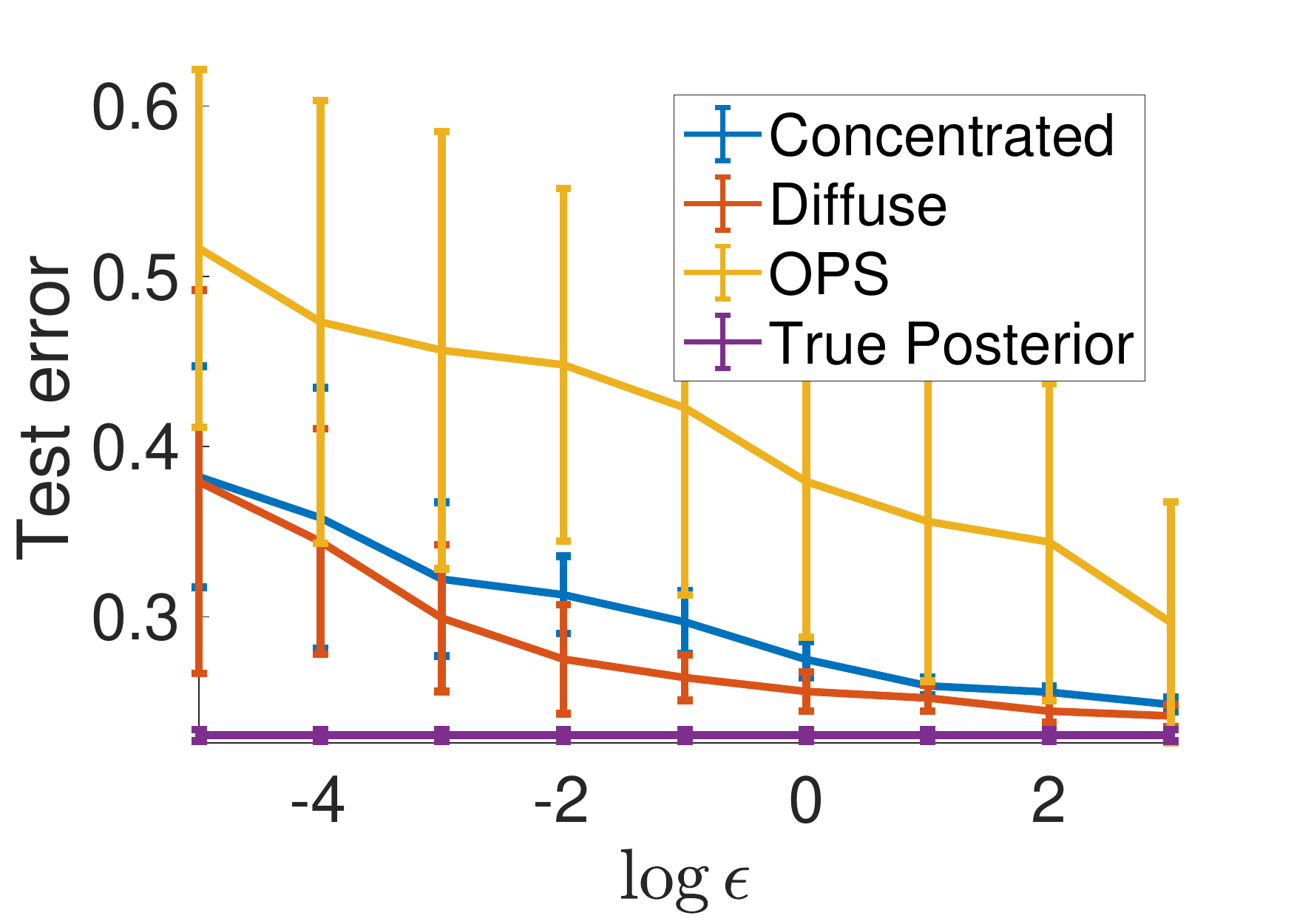}
\vspace{-15pt}\caption{Abalone.}
\end{subfigure}
\hfill
\begin{subfigure}[b]{0.325\textwidth}
\includegraphics[width=\textwidth]{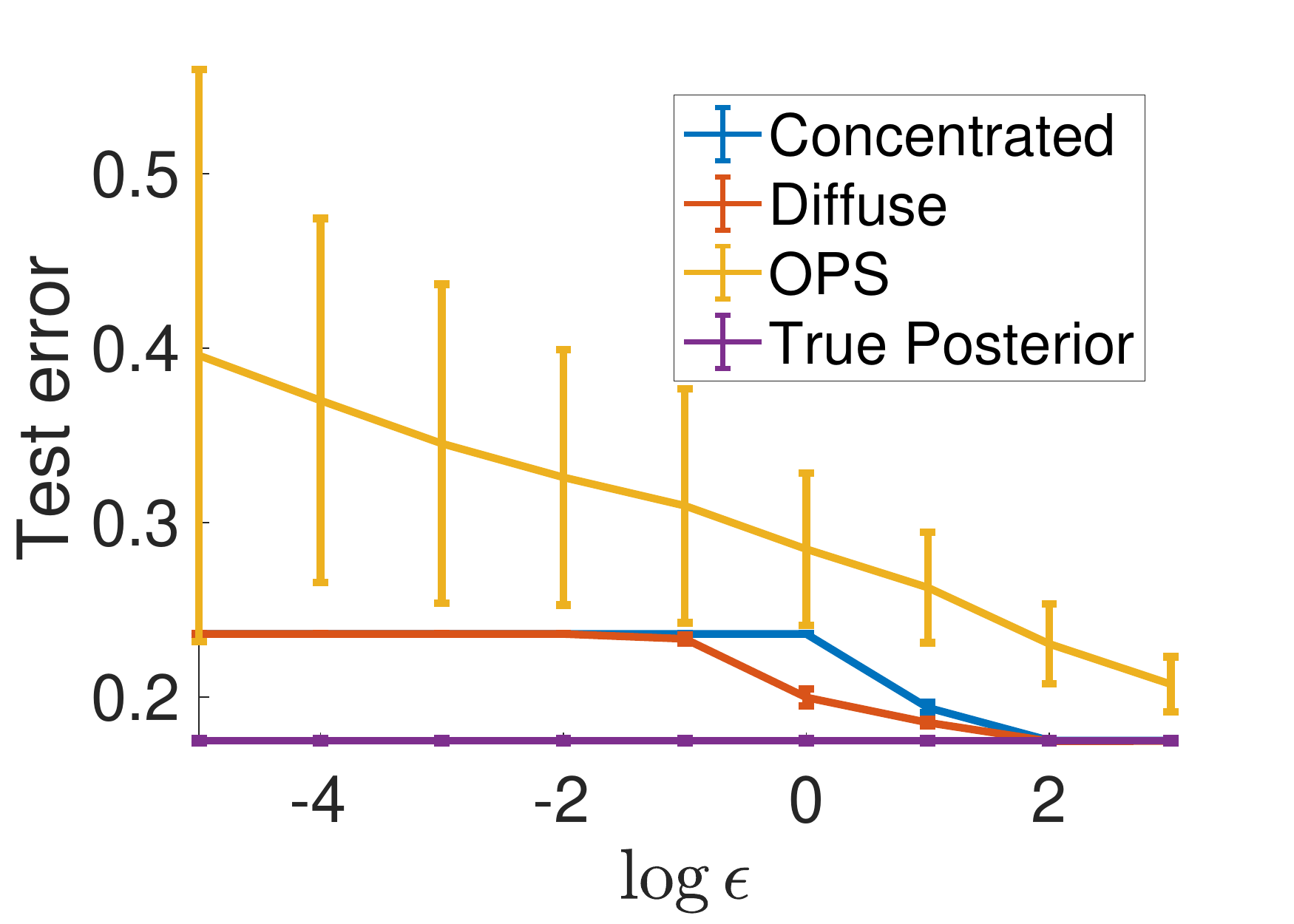}
\vspace{-15pt}\caption{Adult.}
\end{subfigure}
\hfill
\begin{subfigure}[b]{0.325\textwidth}
\includegraphics[width=\textwidth]{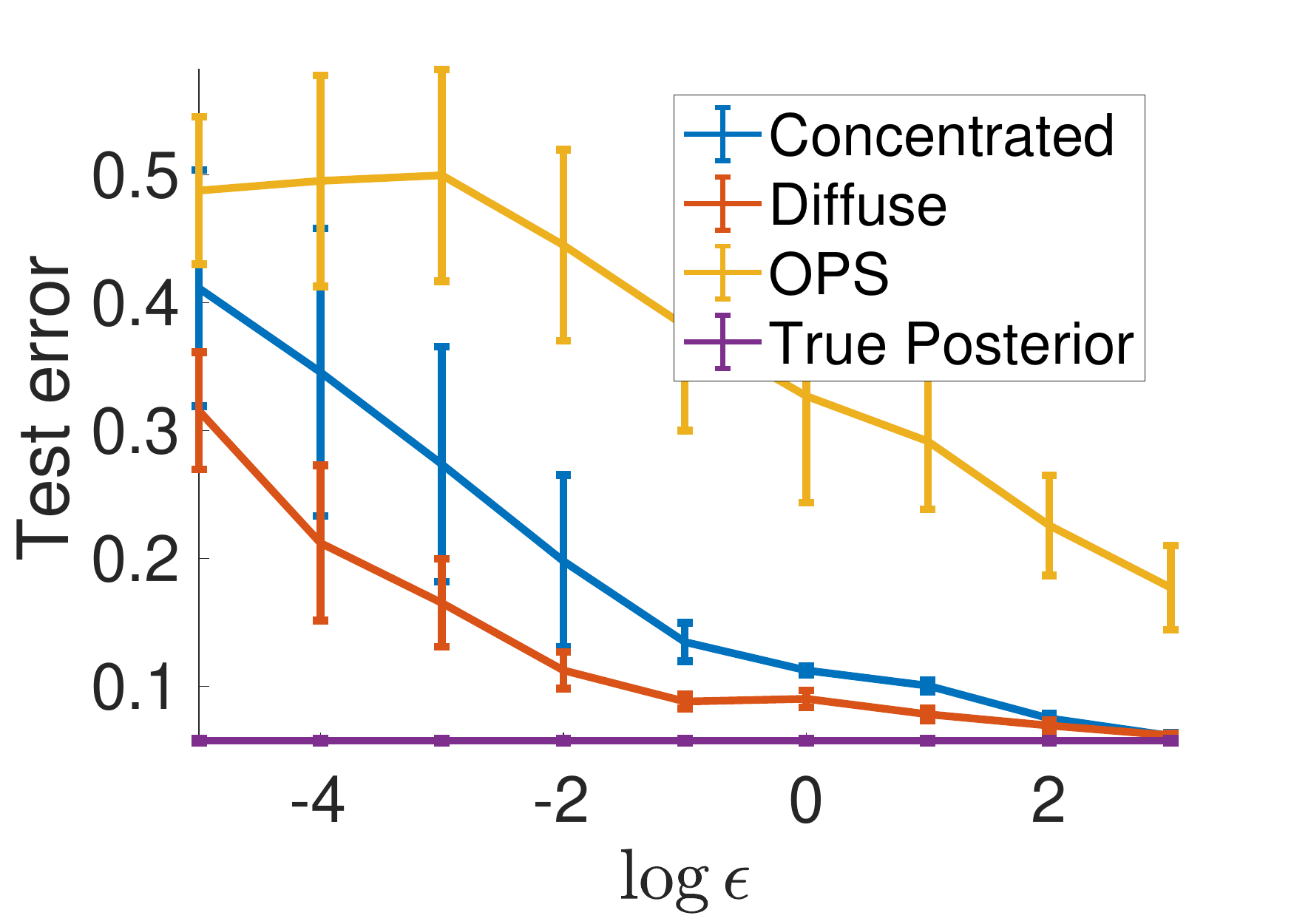}
\vspace{-15pt}\caption{MNIST 3vs8.}
\end{subfigure}
\caption{Test error vs. privacy parameter $\epsilon$. $\lambda=1$, $10$, $100$ from top to bottom.}
\label{fig:lr_test}
\end{figure*}

For all algorithms, we use an original Gaussian prior with $\reg = 10^{-3}$.
The posterior sampling is done using slice sampling with $1000$ burn-in samples. 
Notice that slice sampling does not give samples from the exact posterior. However, a number of MCMC methods are known to converge in total variational distance in time polynomial in the data dimension for log-concave posteriors (which is the case here)~\cite{vempala2005geometric}. Thus, provided that the burn-in period is long enough, we expect the induced distribution to be quite close, and we leave an exact \rdp\ analysis of the MCMC sampling as future work. 
%
For privacy parameters, we set $\lambda = 1, 10, 100$ and $\epsilon \in \{e^{-5}, e^{-4}, \dots, e^{3}\}$.  Figure~\ref{fig:lr_test} shows the test error averaged over $50$ repeated runs. More experiments for test log-likelihood presented in the Appendix. 

 
%
We see that both Algorithm~\ref{alg:lr_concentrate} and~\ref{alg:lr_diffuse} achieve lower test error than OPS at all privacy levels and across all datasets. This is to be expected, since OPS guarantees pure differential privacy which is stronger than \rdp. Comparing Algorithm~\ref{alg:lr_concentrate} and \ref{alg:lr_diffuse}, we can see that the latter always achieves better utility.

\section{Conclusion}

The inherent randomness of posterior sampling and the mitigating influence of a prior can be made to offer a wide range of privacy guarantees. Our proposed methods outperform existing methods in specific situations. The privacy analyses of the mechanisms fit nicely into the recently introduced RDP framework, which continues to present itself as a relaxation of DP worthy of further investigation.

\subsection*{Acknowledgements}
This work was partially supported by NSF under IIS 1253942, ONR under N00014-16-1-2616, and a Google Faculty Research Award.


\bibliographystyle{abbrv}
\bibliography{references}

\newpage
\appendix
\section{Appendix}


\subsection{Proofs of Exponential Family Sampling Theorems}

Our proofs will make extensive use of the definitions laid out in Section \ref{sec:expfambackground}. We will however need an additional definition for a modified version of $pset$, and as well the set of possible updates to the posterior parameter that might arise from the data.

\begin{Defn}
Let $lpset(\eta_0, n, b) = pset(\eta_0, n) + bDiff$. This is the set of posterior parameters that are $b$-neighboring at least one of the elements of $pset(\eta_0,n)$
\end{Defn}

\begin{Defn}
Let $U$ be the set of posterior updates for an exponential family, where $U$ is the convex hull of all vectors of the form $( S(x), 1)$ for $x,y \in \calX$.
\end{Defn}

We begin by noting that observing a data set when starting at a normalizable prior $\eta_0$ must result in a normalizable posterior parameter $\eta'$. 

\begin{Observation} \label{obs:validpost}
In a minimal exponential family, for any prior parameter $\eta_0$, any $n>0$, and any posterior update, every possible posterior parameter in the set $\eta_0+nU$ is also normalizable. As $C(\eta)$ must be a convex function for minimal families, this must apply to positive non-integer values of $n$ as well.
\end{Observation}

With this observation, we are ready to prove our result on the conditions under which sampling from our posterior gives a finite $(\lambda,\epsilon)$-RDP guarantee.

\begin{theorem} \label{thm:privacyDirectExpFamily}

For a $\Delta$-bounded minimal exponential family of distributions $p(x|\theta)$ with continuous log-partition function $A(\theta)$, there exists $\lambda^* \in (1,\infty]$ such Algorithm \ref{alg:expfamdirect} achieves $(\lambda,\epsilon(\eta_0,n,\lambda))$-RDP for $\lambda < \lambda^*$.

$\lambda^*$ is the supremum over all $\lambda$ such that all $\eta$ in the set $\eta_0 + (\lambda - 1)\Diff$ are normalizable.

\end{theorem}

\begin{proof2}

Algorithm \ref{alg:expfamdirect} samples directly from the posterior $\eta_{post} = \eta_0 + \sum_i (S(x_i), 1)$. When applied to neighboring data sets $\X$ and $\X'$, it selects posterior parameters that are neighboring.

The theorem can be reinterpreted as saying there exists $\lambda^*$ such that for $\lambda < \lambda^*$ we have

\begin{equation}
\sup_{\text{neighboring } \eta_P,\eta_Q \in pset(\eta_0,n)} D_\lambda ( p (\theta | \eta_P) || p(\theta | \eta_Q) ) < \infty \mbox{.}
\end{equation}

For these two posteriors from the same exponential family, we can write out the Rényi divergence in terms of the log-partition function $C(\eta)$.

\begin{equation} \label{eqn:expfamrenyiproof}
D_\lambda(p (\theta | \eta_P) || p(\theta | \eta_Q)) = \frac{C(\lambda \eta_P + (1-\lambda)\eta_Q)- \lambda C(\eta_P)}{\lambda - 1}+ C(\eta_Q)
\end{equation}

We wish to show that this is bounded above over all neighboring $\eta_P$ and $\eta_Q$ our mechanism might generate, and will do so by showing that $|C(\eta)|$ must be bounded every where it is applied in equation \eqref{eqn:expfamrenyiproof} if $\lambda < \lambda^*$. To find this bound, we will ultimately show each potential application of $C(\eta)$ lies within a closed subset of $E$, from which the continuity of $C$ will imply an upperbound.

Let's begin by observing that $\eta_P$ and $\eta_Q$ must lie within $pset(\eta_0,n)$ as they arise as posteriors for neighboring data sets $\X$ and $\X'$. The point $\eta_L = \lambda \eta_P + (1-\lambda)\eta_Q = \eta_P + (\lambda - 1) (\eta_P - \eta_Q)$ might not lie within $pset(\eta_0, n)$. However, we know $\eta_P - \eta_Q$ lies within $\Diff$ and that $\eta_L - \eta_P$ is within $(\lambda-1)\Diff$. This means for any neighboring data sets, $\eta_P$, $\eta_Q$, and $\eta_L$ lie inside $lpset(\eta_0,n,\lambda-1)$.

If $\lambda < \lambda^*$, then $\eta_0 + (\lambda -1)Diff \subseteq E$. The set $\eta_0 + (\lambda -1)Diff$ is potentially an open set, but the closure of this set must be within $E$ as well, since we can always construct $\lambda' \in (\lambda,\lambda^*)$ where $\eta_0 + (\lambda' -1)Diff \subseteq E$, and the points inside $\eta_0 + (\lambda -1)Diff$ can't converge to any point outside of $\eta_0 + (\lambda' -1)Diff$.

Any point in $\eta \in lpset(\eta_0,n,\lambda-1)$ can be broken down into three components using the definition of $lpset$: $\eta = \eta_0 + u + d$, where $u \in nU$ and $d \in (\lambda-1)Diff$. For any point in this $lpset$, we can therefore subtract off the component $u$ to reach a point in the set $\eta_0 + (\lambda - 1)Diff$. With Observation \ref{obs:validpost}, we can conclude that $\eta$ is normalizable if $\eta -u$ is normalizable, and therefore the closure of $lpset(\eta_0,n,\lambda-1)$ is a subset of $E$ if $\eta_0 + (\lambda -1)Diff$ is a subset of $E$, which we have shown for $\lambda < \lambda^*$.

As $C(\eta)$ is a continuous function, we know that the supremum of $|C(\eta)|$ over the closure of $lpset(\eta_0, n, \lambda - 1)$ must be finite. Remember that for any neighboring data sets, $\eta_P$,$\eta_Q$, and $\eta_L$ are inside  $lpset(\eta_0, n, \lambda - 1)$. Since $|C(\eta)|$ is bounded over this $lpset$, so too must our expression for $D_\lambda(p(\theta|\eta_P)||p(\theta|\eta_Q))$ in equation \eqref{eqn:expfamrenyiproof}. Therefore there exists an upper-bound for the order $\lambda$ Rényi divergence across all pairs of posterior parameters selected by Algorithm \ref{alg:expfamdirect} on neighboring data sets. This finite upper-bound provides a finite value for $\epsilon(\eta_0, n,\lambda)$ for which Algorithm \ref{alg:expfamdirect} offers $(\lambda, \epsilon(\eta_0,n,\lambda))$-RDP .

\end{proof2}

To prove our results for Algorithm \ref{alg:expfamdiffuse} and Algorithm \ref{alg:expfamconc}, we'll need an additional result that bounds the Rényi divergence in terms of the Hessian of the log-partition function and the distance between the two distribution parameters.

\begin{Lem} \label{lem:hessbound}

For $\lambda > 1$, if $||\nabla^2 C(\eta)|| < H$ over the set $\{ \eta_P + x(\eta_P - \eta_Q) | x \in [-\lambda + 1, \lambda - 1]\}$, then

\begin{equation}
D_\lambda(p(\theta|\eta_P)||p(\theta|\eta_Q)) \le ||\eta_P-\eta_Q||^2 H \lambda
\end{equation}

\end{Lem}

\begin{proof2}

Define the function $g(x) = C(\eta_P + xv)$ where $x\in \mathbb{R}$ and $v = \eta_P - \eta_Q$. This allows us to rewrite the R\'{e}nyi divergence as

\begin{equation}
D_\lambda(P||Q) = \frac{g(1-\lambda)- \lambda g(0)}{\lambda - 1}+ g(1)
\end{equation}

Now we will replace $g$ with its first order Taylor expansion

\begin{equation}
g(x) = g(0) + xg'(0) + e(x)
\end{equation}

where $e(x)$ is the approximation error term, satisfying $|e(x)| \leq x^2 \max_{y \in [-x,x]} g''(y)/2$.

This results in 

\begin{align}
D_\lambda(p(\theta|\eta_P)||p(\theta|\eta_Q)) &= \frac{g(0) + (1-\lambda)g'(0) + e(1-\lambda) - \lambda g(0)}{\lambda - 1}+ g(0) + g'(0) + e(1) \\
&= -\frac{e(1-\lambda)}{\lambda-1} + e(1) \\
&\leq \frac{(\lambda-1)^2}{\lambda - 1} \max_{y \in [-\lambda+1, \lambda-1]} g''(y)/2 + \max_{y \in [-1,1] } g''(y)/2 \mbox{ .}
\end{align}

Further, we can express $g''$ in terms of $C$ and $v$.

\begin{align}
g''(y) &= v^\intercal \nabla^2 C(\eta_P + yv) v\\
&\le ||\eta_P-\eta_Q||^2 ||\nabla^2 C(\eta_P + yv)||\\
&\le ||\eta_P-\eta_Q||^2H
\end{align}

Plugging in this bound on $g''$ gives the desired result.

\begin{align}
D_\lambda(p(\theta|\eta_P)||p(\theta|\eta_Q)) &\leq \frac{(\lambda-1)^2}{\lambda - 1} \max_{y \in [-\lambda+1, \lambda-1]} g''(y)/2 + \max_{y \in [-1,1] } g''(y)/2\\
&\leq (\lambda -1) ||\eta_P-\eta_Q||^2H/2 + ||\eta_P-\eta_Q||^2H/2\\
&\leq ||\eta_P-\eta_Q||^2H\lambda/2 \\
&\leq ||\eta_P-\eta_Q||^2H\lambda
\end{align}

\end{proof2}

We will also make use of the following standard results about the Hessian of the log-partition function of minimal exponential families, given in \citep{liese2007statistical} as Theorem 1.17 and Corollary 1.19 and rephrased for our purposes.

\begin{Thm}{(Theorem 1.17 from \citep{liese2007statistical})} \label{thm:allderivatives}
The log-partition function $C(\eta)$ of a minimal exponential family is infinitely often differentiable at parameters $\eta$ in the interior of the normalizable set $E$. 
\end{Thm}

\begin{Thm}{(Corollary 1.19 from \citep{liese2007statistical})} \label{thm:fullrank}
For minimal exponential family, the Hessian of the log-partition function $\nabla^2 C(\eta)$ is nonsingular for every parameter $\eta$ in the interior of the normalizable set $E$.
\end{Thm}

These results imply that the Hessian $\nabla^2 C(\eta)$ must exist and be continuous over $\eta$ in the interior of $E$, as well as having non-zero determinant.

\begin{theorem}

For any $\Delta$-bounded minimal exponential family with prior $\eta_0$ in the interior of $E$, any $\lambda > 1$, and any $\epsilon > 0$, there exists $r^* \in (0,1]$ such that using $r \in (0,r^*]$ in Algorithm \ref{alg:expfamdiffuse} will achieve $(\lambda,\epsilon)$-RDP.

\end{theorem}

\begin{proof2}

Recall that Algorithm \ref{alg:expfamdiffuse} uses the posterior parameter $\eta' = \eta_0 + r\sum_i^n(S(x),1)$ where the data contribution has been scaled by $r$. Our first step of this proof is to show that there exists $r_0 \in (0,1]$ such that the order $\lambda$ Rényi divergences of the generated parameters are finite for $r < r_0$.

Similar to the proof of Theorem \ref{thm:privacyDirectExpFamily}, we will do so by creating a closed set where $C(\eta)$ is finite and that must contain $\eta_P, \eta_Q,$ and $\eta_L$ for any choice of neighboring data sets.  On neighboring data sets, this generates $r$-neighboring parameters $\eta_P$ and $\eta_Q$. The point $\eta_L = \lambda \eta_P + (1-\lambda)\eta_Q$ is therefore $r(\lambda-1)$-neighboring $\eta_P$. These points must be contained in the set $lpset(\eta_0, rn, r(\lambda-1)) = \eta_0 + rnU + r(\lambda-1)Diff$. For any point in this set, we can subtract off the component in $rnU$ to get to a modified prior that is $r(\lambda-1)$-neighboring $\eta_0$.

By the assumption that $\eta_0$ is in the interior of $E$, there exists $\delta > 0$ such that the ball $\mathcal{B}(\eta_0,\delta) \subseteq E$. For the choice $r_0 = \frac{\delta}{2(\lambda-1)\Delta}$, for any $r \in (0,r_0)$, the modified prior we constructed for each point in $lpset(\eta_0, rn, r(\lambda-1))$ is within distance $r(\lambda-1)\Delta$ of $\eta_0$ and therefore within $\mathcal{B}(\eta_0,\delta/2) \subset \mathcal{B}(\eta_0,\delta) \subseteq E$. Observation \ref{obs:validpost} then allows us to conclude that every point $\eta$ in $lpset(\eta_0, rn, r(\lambda-1))$ has an open neighborhood of radius $\delta_2$ where $C(\eta)$ is finite. This is enough to conclude that the closure of this $lpset$ must also lie entirely within $E$, and $C(\eta)$ is finite and continuous over this closed set. As in Theorem \ref{thm:privacyDirectExpFamily}, this suffices to show that the supremum of order $\lambda$ Rényi divergences on neighboring data sets is bounded above.

We have thus shown there exists $r_0$ where the $\epsilon$ of our $(\lambda,\epsilon)$-RDP guarantee is finite for $r<r_0$. However, our goal was to achieve a specific $\epsilon$ guarantee. Our proof of the existence of $r^*$ centers around the claim that there must exist a bound $H$ for the Hessian of $C(\eta)$ over all choices of $r \in [0,r_0)$.

We can construct the set $D = \cup_{r \in [0,r_0]} lpset(\eta_0, rn, r(\lambda-1))$, which will contain every possible $\eta_P, \eta_Q$, and $\eta_L$ that might arise from any pair neighboring data sets and any choice of $r$ in that interval. The previous argument still applies: each point in this union must have an open neighborhood of radius $\delta/2$ that is a subset of $E$. This is enough to conclude that closure of $D$ is also a subset of $E$. Theorem \ref{thm:allderivatives} implies $\nabla^2 C(\eta)$ exists and is continuous on the interior of $E$, and this further implies that there must exist $H$ such that for all $\eta$ in this closure we have $||\nabla^2 C(\eta)|| \leq H$.

For any value $r$, we know that $\eta_P$ and $\eta_Q$ are $r$-neighboring, so we know $||\eta_P - \eta_Q|| \leq r\Delta$. Since $D$ contains $lpset(\eta_0, rn, r(\lambda-1))$, the bound $H$ must apply for all $\eta$ in the set $\{ \eta_P + x(\eta_P - \eta_Q) | x \in [-\lambda + 1, \lambda - 1]\}$. This allows us to use Lemma \ref{lem:hessbound} to get the following expression:

\begin{align}
D_\lambda(p(\theta|\eta_P)||p(\theta|\eta_Q)) &\le ||\eta_P-\eta_Q||^2 H \lambda \\
&\le r\Delta^2H\lambda \mbox{.}
\end{align}

If we set $r^* = \frac{\epsilon}{\Delta^2H\lambda}$, then for $r < r^*$ the order $\lambda$ Rényi divergence of Algorithm \ref{alg:expfamdiffuse} is bounded above by $\epsilon$, which gives us the desired result.

\end{proof2}

The concentrated mechanism is a bit more subtle in how it reduces the influence of the data, and so we need this result modified from Lemmas 9 and 10 in the appendix of \citep{foulds2016theory}. These results are presented here in a way that matches our notation. It effectively states that if we start at a prior $\eta_0$ satisfy mild but technical regularity assumptions, then the Hessians $C(k\eta_0)$ must converge to zero as $k$ grows. In practical terms, this implies the covariance of our prior distribution must shrink as we increase the number of pseudo-observations.

\begin{Defn}
Let $T^*_\eta = T(\argmax_{\theta\in\Theta} \eta \cdot T(\theta))$. This represents the mode of the sufficient statistics under the distribution $p(T(\theta) | \eta)$.
\end{Defn}

\begin{Lem}{(Lemma 9 from \citep{foulds2016theory})} \label{lem:modemapinterior}
If $A(\theta)$ is continuously differentiable and $\eta_0$ is in the interior of $E$, then $\argmax_{\theta\in\Theta} \eta \cdot T(\theta)$ must be in the interior of $\Theta$.
\end{Lem}

\begin{Lem}{(Lemma 10 from \citep{foulds2016theory})} \label{lem:bigetalowvar}
If we have a minimal exponential family in which $A(\theta)$ is differentiable of all orders, there exists $\delta_1 > 0$ such that the ball $\mathcal{B}(\eta_0,\delta_1)$ is a subset of $E$, there exists $\delta_2 > 0$ and a bound $L$  such that  all the seventh order partial derivatives of $A(\theta)$ on the set  $D_{\eta_0,\delta_1,\delta_2} = \{ \theta | \min_{\eta \in \mathcal{B}(\eta_0,\delta_1)} ||T(\theta) - T^*_\eta|| < \delta_2 \}$ are bounded by $P$, and the determinant of $\nabla^2 A(\theta)$ is bounded away from zero on $ D_{\eta_0,\delta_1,\delta_2}$, then there exists real number $V,K$ such that for $k > K$ we have

\begin{equation}
\forall \eta \in \mathcal{B}(\eta_0,\delta_1)  \hskip5pt ||\nabla^2 C(k\eta)|| < \frac{V}{k} \mbox{.}
\end{equation}

\end{Lem}

\begin{theorem}

For any $\Delta$-bounded minimal exponential family with prior $\eta_0$ in the interior of $E$, for any $\lambda > 1$, and any $\epsilon > 0$, there exists $m^* \in (0,1]$ such that using $m \in (0,m^*]$ in Algorithm \ref{alg:expfamconc} will achieve $(\lambda,\epsilon)$-RDP.

\end{theorem}

\begin{proof2}

For a fixed value of $m$, recall that Algorithm \ref{alg:expfamconc} selects the posterior parameter $\eta' = m^{-1}\eta_0 + \sum_{i=1}^n(S(x_i),1)$. For neighboring data sets $\X$ and $\X'$, the selected posterior parameters $\eta_P$, $\eta_Q$, and $\eta_L = \lambda \eta_P + (1-\lambda)\eta_Q$ lie within $lpset(m^{-1}\eta_0,n,\lambda-1) = m^{-1}\eta_0 + nU + (\lambda-1)Diff$.

We start by showing that the conditions of Lemma \ref{lem:bigetalowvar} are met. As we assumed $\eta_0$ is in the interior of $E$, there exists $\delta_1 > 0$ such that we have the ball $\mathcal{B}(\eta_0,\delta_1) \subseteq E$.  By Theorem \ref{thm:allderivatives}, the log-partition function of the data likelihood $A(\theta)$ is differentiable of all orders, and Theorem \ref{thm:fullrank} tells us that the Hessian $\nabla^2 A(\theta)$ is non-singular with non-zero determinant on the interior of $\Theta$. This permits the application of Lemma \ref{lem:modemapinterior}, offering a mapping from $\eta$ in the interior of $E$ to their mode $T^*_\eta$ corresponding to a parameter $\theta$ in the interior of $\Theta$. Knowing that $A(\theta)$ is infinitely differentiable on the interior of $\Theta$ further implies that the seventh order derivatives are well-behaved in a neighborhood around each mode resulting from this mapping.  This provides the rest of the requirements for Lemma \ref{lem:bigetalowvar}.

Therefore there exists $V$ and $K$ such that the following holds

\begin{equation}
\forall \eta \in \mathcal{B}(\eta_0,\delta_1) \mbox{ : }||\nabla^2 C(k\eta)|| \leq \frac{V}{k} \mbox{.} \label{eqn:conchessbound}
\end{equation}

We wish to show that $||\nabla^2C(\eta)||$ must be bounded on the expanded set $lpset(m^{-1}\eta_0, n, \lambda - 1) = m^{-1}\eta_0 + nU + (\lambda-1)\Diff$, and will do so by showing that for small enough $m$ we can use equation \eqref{eqn:conchessbound} to bound the Hessians.

Let $\alpha(\eta)$ denote the last coordinate of $\eta$. This represents the pseudo-observation count of this parameter, and notice that $\forall u \in U \mbox{ : } \alpha(u) = 1$ and $\forall v \in Diff \mbox { : } \alpha(v) = 0$. We are going to analyze the scaled set $c_m\cdot lpset(m^{-1}\eta_0, n, \lambda - 1)$ where $c_m$ is a positive scaling constant that will depend on $m$.

\begin{equation}
c_m\cdot lpset(m^{-1}\eta_0, n, \lambda - 1) = c_mm^{-1}\eta_0 + c_mnU + c_m(\lambda-1)\Diff
\end{equation}. 

For each $\eta$ in this $c_m\cdot lpset$, we have 

\begin{equation}
\alpha(\eta) = c_mm^{-1}\alpha(\eta_0) + c_mn\cdot1 + c_m(\lambda-1)\cdot0 =  c_m(m^{-1}\alpha(\eta_0) + n) \mbox{ .}
\end{equation}

Setting $c_m = \frac{\alpha(\eta_0)}{m^{-1}\alpha(\eta_0) +n}$ thus guarantees that for all $\eta$ in $c_m\cdot lpset(m^{-1}\eta_0, n, \lambda - 1)$ we have $\alpha(\eta) = \alpha(\eta_0)$. We want to know how far the points in this $c_m\cdot lpset$ are from $\eta_0$, so we simply subtract $\eta_0$ to get a set $D_m$ of vectors. These offset vectors have the form $c_m\cdot lpset(m^{-1}\eta_0, n, \lambda - 1) - \eta_0$ and therefore lie in the set 

\begin{equation}
 D_m =(c_mm^{-1} - 1)\eta_0 + c_mnU + c_m(\lambda -1)\Diff \mbox{.}
\end{equation}

Using our expression of $c_m$ as a function of $m$, we can see the following limiting behavior:

\begin{align}
\lim_{m \rightarrow 0} c_m &= \lim_{m \rightarrow 0} \frac{\alpha(\eta_0)}{m^{-1}\alpha(\eta_0)  +n} = 0\\
\lim_{m \rightarrow 0} c_mm^{-1} - 1 &= \lim_{m \rightarrow 0} \frac{m^{-1}\alpha(\eta_0)}{m^{-1}\alpha(\eta_0)  +n} - 1 =  1 -1 = 0 \mbox{.}
\end{align}

These limits lets us take the limit of the size of the vectors in $D_m$ as $m\rightarrow0$:
\begin{align}
\lim_{m \rightarrow 0}\sup_{v\in D_m} ||v|| &\leq \lim_{m \rightarrow 0} (c_mm^{-1} - 1)||\eta_0|| + c_mn\sup_{u_1\in U}||u_1|| + c_m(\lambda-1)\sup_{u_2\in\Diff}||u_2|| \\
&\leq 0\cdot ||\eta_0|| + 0\cdot\sup_{u_1\in U}||u_1|| + 0\cdot\sup_{u_2\in\Diff}||u_2|| \\
&\leq 0\mbox{.} \label{eqn:differencelimit}
\end{align}

This limit supremum on $D_m$ tells us that as $m \rightarrow 0$, the maximum distance between points in the scaled set $c_m\cdot lpset(m^{-1}\eta_0,n,\lambda-1)$ and $\eta_0$ gets arbitrarily small. This means there exists some $m_0$ such that for $m < m_0$ the scaled set $c_m\cdot lpset(m^{-1}\eta_0,n,\lambda-1)$ lies within $\mathcal{B}(\eta_0,\delta_1)$. This scaling mapping can be inverted, and it implies $lpset(m^{-1}\eta_0,n,\lambda-1)$ is contained within $\frac{1}{c_m}\mathcal{B}(\eta_0,\delta_1)$. Being contained within this scaled ball is precisely what we need to use equation \eqref{eqn:conchessbound} with $\frac{1}{k} = c_m$.

Equation \eqref{eqn:conchessbound} bounds $||\nabla^2C(\eta)|| \leq H_m = Vc_m$ for all $\eta$ in $lpset(m^{-1}\eta_0,n,\lambda-1)$, which in turn lets us use Lemma \ref{lem:hessbound} to bound our Rényi divergences.

\begin{align}
D_\lambda(p(\theta|\eta_P)||p(\theta|\eta_Q)) &\le ||\eta_P-\eta_Q||^2 H_m \lambda \\
&\le \Delta^2Vc_m\lambda \mbox{.}
\end{align}

As we have $c_m \rightarrow 0$ as $m \rightarrow 0$, we know there must exist $m^*$ such that for $m < m^*$ we have $c_m \leq \frac{\epsilon}{\Delta^2V\lambda}$. This means the order $\lambda$ Rényi divergences of Algorithm \ref{alg:expfamconc} on neighboring data sets is bounded above by $\epsilon$, which gives us the desired result.

\end{proof2}

We have one last theorem to prove, the result claiming the Rényi divergences of order $\lambda$ between $\eta_P$ and its neighbors is convex, which greatly simplifies finding the supremum of these divergences over the convex sets being considered.

\begin{theorem} \label{thm:appdivconvexity}

Let $e(\eta_P,\eta_Q,\lambda) =  D_\lambda\left(p(\theta|\eta_P) || p(\theta|\eta_Q)\right)$.

For a fixed $\lambda$ and fixed $\eta_P$, the function $e$ is a convex function over $\eta_Q$.

If for any direction $v \in \Diff$, the function $g_v(\eta) = v^\intercal \nabla^2C(\eta) v$ is convex over $\eta$, then for a fixed $\lambda$, the function

\begin{equation}
f_\lambda(\eta_P) = \sup_{\eta_Q \hskip3pt r-\text{neighboring } \eta_P} e(\eta_P,\eta_Q,\lambda)
\end{equation}

is convex over $\eta_P$ in the directions spanned by $\Diff$.

\end{theorem}

\begin{proof2}

First, we can show that for a fixed $\eta_P$ and fixed $\lambda$, the choice of $\eta_Q$ in the supremum must lie on the boundary of possible neighbors. This is derived from showing that $D_\lambda(P||Q)$ is convex over the choice of $\eta_Q$.

Consider once again the expression for our Rényi divergence, expressed now as the function $e(\eta_P,\eta_Q,\lambda)$:

\begin{align}
e(\eta_P,\eta_Q,\lambda) = D_\lambda(P||Q) = \frac{C(\lambda \eta_P + (1-\lambda)\eta_Q)- \lambda C(\eta_P)}{\lambda - 1}+ C(\eta_Q) \mbox{.}
\end{align}

Let $\nabla_{\eta_Q} e(\eta_P,\eta_Q,\lambda) $ denote the gradient of the divergence with respect to $\eta_Q$.

\begin{align}
\nabla_{\eta_Q} e(\eta_P,\eta_Q,\lambda) &= \nabla C(\eta_Q) + \frac{1-\lambda}{\lambda-1} \nabla C(\lambda \eta_P + (1-\lambda)\eta_Q) \\
&= \nabla C(\eta_Q) - \nabla C(\lambda \eta_P + (1-\lambda)\eta_Q) \mbox{.}
\end{align}

We can further find the Hessian with respect to $\eta_Q$: 

\begin{align}
\nabla^2_{\eta_Q} e(\eta_P,\eta_Q,\lambda) &= \nabla^2 C(\eta_Q) - (1-\lambda) \nabla^2 C(\lambda \eta_P + (1-\lambda)\eta_Q) \mbox{.}
\end{align}

By virtue of being a minimal exponential family, we know $C$ is convex and thus $\nabla^2 C$ is PSD everywhere. Combined with the fact that $\lambda > 1$, this is enough to conclude that $\nabla^2_{\eta_Q} e(\eta_P,\eta_Q,\lambda)$ is also PSD for everywhere with $\lambda > 1$. This means $e(\eta_P,\eta_Q,\lambda)$ is a convex function with respect to $\eta_Q$ for any fixed $\eta_P$ and $\lambda$.

We now wish to characterize the function $f_\lambda(\eta_P)$, which takes a supremum over $\eta_Q \in \eta_P + r\Diff$ of $e(\eta_P,\eta_Q,\lambda)$.

\begin{equation}
f_\lambda(\eta_P) = \sup_{\eta_Q r-\text{neighboring } \eta_P} e(\eta_P,\eta_Q,\lambda)
\end{equation}

We re-parameterize this supremum in terms of the offset $b = \eta_Q - \eta_P$.

\begin{equation}
f_\lambda(\eta_P) = \sup_{b \in r\Diff} e(\eta_P,\eta_P+b,\lambda)
\end{equation}

Now for any fixed offset $b$, x we can find the expression for the Hessian of $\nabla^2_{\eta_P} e(\eta_P,\eta_P+b,\lambda)$. 

\begin{align}
\nabla^2_{\eta_P}e(\eta_P,\eta_P+b,\lambda) = \nabla^2 C(\eta_P + b) - \frac{\lambda}{\lambda-1}\nabla^2 C(\eta_P) + \frac{1}{\lambda - 1} \nabla^2 C(\eta_p + (1-\lambda)b)
\end{align}

We wish to show this Hessian is PSD, i.e. for any vector $v$ we have $v^\intercal \nabla^2_{\eta_P}e(\eta_P,\eta_P+b,\lambda) v$ is non-negative. We can rewrite this in terms of the function $g_v(\eta)$ introduced in the theorem statement.

\begin{align}
v^\intercal \nabla^2_{\eta_P}e(\eta_P,\eta_P+b,\lambda) v &= g_v(\eta_P + b) - \frac{\lambda}{\lambda-1}g_v(\eta_P) + \frac{1}{\lambda - 1} g_v(\eta_p + (1-\lambda)b) \\
&= \frac{\lambda}{\lambda-1}\left( \frac{\lambda-1}{\lambda}g_v(\eta_P + b) - g_v(\eta_P) + \frac{1}{\lambda} g_v(\eta_p + (1-\lambda)b)\right) \label{eqn:usejensen}
\end{align}

We know $\frac{\lambda}{\lambda-1} > 0$ and that $\eta_P$ must lie between $\eta_P+b$ and $\eta_P - (\lambda-1)b$. Our assumption that $g_v(\eta)$ is convex over $\eta$ for all directions $v$ then lets us use Jensen's inequality to  see that the expression \eqref{eqn:usejensen} must be non-negative.

This lets us conclude that $v^\intercal (\nabla^2_{\eta_P} e(\eta_P,\eta_P+b,\lambda) v \ge 0$ for all $v$, and thus this Hessian is PSD for any $\eta_P$. This in turn means our divergence $e(\eta_P,\eta_P+b,\lambda)$ is convex over $\eta_P$ assuming a fixed offset $b$.

We return to $f_\lambda(\eta_P)$, and observe that it is a supremum of functions that are convex, and therefore it is convex as well.

\end{proof2}

\subsection{Additional Beta-Bernoulli Experiments}

\begin{figure*}[!t]
\centering
\begin{subfigure}[b]{0.32\textwidth}
\includegraphics[width=\textwidth]{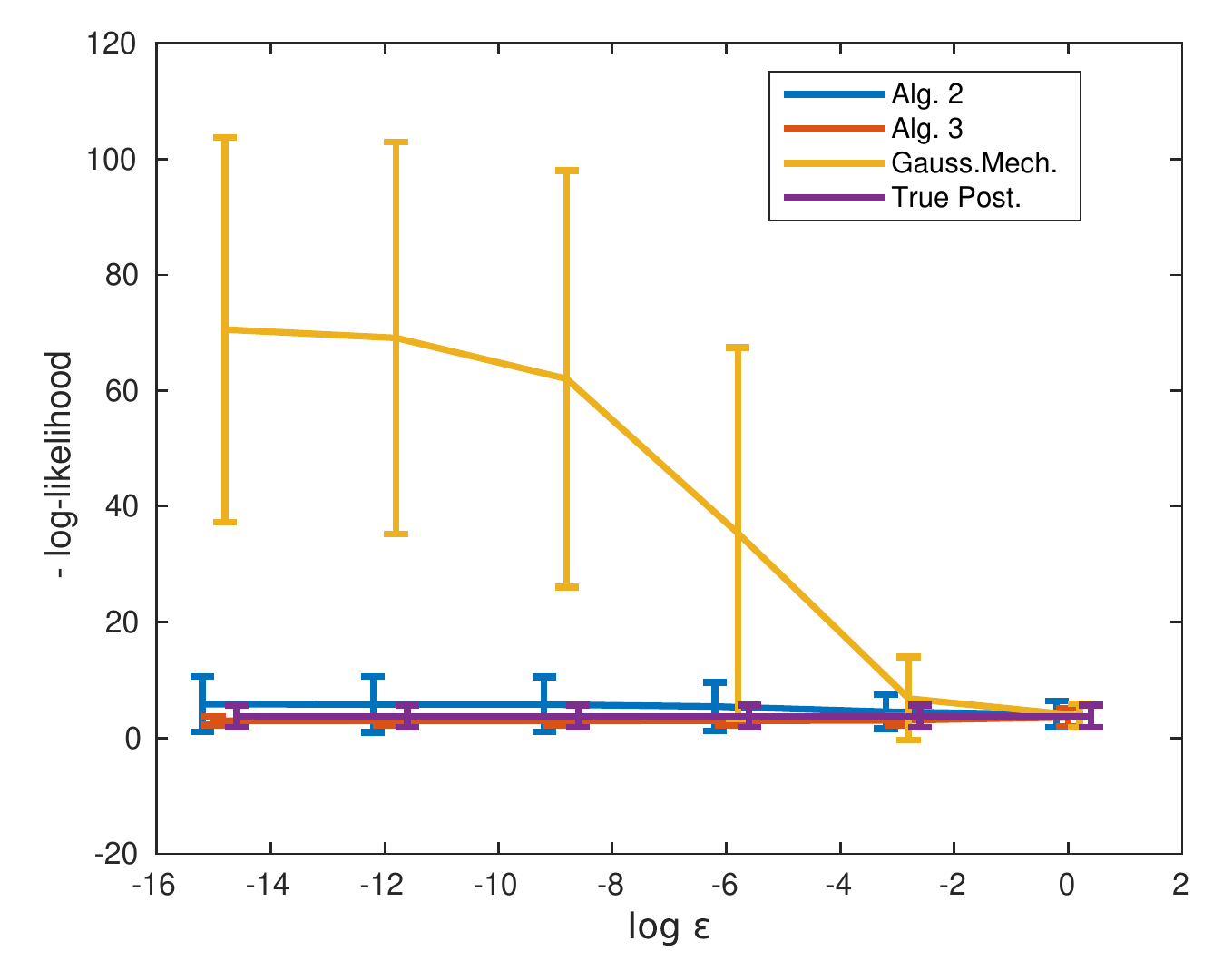}
\caption{$\rho = 1/3$ (high match with $\eta_0$)}
\end{subfigure}
\begin{subfigure}[b]{0.32\textwidth}
\includegraphics[width=\textwidth]{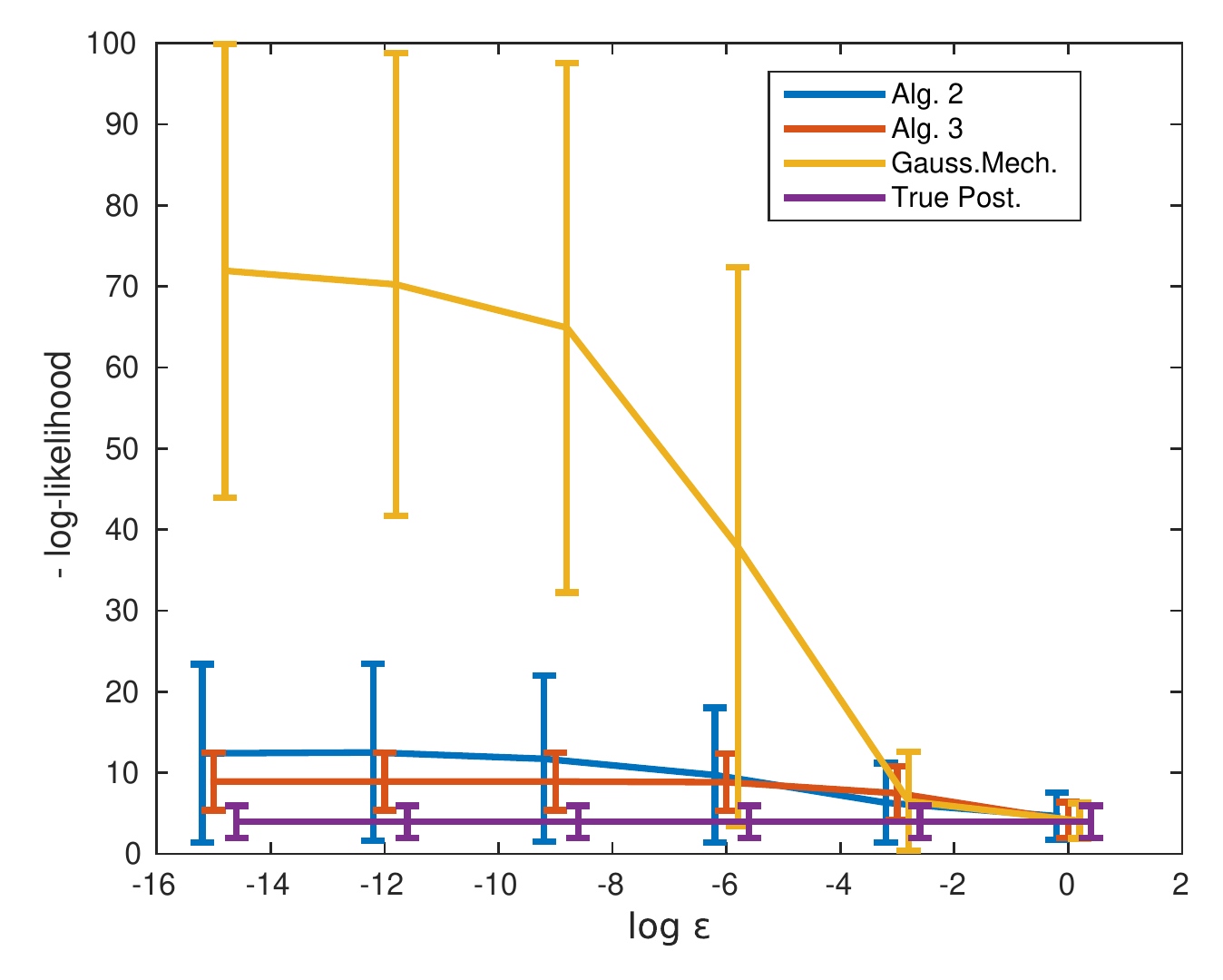}
\caption{$\rho = 1/2$}
\end{subfigure}
\begin{subfigure}[b]{0.32\textwidth}
\includegraphics[width=\textwidth]{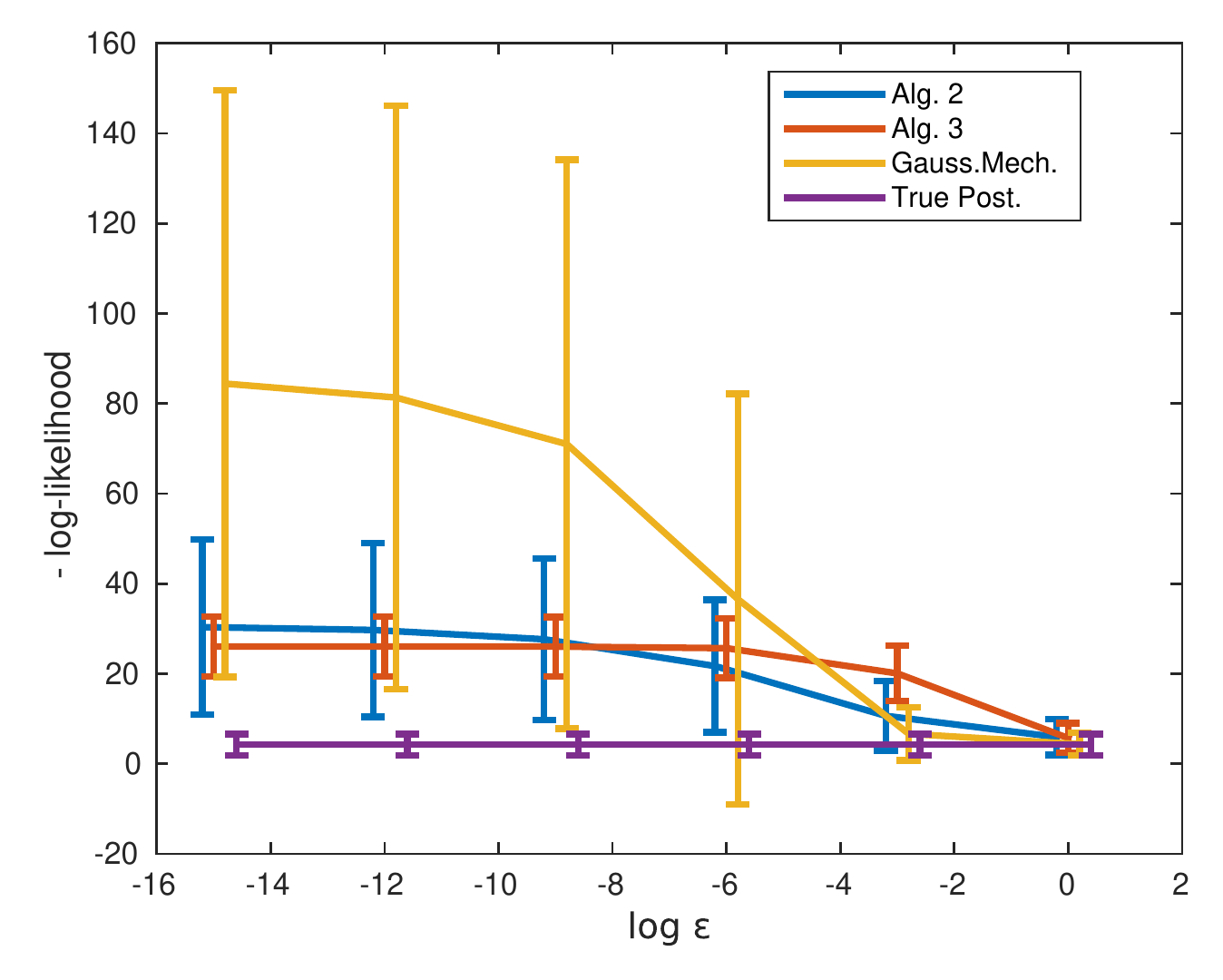}
\caption{$\rho = 2/3$ (low match with $\eta_0$)}
\end{subfigure}
\caption{Utility Comparison for a fixed $\eta_0$ but varying true population parameter}
\label{fig:priormatching}
\end{figure*}

The utility of the prior-based methods (Algorithms \ref{alg:expfamdiffuse} and \ref{alg:expfamconc}) depends on how well the prior matches the observed data. Figure \ref{fig:priormatching} shows several additional situations for the experimental procedure of measuring the log-likelihood of the data.

In each case, the prior $\eta_0 = (6,18)$ was used, and both $\X$ and $\X_H$ had 100 data points. $\lambda=15$ was fixed in these additional experiments. The only thing that varies is the true population parameter $\rho$. In (a), $\rho = 1/3$ closely matches the predictions of the prior $\eta_0$. In (b), $\rho = 0.5$, presented as an intermediate case where the prior is misleading. Finally, in (c), $\rho = 2/3$, which is biased in the opposite direction as the prior. In all cases, the proposed methods act conservatively in the face of high privacy, but in (a) this worst case limiting behavior still has high utility. Having a strong informative prior helps these mechanisms. The setting in which the prior is based off of a representative sample of non-private data from the same population as the private data is likely to be beneficial for Algorithms \ref{alg:expfamdiffuse} and \ref{alg:expfamconc}.

\begin{figure*}[!t]
\centering
\includegraphics[width=0.4\textwidth]{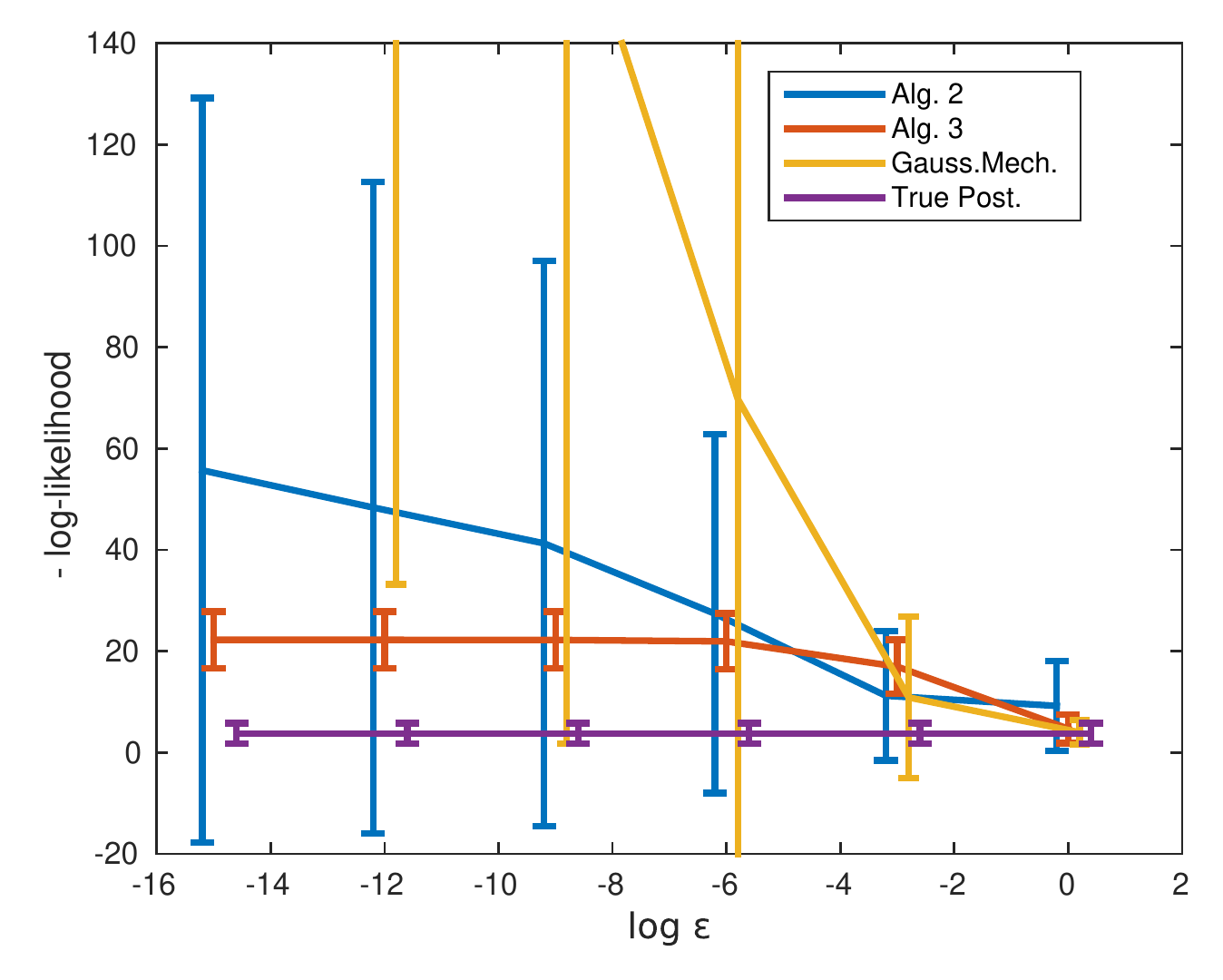}
\caption{Utility Experiment for the non-informative uniform prior}
\label{fig:expfamuniform}
\end{figure*}

One other case is presented in Figure \ref{fig:expfamuniform}, where $\rho = 0.2$ but the prior has been changed to $\eta_0 = (1,2)$. $\lambda$ is still 15, and the number of data points is still 100. This prior corresponds to the uniform prior, as it assigns equal probability to all estimated data means on $(0,1)$. It represents an attractive case on a non-informative prior, but also represents a situation in which privacy is difficult. In particular, $\lambda^* = 2$ in this setting. When Algorithm \ref{alg:expfamconc} scales up this prior, it becomes concentrated around $\rho = 0.2$, so this setting also corresponds to a case where the true population parameter does not match well with the predictions from the prior.

\subsection{Application to other exponential families}

\subsubsection{Dirichlet-Categorical}

The Categorical family is a higher dimension generalization of the Bernoulli family. Instead of just two possible values, (e.g. "failure" or "success", 0 or 1), a categorical variable is allowed to take any of $d$ discrete values. The parameters of a categorical distribution assign a probability to each of the discrete values. These probabilities are constrained to sum to one in order to be a valid distribution, so this family of distribution can be described with only $d-1$ parameters.

Our propsed method works with this family as well, but the proof is a little more difficult due to the higher dimensions.

Let the space of observations $\calX = \{ 1, 2, \ldots, d\}$. The sufficient statistics of an observation $x$ is a vector of indicator variables, $S(x) = \{ \mathbb{I}_1(x), \ldots, \mathbb{I}_{d-1}(x)\}$. Notice that $\mathbb{I}_d(x)$ is not included, since it can be derived from the other coordinates of $S(x)$. Including this last indicator variable would make the family non-minimal where the sufficient statistics satsify the linear relationship $\sum_{i=1}^d \mathbb{I}_i = 1$.

The conjugate prior family is the Dirichlet family. Under our construction of the conjugate prior, we want the parameter $\eta$ to satisfy the relationship $\eta_{posterior} = \eta_{prior} + (S(x), 1)$. This means that $\eta$ is $d$ dimensional, 
and the last coordinate of $\eta$ measures an effective count of observations. Since each coordinate of $S(x)$ is bounded by one, we also have the relationship that for any posterior, $\eta^{(d)} \geq \eta^{(i)}$ for $i \in [d]$.

When $d=2$, this derivation exactly matches the one from the Beta-Bernoulli system and it is $\Delta$-bounded for $\Delta = 1$.

For $d > 2$, this family is $\Delta$-bounded for $\Delta = \sqrt{2}$. For any two observations, $S(x) - S(y)$ is non-zero in atmost two locations, and each location has a difference of at most 1.

Further, this Dirichlet-Categorical system satisfies the requirements of Theorem \ref{thm:appdivconvexity}. The necessary requirement is that for any direction $v \in \Diff$, the function $g_v(\eta) = v^\intercal \nabla^2C(\eta) v$ is convex over $\eta$. For this system, we have an expression for $C(\eta):$

\begin{equation}
C(\eta) = \sum_{k = 1}^{d-1} \ln{ \Gamma(\eta^{(k)})} + \ln{\Gamma (\eta^{(d)} - \sum_{i = k}^{d-1} \eta^{(k)})} - \ln{\Gamma(\eta^{(d)})}
\end{equation}

This value is merely the sum of the log-Gamma function applied to the count of observations at each value, minus the log-Gamma function applied to the total count of observations. The expression $\eta^{(d)} - \sum_{i = 1}^{d-1} \eta^{(i)}$ evaluates to the count of observations located at the implicit $d^{th}$ value, since $\eta^{(d)}$ carries the total count of observations seen.

With this expression, we can calculate the gradient and Hessian. The digamma function $\psi_0$ is the derivative of the log-Gamma function $\ln{\Gamma(\cdot)}$, and the trigamma function $\psi_1$ is the derivative of the digamma function.

\begin{equation}
\nabla C(\eta)^{(i)} = \begin{cases} 
      \psi_0\left(\eta^{(i)}\right) - \psi_0\left(\eta^{(d)} - \sum_{k = 1}^{d-1} \eta^{(k)}\right) & i \not = d \\
      \psi_0\left(\eta^{(d)} - \sum_{k = 1}^{d-1} \eta^{(k)}\right) - \psi_0\left(\eta^{(d)}\right) &  i = d
   \end{cases}
\end{equation}

\begin{equation}
\nabla^2 C(\eta)^{(i,j)} = \begin{cases} 
      \psi_1\left(\eta^{(i)}\right) + \psi_1\left(\eta^{(d)} - \sum_{k = 1}^{d-1} \eta^{(k)}\right) & i = j \not = d \\
      \psi_1\left(\eta^{(d)} - \sum_{k = 1}^{d-1} \eta^{(k)}\right) &  i \not = j, i \not = d, j \not = d \\
      -\psi_1\left(\eta^{(d)} - \sum_{k = 1}^{d-1} \eta^{(k)}\right) &  i \not = j, i  = d, j \not = d \\
      -\psi_1\left(\eta^{(d)} - \sum_{k = 1}^{d-1} \eta^{(i)}\right) &  i \not = j, i \not = d, j  = d \\
      \psi_1\left(\eta^{(d)} - \sum_{k = 1}^{d-1} \eta^{(k)}\right)- \psi_1(\eta^{(d)})& i = j = d
   \end{cases} \label{eq:cathessmatrix}
\end{equation}

When $v\in Diff$, the last coordinate of $v$ is zero since changing one observation does not change the total count of observations. This means the expression $g_v(\eta) = v^\intercal \nabla^2C(\eta) v$ can ignore the last coordinate of $v$, as well as the last row and column of $\nabla^2 C(\eta)$. This means we are only concerned with the entries matching the first two cases of equation \eqref{eq:cathessmatrix}.   Let $\tilde{v}$ denote the vector formed by the first $d-1$ coordinates of $v$.

A careful examination the matrix $M$ equal to the top $d-1$ rows and and leftmost $d-1$ columns of $\nabla^2 C(\eta)$ reveals that $M$ decomposes as

\begin{equation}
M = \psi_1\left(\eta^{(d)} - \sum_{k = 1}^{d-1} \eta^{(k)}\right) [\mathbf{1}] + diag\left(\psi_1(\eta^{(1)}),\ldots,\psi_1(\eta^{(d-1)})\right) 
\end{equation}

where $[\mathbf{1}]$ is the $d-1$ by $d-1$ matrix where all entries are 1, and $diag$ constructs a diagonal matrix from the given values. This means for all $v \in Diff$, we have the following expression:

\begin{align}
g_v(\eta) &= v^\intercal \nabla^2C(\eta) v \\
&= \tilde{v}^\intercal M \tilde{v} \\
&= \tilde{v}^\intercal  \left( \psi_1(\eta^{(d)} - \sum_{k = 1}^{d-1} \eta^{(k)}) [\mathbf{1}] + diag(\psi_1(\eta^{(1)},\ldots,\eta^{(d-1)}) \right)\tilde{v} \\
&= \psi_1(\eta^{(d)} - \sum_{k = 1}^{d-1} \eta^{(k)}) \left(\tilde{v}^\intercal [\mathbf{1}] \tilde{v}\right) + \sum_{i=1}^{d-1} \psi_1(\eta^{(i)}) (\tilde{v}^{(i)})^2 \\
\end{align}

With the fact that $[\mathbf{1}]$ is PSD and that $(\tilde{v}^{(i)})^2$ is always positive, the above calculations show that $g_v(\eta)$ is the sum of many applications of the digamma function $\psi_1$. Each of these applications has a positive coefficient, and the function $\psi_1$ is convex. This concludes the proof that $g_v(\eta)$ is convex over $\eta$ for any $v\in Diff$. (When $d=2$, this expression for $g_v$ in fact matches the one derived from the Beta-Bernoulli system.)

This means that the expression for the worst-case Rényi divergence between neighboring posterior parameters is convex, and so the maximum must be located at the boundaries. In this case, the $pset$ is a shifted simplex, so the maximum must occur at one of the vertices.

The potential pairs of posterior parameters that must be checked in order to evaluate the RDP guarantee grows as $O(d^3)$.

\subsubsection{Gaussian-Gaussian and non-$\Delta$-bounded families}

Another interesting setting is estimating the mean of a Gaussian variable when the variance is known. In this case, the conjugate prior is also a Gaussian distribution.

This system satisfies the $v^\intercal \nabla^2C(\eta) v$ convexity requirement for Theorem \ref{thm:appdivconvexity}, since the variance $\nabla^2 C(\eta)$ is constant when the final coordinate (the total count of observations) is fixed. Thus for any $v \in Diff$, the function $g_v(\eta)$ is constant and therefore convex.

However, this setting does not satisfy the $\Delta$-bounded assumption. The observations can be arbitrarily large, and changing a single observation can therefore lead to arbitrarily large changes to posterior parameters and thus also arbitrarily large Rényi divergences between neighboring data sets. 

The exponential family mathematics behind our results did not directly depend on the $\Delta$-boundedness assumption. Instead, this bound was used only to bound the $pset$ of possible posterior parameters in order to bound the distance $||\eta_P - \eta_Q||$ when considering neighboring data sets. This bounded $pset$ then ensured our privacy guarantee was finite. 

For any given data set $\X$, we can bound the Rényi divergence between the posterior from $\X$ and the posterior from any other data set $\X'$ satisfying $S(\X) - S(\X') \leq  \Delta$. This is true even when the exponential family is not $\Delta$-bounded.

This permits two different approaches: we can relax the RDP framework further, protecting only data sets and a select bounded range of neighboring data sets rather than all the neighbors, or we can include a data preprocessing step that projects the observations onto a set with bounded sufficient statistics. The latter approach permits the use of the RDP framework without introducing further relaxations.

For example, we could replace the observations $\X$ with $\tilde{\X} = f(X)$ where the following function $f$ was applied to each observation $x$ in $\X$:

\begin{equation}
f(x) = \begin{cases} 
      -\Delta & x \leq -\Delta \\
      x & -\Delta < x < \Delta \\
      \Delta & \Delta \leq x 
   \end{cases}
\end{equation}

Although the statistical model still believes arbitrarily large observations are possible, the preprocessing projection step allows us to bound $||\eta_P - \eta_Q|| \leq \Delta$ where $\eta_P$ is the posterior for $f(\X)$ and $\eta_Q$ is the posterior for $f(\X')$ with any neighboring set of observations $\X'$.

This comes with the caveat that our model no longer matches reality, since it is unaware of the distortions introduced by our  preprocessing step. We leads to a potential degradation of utility for the mechanism output, but our privacy guarantees will hold. If the data altered by $f$ is sufficiently rare, these distortions should be minimal.

\subsection{Proofs in Section~\ref{sec:glm}}
\subsubsection{GLMs Privacy Proof}
In this section we prove Theorem~\ref{lem:LR_privacy0}, \ref{lem:LR_privacy}.
Here we state and prove a more general version of the theorems.
Consider any problem with likelihood in the form
\begin{align*}
p(D|w) = \exp{-\sum_{i=1}^n \ell(w,x_i,y_i)}
\end{align*}
and posterior of the following form 
\begin{align}\label{eqn:logistic def general}
p(w | D) = \frac{\exp{-\sum_{i=1}^n \rho \ell(w,x_i,y_i)} p(w)}{\int_{\mathbb{R}^d} \exp{-\sum_{i=1}^n \rho \ell(w',x_i,y_i)} p(w') dw'},
\end{align}
where in the case of logistic regression, $\ell$ is the logistic loss function.

Then we have the following lemma. 
\begin{lemma}\label{lem:LR_privacy_general}
Suppose $\ell(\cdot,x, y)$ is $L$-Lipschitz and convex, and $-\log p(w)$ is twice differentiable and $m$-strongly convex.  
Posterior sampling from \eqref{eqn:logistic def general} satisfies $(\lambda, \frac{2 \rho^2 L^2}{m}\lambda)$-\rdp\ for all $\lambda \geq 1$.
\end{lemma}

\begin{proof}(of Lemma~\ref{lem:LR_privacy_general})
The proof follows from the same idea as in the proof of Theorem 7 of \citep{without_sensitivity}.
The basic idea is that the posterior distribution $p(\cdot|D)$ satisfies Logarithmic Sobolev inequality (LSI), which implies sub-Gaussian concentration on $\log \frac{p(w|D)}{p(w|D')}$; and sub-Gaussian concentration implies \rdp.

Before the proof, we define LSI and introduce the relation between sub-Gaussian concentration and LSI.

\begin{definition}\label{def:LSI}
A distribution $\mu$ satisfies the Log-Sobolev Inquality (LSI) with constant $C$ if for any integrable function $f$, 
$$\Expect{\mu}{f^2 \log f^2} - \Expect{\mu}{f^2} \log \Expect{\mu}{f^2} \leq 2C \Expect{\mu}{\|\nabla f\|^2}. $$
\end{definition}

\begin{theorem}(Herbst's Theorem)\label{lem:herbst}
If $\mu$ satisfies LSI with constant C. Then for every $L$-Lipschitz function $f$, for any $\lambda$,
$\Expect{}{\exp{\lambda (f-\Expect{\mu}{f})}} \leq \exp{C \lambda^2 L^2/2}$.
\end{theorem}

\begin{lemma}\label{lem:log-concave_LSI}
Let $U: \mathbb{R}^d \rightarrow \mathbb{R}$ be a twice differential, $m$-strongly convex and integrable function. Let $\mu$ be a probability measure on $\mathbb{R}^d$ whose density is proportional to $\exp{-U}$. Then $\mu$ satisfies LSI with constant $C = 1/m$.
\end{lemma}

Now we prove \rdp\ bound of posterior sampling from \eqref{eqn:logistic def general}.

Firstly, notice that negative of log of the prior, $-\log p(w)$, is twice differentiable, $m$-strongly convex and integrable. And therefore negative of log of the posterior, $\rho \sum_{i=1}^n \ell(w,x_i,y_i) - \log p(w)$ is $m$-strongly convex.
According to Lemma \ref{lem:log-concave_LSI}, distribution $p(w|D)$ satisfies LSI with constant $1/m$.

Then, set $f$ in Theorem~\ref{lem:herbst} as $f(D,D',w) = \log \frac{p(w | D)}{p(w | D')}$. Since the $\ell(\cdot,x,y)$ is $L$-Lipschitz, we know that $f(D,D',w)$ is $2 \rho L$-Lipschitz. According to Theorem~\ref{lem:herbst}, for any $\lambda \in \mathbb{R}$,
\begin{align*}
\Expect{w \sim p(w | D)}{\exp{\lambda\left(\log \frac{p(w | D)}{p(w | D')} - \KL{p(w | D)}{p(w | D')}\right)}} \leq e^{2 \lambda^2 \rho^2 L^2/ m}.
\end{align*}

Let $a = 2 \rho^2 L^2/ m$. Equivalently, then for any $\lambda \in \mathbb{R}$,
\begin{align*}
\Expect{w \sim p(w | D)}{\exp{\lambda \log \frac{p(w | D)}{p(w | D')}}} \leq \exp{a \lambda^2 + \lambda \KL{p(w | D)}{p(w | D')}}.
\end{align*}
And setting $\lambda$ to $\lambda-1$, we have
\begin{align*}
\Expect{w \sim p(w | D)}{\exp{(\lambda-1) \log \frac{p(w | D)}{p(w | D')}}} 
\leq& \exp{a (\lambda-1)^2 + (\lambda-1) \KL{p(w | D)}{p(w | D')}} \\
\leq& \exp{(\lambda-1) \left(a \lambda + \KL{p(w | D)}{p(w | D')} - a\right)}.
\end{align*}
If $\lambda \geq 1$, the expectation is upper bounded by
\begin{align*}
\exp{(\lambda-1) \left(a \lambda + \max_{d(D,D')=1} \KL{p(w | D)}{p(w | D')} - a\right)}.
\end{align*}

According to the definition of zCDP in \citep{bun2016concentrated}, this implies zCDP with
\begin{align*}
\rho =& \frac{2 \rho^2 L^2}{m}, \\
\xi =& \max_{d(D,D')=1} \KL{p(w | D)}{p(w | D')} - \frac{2 \rho^2 L^2}{m},
\end{align*}
which is equivalent to $(\lambda, \frac{2 \rho^2 L^2}{m}\lambda + \max_{d(D,D')=1} \KL{p(w | D)}{p(w | D')} - \frac{2 \rho^2 L^2}{m})$-\rdp\ for any $\lambda \geq 1$.

Finally, we aim at bounding $\KL{p(w | D)}{p(w | D')}$. 
Let $F(w) = \frac{p(w | D)}{p(w | D')}$.
According to the definition of KL-divergence, we have
\begin{align*}
\KL{p(w | D)}{p(w | D')} 
= \Expect{p(w|D)}{\log F}
= \Expect{p(w|D')}{F\log F} - \Expect{p(w|D')}{F} \Expect{p(w|D')}{\log F},
\end{align*}
which, by setting $f = \sqrt{F}$ in Definition~\ref{def:LSI} and having $C = 1/m$, can be upper bounded by 
\begin{align}\label{eqn:lem:logistic proof}
\KL{p(w | D)}{p(w | D')} \leq
\frac{2}{m} \Expect{p(w | D')}{\|\nabla \sqrt{F}\|_2^2}.
\end{align}
We have
\begin{align*}
&\|\nabla \log F\|_2 \\
=& \rho \|\nabla\log{p(w | D)} - \nabla\log{p(w | D')}\|_2 \\
=& \rho \|\nabla \log(p(D|w)p(w)) - \nabla \log(p(D'|w)p(w)) \|_2 \\
=& \rho \|\nabla \log p(D|w) - \nabla \log p(D'|w) \|_2 \\
\leq& 2 \rho L,
\end{align*}
and therefore
\begin{align*}
\|\nabla \sqrt{F}\|_2^2
= \|\nabla \exp{\frac{1}{2}\log{F}}\|_2^2
= \|\frac{\sqrt{F}}{2} \nabla \log F\|_2^2
= \frac{F}{4} \|\nabla \log F\|_2^2
\leq \rho^2 L^2 F.
\end{align*}
So the KL-divergence in \eqref{eqn:lem:logistic proof} is upper bounded by
\begin{align*}
	 \frac{2 \rho^2 L^2}{m} \Expect{p(w | D')}{F}
=	 \frac{2 \rho^2 L^2}{m}.
\end{align*}


Therefore Bayesian logistic regression satisfies $(\lambda, \frac{2 \rho^2 L^2}{m}\lambda)$-\rdp\ for any $\lambda$.

For readers familiar with the proof of Theorem 7 in \citep{without_sensitivity}, the proof here is exactly the same except that the tail bound of sub-Gaussian concentration in Equation 21 and consequently 25 there are replaced by the moment generating function bound.
The reason for not using the tail bound to imply moment generating function bound is because that loses constant factor.  
\end{proof}

For GLMs, we have
\begin{align*}
\ell(w,x,y) = -\log h(y) + A(w^\top x) - y w^\top x,
\end{align*}
and thus
\begin{align*}
\nabla_w \ell(w,x,y) = (\mu - y) x = (g^{-1}(w^\top x) - y) x.
\end{align*}
Then, by the condition in Theorem 16 and 17,
$\|\nabla_w \ell(w,x,y)\|_2$ is upper bounded by $B c$ and $\ell(\cdot,x,y)$ is $B c$-Lipschitz.


\subsubsection{Logistic Regression Tightness}
\begin{proof}(of Theorem~\ref{lem:LR_tightness})
We aim to upper bound
\begin{align}\label{eqn:term1 times term2}
\int \frac{p(w|D)^{\lambda}\hfill}{p(w|D')^{\lambda-1}} dw = 
\int p(w) \frac{p(D | w)^{\lambda}\hfill}{p(D' | w)^{\lambda-1}} dw \times
\frac{[\int p(w) p(D' | w) dw]^{\lambda-1}}{[\int p(w) p(D | w) dw]^{\lambda}\hfill}
\end{align}
for all $\lambda > 1$.

For convenience, here we assume $\calY = \{-1, 1\}$ instead of $0, 1$, and thus $p(y|w,x)$ can be written as $1/(1 + e^{-y w^\top x})$. Let $\sigma^2 = (n\reg)^{-1}$ denote the variance of the Gaussian prior.

Firstly, we consider the case when $|D| = 1$. We will extend the analysis to $|D| > 1$ later.

Consider any $x \in \mathbb{R}^d$ with $\|x\|=c$. Let $D = \{(x, y)\}$ and $D' = \{(x', y')\}$, where $x'=x$ and $y' = -y = -1$. 

Firstly, we prove an equation that will be used later. Let $*_j$ be the $j$-th dimension of vector $*$.  For any $i$ and $d$, we have
\begin{align}\label{eqn:term1 proof basic}
&	\frac{1}{\sqrt{2\pi\sigma^2}^d} \int_{\mathbb{R}^d} \exp{-\frac{\|w\|^2}{2\sigma^2}} \exp{-i w^\top x} dw \\
=&	\prod_{j=1}^{d} \frac{1}{\sqrt{2\pi\sigma^2}} \int_{\mathbb{R}} \exp{-\frac{w_j^2}{2\sigma^2}} \exp{-i w_j x_j} dw_j \nonumber\\
=&	\prod_{j=1}^{d} \frac{1}{\sqrt{2\pi\sigma^2}} \int_{\mathbb{R}} \exp{-\frac{(w_j + i x_j \sigma^2)^2}{2\sigma^2}} \exp{\frac{i^2 x_j^2 \sigma^2}{2}} dw_j \nonumber\\
=&	\prod_{j=1}^{d} \exp{\frac{i^2 x_j^2 \sigma^2}{2}} \nonumber\\
=&	\exp{\frac{i^2 \sigma^2 \|x\|_2^2}{2}}.\nonumber
\end{align}

Now we will consider the two terms in \eqref{eqn:term1 times term2} separately.

For the first term, we have
\begin{align}\label{eqn:term1}
&	\int_{\mathbb{R}^d} p(w) \frac{p(D | w)^{\lambda}\hfill}{p(D' | w)^{\lambda-1}} dw \\
=&	\frac{1}{\sqrt{2\pi\sigma^2}^d} \int_{\mathbb{R}^d} \exp{-\frac{\|w\|^2}{2\sigma^2}} \frac{(1+\exp{-y' w^\top x'})^{\lambda-1}}{(1+\exp{-y w^\top x})^{\lambda}} dw \nonumber\\
=&	\frac{1}{\sqrt{2\pi\sigma^2}^d} \int_{\mathbb{R}^d} \exp{-\frac{\|w\|^2}{2\sigma^2}} \frac{(1+\exp{w^\top x})^{\lambda-1}}{(1+\exp{-w^\top x})^{\lambda}} dw \nonumber\\
=&	\frac{1}{\sqrt{2\pi\sigma^2}^d} \int_{\mathbb{R}^d} \exp{-\frac{\|w\|^2}{2\sigma^2}} \frac{\exp{(\lambda-1) w^\top x}}{1+\exp{-w^\top x}} dw.\nonumber
\end{align}
Let $S_{+}$ be any half-space of $\mathbb{R}^d$ and $S_{-} = \mathbb{R}^d \backslash S_{+}$. The above equals to
\begin{align*}
&	\frac{1}{\sqrt{2\pi\sigma^2}^d} \int_{S_{+}} \exp{-\frac{\|w\|^2}{2\sigma^2}} \frac{\exp{(\lambda-1) w^\top x}}{1+\exp{-w^\top x}} dw + \frac{1}{\sqrt{2\pi\sigma^2}^d} \int_{S_{-}} \exp{-\frac{\|w\|^2}{2\sigma^2}} \frac{\exp{(\lambda-1) w^\top x}}{1+\exp{-w^\top x}} dw
\end{align*}
For any $x$ and any $w\in S_{-}$, we have $-y w^\top x = y (-w)^\top x$. By changing variable in the second integral from $w$ to $-w$, the above equals to
\begin{align*}
&	\frac{1}{\sqrt{2\pi\sigma^2}^d} \int_{S_{+}} \exp{-\frac{\|w\|^2}{2\sigma^2}} \frac{\exp{(\lambda-1) w^\top x}}{1+\exp{-w^\top x}} dw + \frac{1}{\sqrt{2\pi\sigma^2}^d} \int_{S_{+}} \exp{-\frac{\|w\|^2}{2\sigma^2}} \frac{\exp{-(\lambda-1) w^\top x}}{1+\exp{w^\top x}} dw\\
=&	\frac{1}{\sqrt{2\pi\sigma^2}^d} \int_{S_{+}} \exp{-\frac{\|w\|^2}{2\sigma^2}} \left(\frac{\exp{(\lambda-1) w^\top x}}{1+\exp{-w^\top x}} + \frac{\exp{-(\lambda-1) w^\top x}}{1+\exp{w^\top x}}\right) dw\\
=&	\frac{1}{\sqrt{2\pi\sigma^2}^d} \int_{S_{+}} \exp{-\frac{\|w\|^2}{2\sigma^2}} \sum_{i=-\lambda+1}^{\lambda-1} \exp{-i w^\top x} (-1)^{i+\lambda-1}  dw\\
=&	\sum_{i=-\lambda+1}^{-1} \frac{(-1)^{i+\lambda-1}}{\sqrt{2\pi\sigma^2}^d} \int_{S_{+}} \exp{-\frac{\|w\|^2}{2\sigma^2}} \exp{-i w^\top x} dw + \frac{(-1)^{\lambda-1}}{\sqrt{2\pi\sigma^2}^d} \int_{S_{+}} \exp{-\frac{\|w\|^2}{2\sigma^2}} dw \\
 &+ \sum_{i=1}^{\lambda-1} \frac{(-1)^{i+\lambda-1}}{\sqrt{2\pi\sigma^2}^d} \int_{S_{+}} \exp{-\frac{\|w\|^2}{2\sigma^2}} \exp{-i w^\top x} dw.
\end{align*}
The middle term equals to ${(-1)^{\lambda-1}}/{2}$. And changing variable from $w$ to $-w$ in the first term, the above equals to
\begin{align*}
=&	\sum_{i=1}^{\lambda-1} \frac{(-1)^{-i+\lambda-1}}{\sqrt{2\pi\sigma^2}^d} \int_{S_{-}} \exp{-\frac{\|w\|^2}{2\sigma^2}} \exp{-i w^\top x} dw \\
 &+ \sum_{i=1}^{\lambda-1} \frac{(-1)^{i+\lambda-1}}{\sqrt{2\pi\sigma^2}^d} \int_{S_{+}} \exp{-\frac{\|w\|^2}{2\sigma^2}} \exp{-i w^\top x} dw + \frac{(-1)^{\lambda-1}}{2}\\
=&	\sum_{i=1}^{\lambda-1} \frac{(-1)^{-i+\lambda-1}}{\sqrt{2\pi\sigma^2}^d} \int_{\mathbb{R}^d} \exp{-\frac{\|w\|^2}{2\sigma^2}} \exp{-i w^\top x} dw + \frac{(-1)^{\lambda-1}}{2}.
\end{align*}

Using the equation in \eqref{eqn:term1 proof basic} with the fact that $\|x\|_2 = c$, the above equals to
\begin{align*}
	\sum_{i=1}^{\lambda-1} (-1)^{-i+\lambda-1} \exp{\frac{i^2 \sigma^2 c^2}{2}} + \frac{(-1)^{\lambda-1}}{2}
=	\sum_{i=0}^{\lambda-1} (-1)^{-i+\lambda-1} \exp{\frac{i^2 \sigma^2 c^2}{2}}.
\end{align*}

Now we consider the second term in \eqref{eqn:term1 times term2}. We have
\begin{align*}
&	\int p(w) p(D | w) dw 
=	\frac{1}{\sqrt{2\pi\sigma^2}^d} \int_{\mathbb{R}^d} \exp{-\frac{\|w\|^2}{2\sigma^2}} \frac{1}{1+\exp{-y w^\top x}} dw. 
\end{align*}
Let $S_{+}$ be any half-space of $\mathbb{R}^d$ and $S_{-} = \mathbb{R}^d \backslash S_{+}$. The above equals to
\begin{align*}
&	\frac{1}{\sqrt{2\pi\sigma^2}^d} \int_{S_{+}} \exp{-\frac{\|w\|^2}{2\sigma^2}} \frac{1}{1+\exp{-y w^\top x}} dw + \frac{1}{\sqrt{2\pi\sigma^2}^d} \int_{S_{-}} \exp{-\frac{\|w\|^2}{2\sigma^2}} \frac{1}{1+\exp{-y w^\top x}} dw.
\end{align*}
For any $x$ and any $w\in S_{-}$, we have $-y w^\top x = y (-w)^\top x$. By changing variable in the second integral from $w$ to $-w$, the above equals to
\begin{align*}
&	\frac{1}{\sqrt{2\pi\sigma^2}^d} \int_{S_{+}} \exp{-\frac{\|w\|^2}{2\sigma^2}} \frac{1}{1+\exp{-y w^\top x}} dw + \frac{1}{\sqrt{2\pi\sigma^2}^d} \int_{S_{+}} \exp{-\frac{\|w\|^2}{2\sigma^2}} \frac{1}{1+\exp{y w^\top x}} dw \\
=&	\frac{1}{\sqrt{2\pi\sigma^2}^d} \int_{S_{+}} \exp{-\frac{\|w\|^2}{2\sigma^2}} \left(\frac{1}{1+\exp{-y w^\top x}} + \frac{1}{1+\exp{y w^\top x}}\right) dw \\
=&	\frac{1}{\sqrt{2\pi\sigma^2}^d} \int_{S_{+}} \exp{-\frac{\|w\|^2}{2\sigma^2}} dw \\
=&	\frac{1}{2}.
\end{align*}
Since the above result holds for any dataset $D$, we know that the second term of \eqref{eqn:term1 times term2} equals to $2$.

Combining the two terms, \eqref{eqn:term1 times term2} equals to
\begin{align*}
2 \sum_{i=0}^{\lambda-1} (-1)^{-i+\lambda-1} \exp{\frac{i^2 \sigma^2 c^2}{2}}.
\end{align*}
As $\lambda$ becomes sufficiently large, this is lower bounded by
$$\exp{\frac{(\lambda-1)^2 \sigma^2 c^2}{2}}. $$

Now we consider the case when $|D| > 1$. We aim to show that there exists $D$ and $D'$, such that the same results hold for them.

Let $D = \{(x,y),(x^1,y^1),\dots,(x^{n-1},y^{n-1})\}$ and $D' = \{(x',y'),(x^1,y^1),\dots,(x^{n-1},y^{n-1})\}$, i.e., $D$ and $D'$ differ in the first example. 
Let $S_1 \subset [d]$, $|S_1| < d$ be a set of dimensions and let $S_2 = [d]\backslash S_1$. Let $x=x'$ be a vector with non-zero dimensions only on $S_1$, i.e., $x_i = 0$, $\forall i\in S_2$, and $\|x\|_2 = c$. 
Let $w_{S_*}$ denote the dimensions of $w$ that belongs to $S_*$. We then have $\|x_{S_1}\|_2 = c$.
Let $y' = -y = -1$. 
Suppose $\forall j\in [n-1]$, $x^j$ has non-zero dimensions only at $\{1,\dots,d\}\backslash S_*$, i.e., $x^j_i = 0$, $\forall i \in S_*$.

We have
\begin{align*}
&	\int p(w) p(D | w) dw \\
=&	\frac{1}{\sqrt{2\pi\sigma^2}^d} \int \exp{-\frac{\|w\|^2}{2\sigma^2}} \frac{1}{1+\exp{-y^j w^\top x}} \prod_j \frac{1}{1+\exp{-y^j w^\top x^j}} dw \\
=&	\frac{1}{\sqrt{2\pi\sigma^2}^{|S_1|}} \int \exp{-\frac{\|w_{S_1}\|^2}{2\sigma^2}} \frac{1}{1+\exp{-y^j w_{S_1}^\top x_{S_1}}} dw_{S_1}\\
\times& \frac{1}{\sqrt{2\pi\sigma^2}^{|S_2|}} \int \exp{-\frac{\|w_{S_2}\|^2}{2\sigma^2}} \prod_j \frac{1}{1+\exp{-y^j w_{S_2}^\top x^j_{S_2}}} dw_{S_2}\\
=&	\frac{1}{2} \frac{1}{\sqrt{2\pi\sigma^2}^{|S_2|}} \int \exp{-\frac{\|w_{S_2}\|^2}{2\sigma^2}} \prod_j \frac{1}{1+\exp{-y^j w_{S_2}^\top x^j_{S_2}}} dw_{S_2}
\end{align*}
where the second step follows from the fact that $x$ is non-zero only at $S_1$ and $s^j$ is non-zero only at $S_2$, and the
last step follows from the normalization term of $|D|=1$ (second term of \eqref{eqn:term1 times term2}).

We also have
\begin{align*}
&	\int p(w) \frac{p(D | w)^{\lambda}}{p(D' | w)^{\lambda-1}} dw \\
=&	\frac{1}{\sqrt{2\pi\sigma^2}^d} \int \exp{-\frac{\|w\|^2}{2\sigma^2}} \frac{(1+\exp{-y' w^\top x'})^{\lambda-1}}{(1+\exp{-y w^\top x})^{\lambda}} \prod_j \frac{1}{1+\exp{-y^j w^\top x^j}} dw \\
=&	\frac{1}{\sqrt{2\pi\sigma^2}^d} \int \exp{-\frac{\|w\|^2}{2\sigma^2}} \frac{(1+\exp{w^\top x})^{\lambda-1}}{(1+\exp{-w^\top x})^{\lambda}} \prod_j \frac{1}{1+\exp{-y^j w^\top x^j}} dw \\
=&	\frac{1}{\sqrt{2\pi\sigma^2}^{|S_1|}} \int \exp{-\frac{\|w_{S_1}\|^2}{2\sigma^2}} \frac{(1+\exp{w_{S_1}^\top x_{S_1}})^{\lambda-1}}{(1+\exp{-w_{S_1}^\top x_{S_1}})^{\lambda}} dw_{S_1} \\
\times&
	\frac{1}{\sqrt{2\pi\sigma^2}^{|S_2|}} \int \exp{-\frac{\|w_{S_2}\|^2}{2\sigma^2}} \prod_j \frac{1}{1+\exp{-y^j w_{S_1}^\top x^j_{S_1}}} dw_{S_2}\\
\leq&	\exp{\exp{\frac{(\lambda-1)^2 \sigma^2 c^2}{2}}} \times \frac{1}{\sqrt{2\pi\sigma^2}^{|S_2|}} \int \exp{-\frac{\|w_{S_2}\|^2}{2\sigma^2}} \prod_j \frac{1}{1+\exp{-y^j w_{S_1}^\top x^j_{S_1}}} dw_{S_2}
\end{align*}
where the last step follows from the calculation of the first term of $|D|=1$ since $x' = x$, $\|x_{S_1}\|_2=c$.

Combining them together, the integration term for $x^j$s cancelled and the Renyi divergence is $\frac{1}{\lambda-1} \log \left(2{\exp{\frac{(\lambda-1)^2 \sigma^2 c^2}{2}}}\right) = O(\frac{\sigma^2 c^2 (\lambda-1)}{2})$, the same as that at $|D|=1$.

\end{proof}

\subsection{Additional Experiments for GLMs}
In this section, we present more experimental results on the same datasets.

\begin{figure*}
\centering
\begin{subfigure}[b]{0.325\textwidth}
\includegraphics[width=\textwidth]{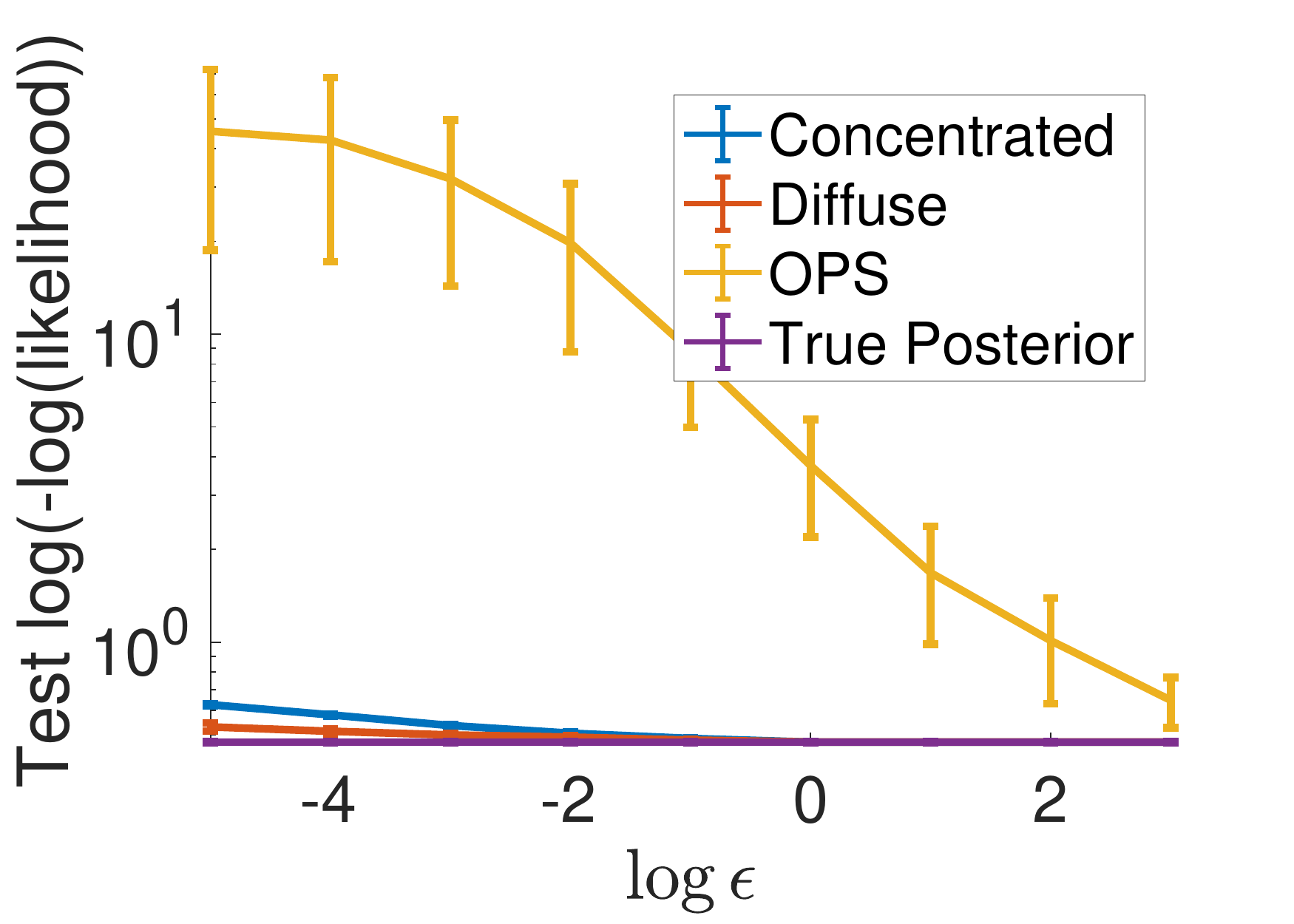}
\end{subfigure}
\hfill
\begin{subfigure}[b]{0.325\textwidth}
\includegraphics[width=\textwidth]{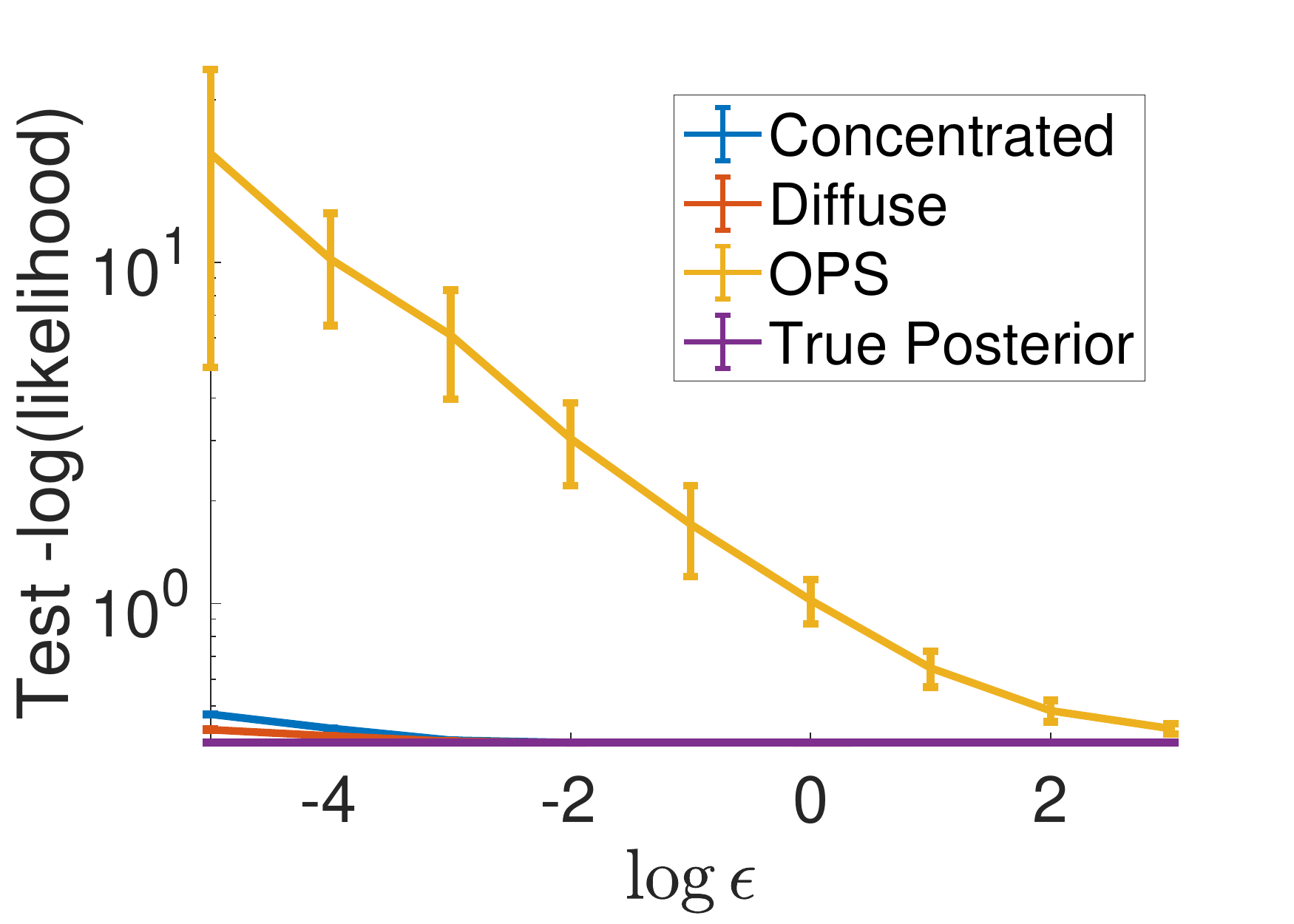}
\end{subfigure}
\hfill
\begin{subfigure}[b]{0.325\textwidth}
\includegraphics[width=\textwidth]{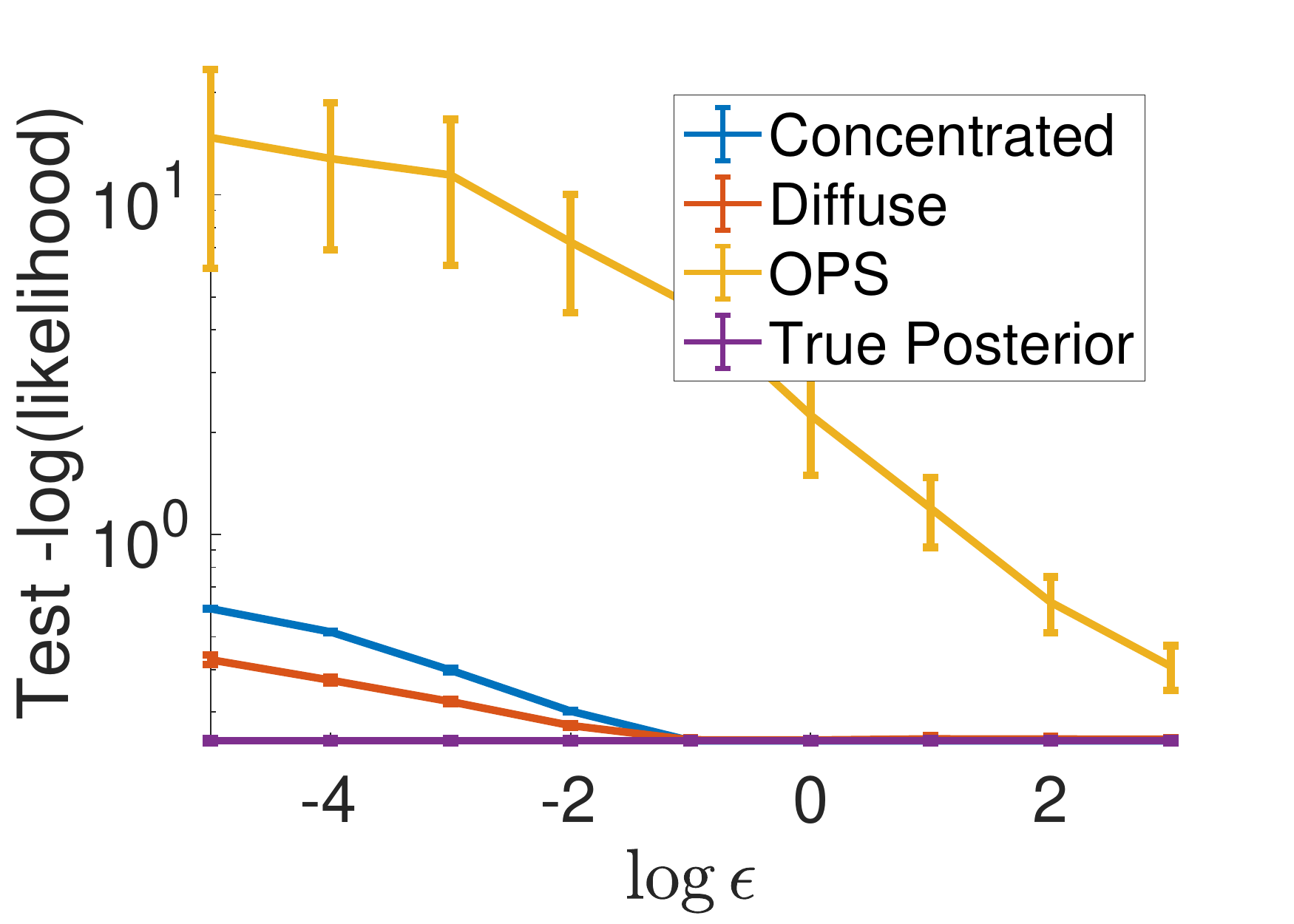}
\end{subfigure}
%
\begin{subfigure}[b]{0.325\textwidth}
\includegraphics[width=\textwidth]{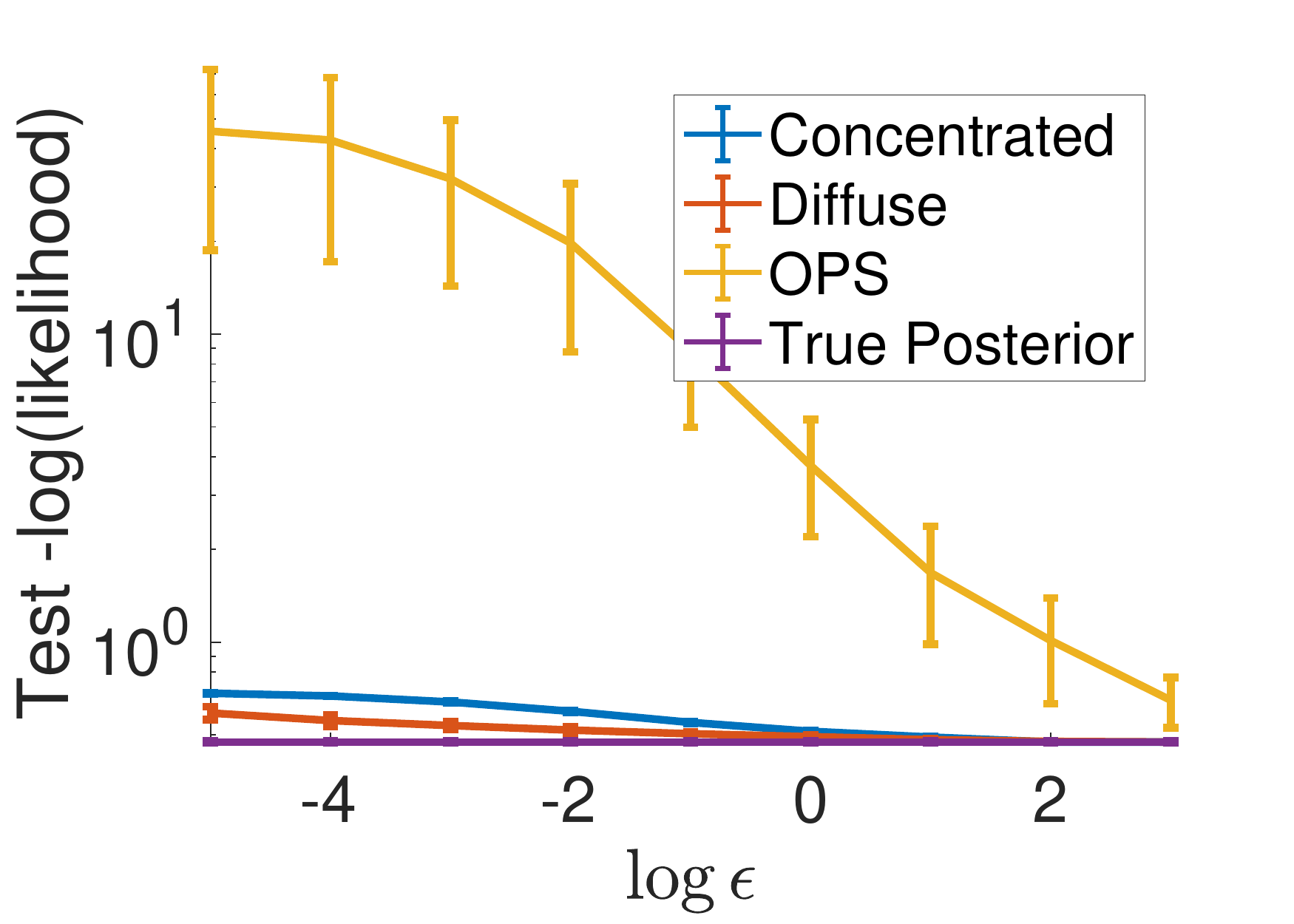}
\end{subfigure}
\hfill
\begin{subfigure}[b]{0.325\textwidth}
\includegraphics[width=\textwidth]{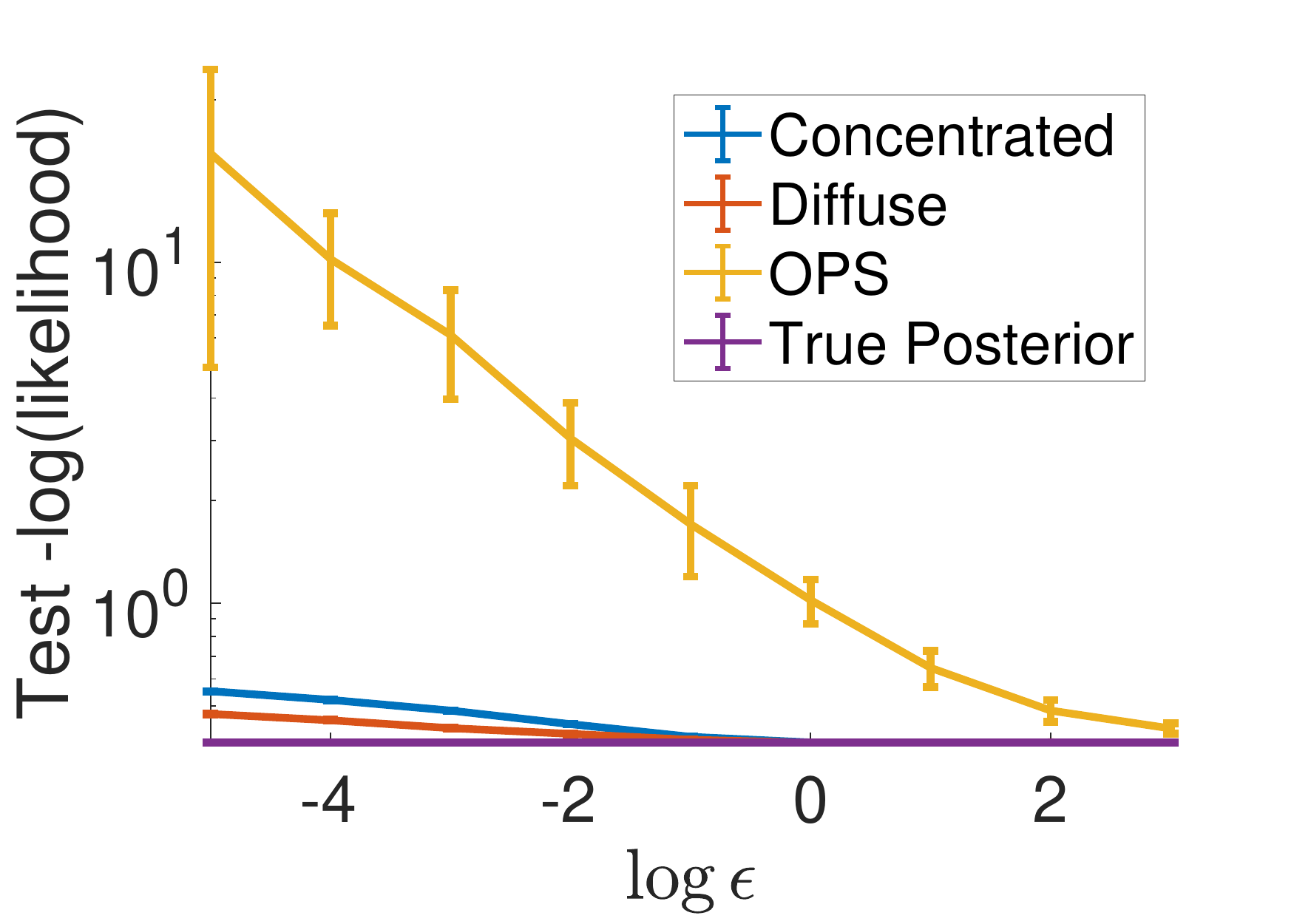}
\end{subfigure}
\hfill
\begin{subfigure}[b]{0.325\textwidth}
\includegraphics[width=\textwidth]{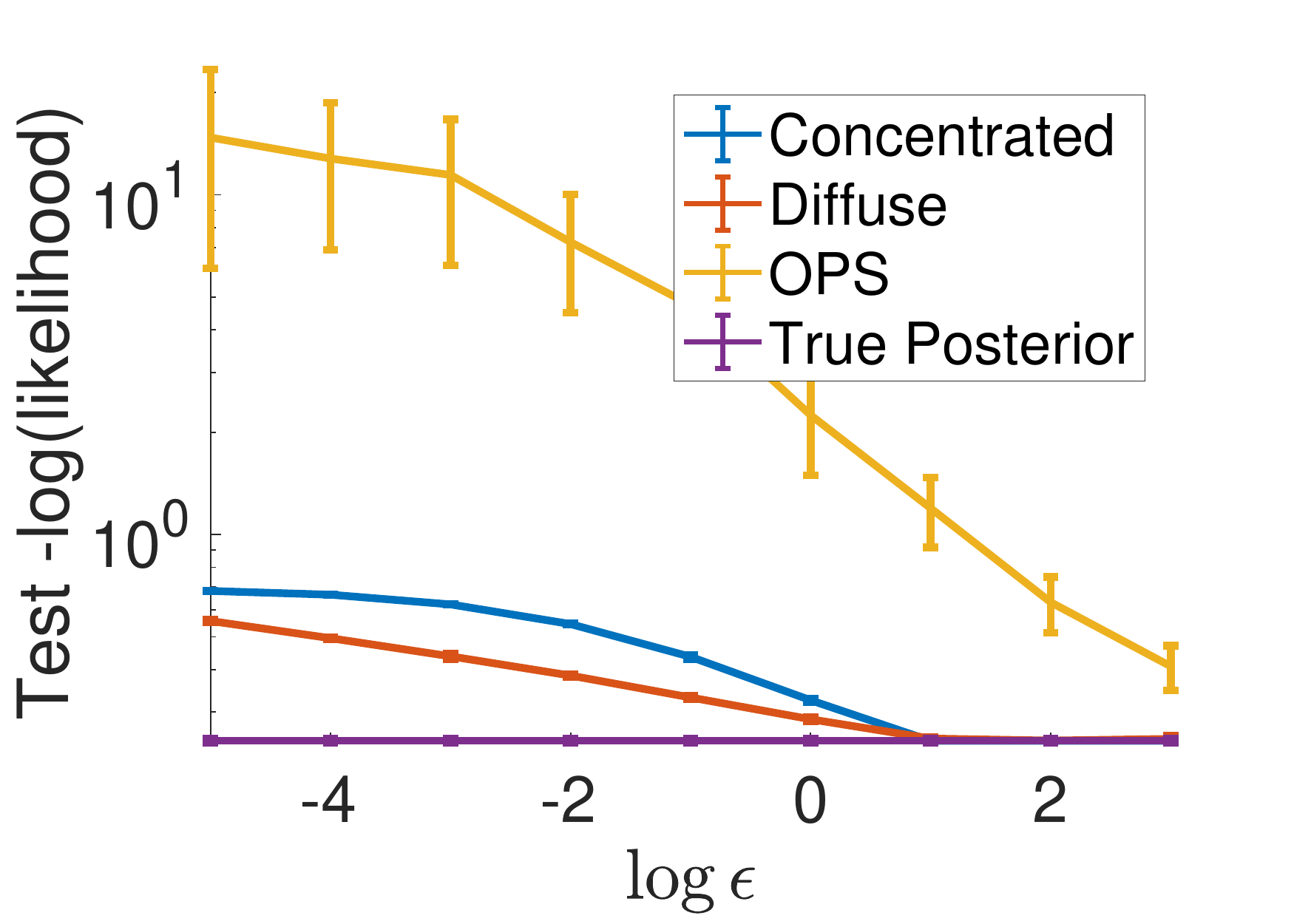}
\end{subfigure}
\begin{subfigure}[b]{0.325\textwidth}
\includegraphics[width=\textwidth]{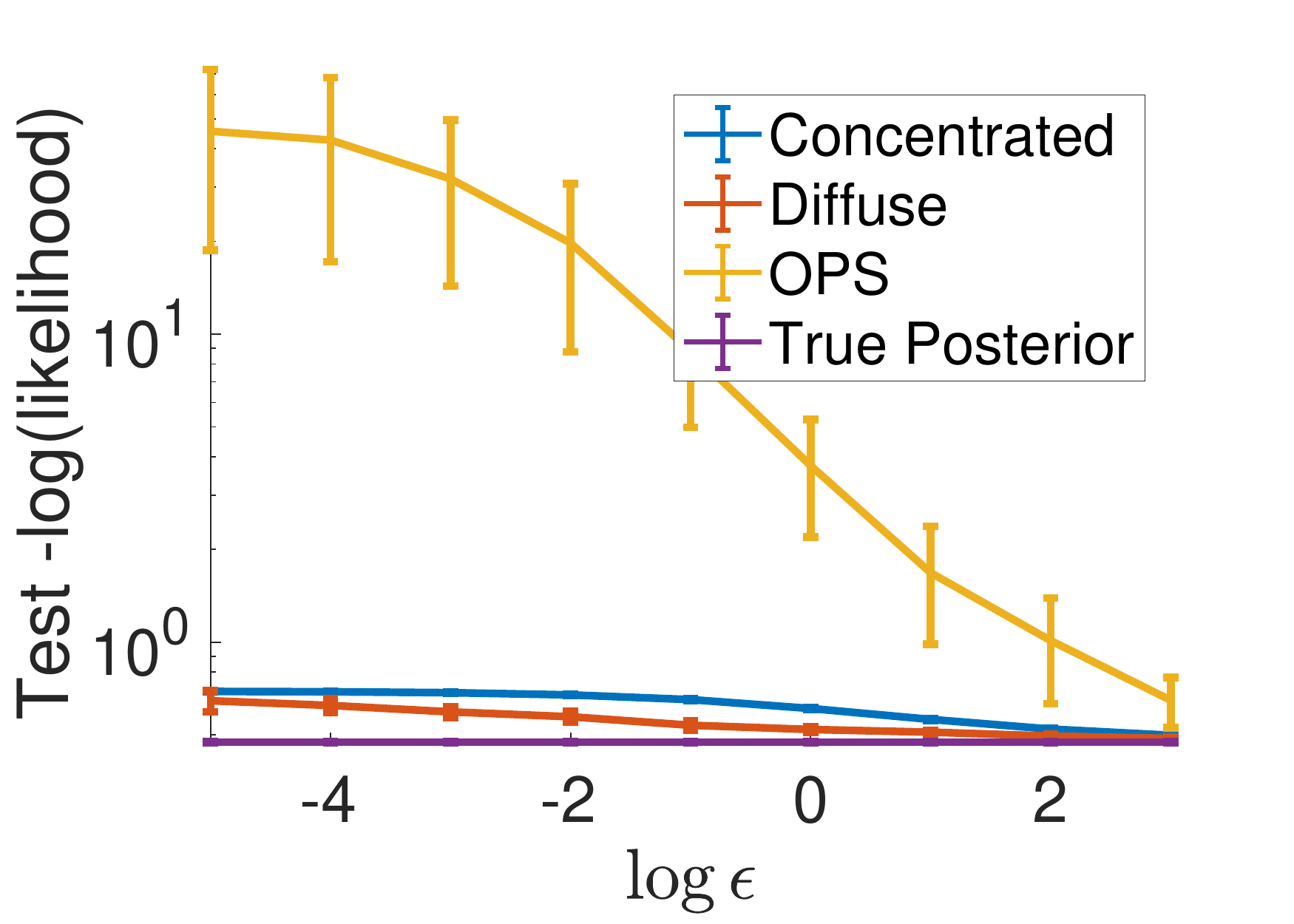}
\vspace{-15pt}\caption{Abalone.}
\end{subfigure}
\hfill
\begin{subfigure}[b]{0.325\textwidth}
\includegraphics[width=\textwidth]{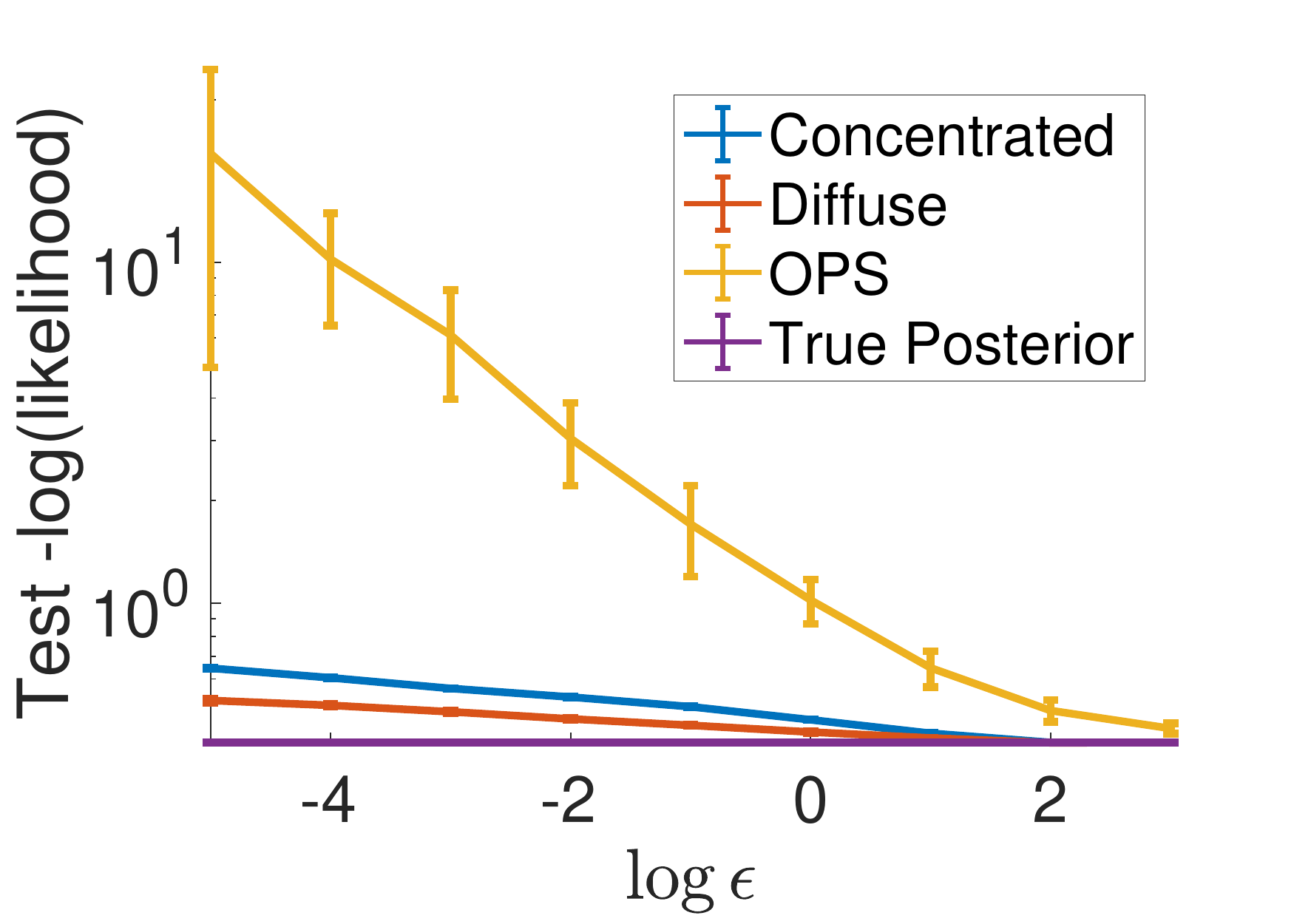}
\vspace{-15pt}\caption{Adult.}
\end{subfigure}
\hfill
\begin{subfigure}[b]{0.325\textwidth}
\includegraphics[width=\textwidth]{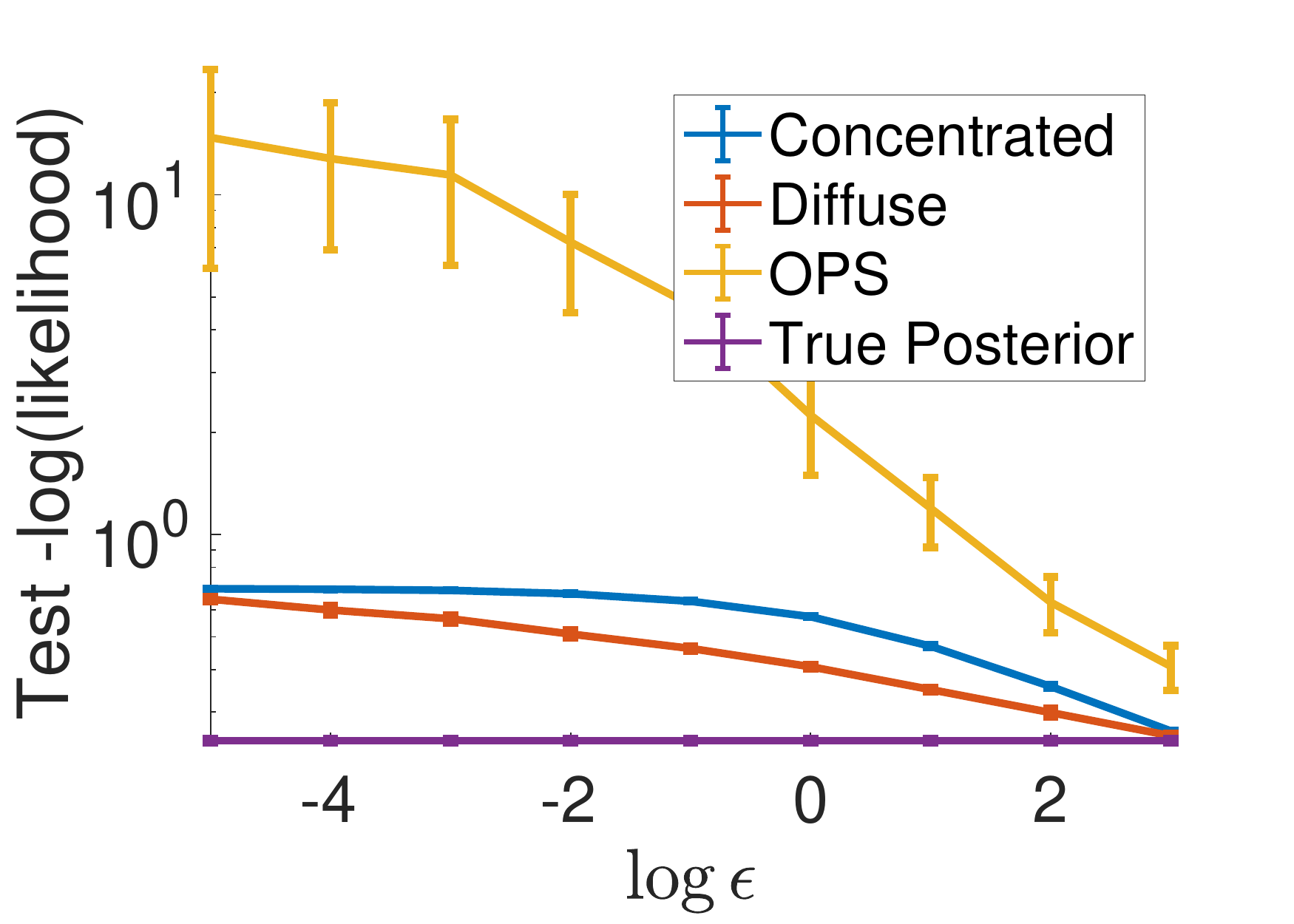}
\vspace{-15pt}\caption{MNIST 3vs8.}
\end{subfigure}
\caption{Negative log-likelihood vs. privacy parameter $\epsilon$. $\lambda=1$, $10$, $100$ from top to bottom. y-axis plotted in log scale.}
\label{fig:lr_likelihood_appendix}
\end{figure*}

We show the negative log-likelihood at $\lambda \in \{1,10,100\}$ in Figure~\ref{fig:lr_likelihood_appendix}. We can see the same trend as that in test error. Both of our proposed algorithms achieves smaller negative log-likelihood, and the diffused algorithm achieves lower negative log-likelihood than the concentrate algorithm.


\end{document}